\documentclass{article}

\usepackage[preprint]{neurips_2025}

\usepackage[utf8]{inputenc} 
\usepackage[T1]{fontenc}    
\usepackage{hyperref}       
\usepackage{url}            
\usepackage{booktabs}       
\usepackage{amsfonts}       
\usepackage{nicefrac}       
\usepackage{microtype}      
\usepackage{xcolor}         

\usepackage[utf8]{inputenc} 
\usepackage[T1]{fontenc}    
\usepackage{hyperref}       
\usepackage{url}            
\usepackage{booktabs}       
\usepackage{amsfonts}       
\usepackage{nicefrac}       
\usepackage{microtype}      
\usepackage{xcolor}         

\usepackage{hyperref}
\usepackage{footnote}
\usepackage{caption}
\usepackage{tikz}

\usepackage{mathrsfs}

\usepackage{graphicx}
\usepackage{wrapfig}
\usepackage{caption}
\usepackage{lipsum} 

\usepackage{algpseudocode}
\usepackage[ruled,linesnumbered]{algorithm2e}
\usepackage{amsmath} 
\usepackage{array}
\usepackage{tabularx}
\usepackage{multirow}
\usepackage[most]{tcolorbox}
\usepackage{makecell}
\usepackage{colortbl}
\usepackage{balance}
\usepackage{float}
\usepackage{graphicx}
\usepackage{subcaption}
\usepackage{diagbox}

\usepackage[ruled,linesnumbered]{algorithm2e}

\newcommand{\Rmnum}[1]{\expandafter\@slowromancap\romannumeral #1@}

\newcommand{\etal}{\emph{et~al.}\xspace}
\newcommand{\eg}{\emph{e.g.},\xspace}

\DeclareMathAlphabet\mathbfcal{OMS}{cmsy}{b}{n}

\newcommand{\eat}[1]{}
\ifodd 1

\else

\fi

\usepackage{pifont}%
\tcbset{
  aibox/.style={
    width=400.18663pt,
    top=10pt,
    colback=black!05,
    colframe=black!20,
    colbacktitle=black!25!gray,
    enhanced,
    center,
    attach boxed title to top center={yshift=-0.1in},
    boxed title style={boxrule=0pt,colframe=white,},
  }
}

\tcbset{
  aiboxnotitle/.style={
    width=400.18663pt,
    top=10pt,
    colback=black!05,
    colframe=black!20,
    enhanced,
    center,
  }
}

\tcbset{
  aiboxbreakable/.style={
    width=400.18663pt,
    top=10pt,
    colback=black!05,
    colframe=black!20,
    colbacktitle=black!50,
    enhanced,
    center,
    breakable,
    attach boxed title to top left={yshift=-0.1in,xshift=0.15in},
    boxed title style={boxrule=0pt,colframe=white,},
  }
}
\newtcolorbox{AIBox}[2][]{aibox,title=#2,#1}
\newtcolorbox{AIBoxNoTitle}[1][]{aiboxnotitle}
\newtcolorbox{AIBoxBreak}[2][]{aiboxbreakable,title=#2,#1}
\usepackage{alltt}
\usepackage{listings}
\title{MM-Agent: LLM as Agents for Real-world Mathematical Modeling   Problem}

\renewcommand\footnotemark{}
\author{
Fan Liu$^{*1}$, Zherui Yang$^{*1}$, Cancheng Liu$^1$, Tianrui Song$^1$, Xiaofeng Gao$^2$, Hao Liu$^{\dagger1}$
\thanks{$^*$Equal contribution.  $^{\dagger}$Correspondence to Hao Liu. }\\
$^1$AI Thrust, The Hong Kong University of Science and Technology (Guangzhou)\\ 
$^2$Department
of Computer Science and Engineering, Shanghai Jiao Tong University\\
\footnotesize{\texttt{fliu236@hkust-gz.connect.edu.cn;}}
\footnotesize{\texttt{yangzhr9@outlook.com;}}\\
\footnotesize{\texttt{\{canchengliu,tsong847\}@hkust-gz.connect.edu.cn;}}\\
\footnotesize{\texttt{gao-xf@cs.sjtu.edu.cn;}\texttt{liuh@ust.hk}
}
}

\begin{document}

\maketitle

\vspace{-0.1in}

\begin{abstract}

Mathematical modeling is a cornerstone of scientific discovery and engineering practice, enabling the translation of real-world problems into formal systems across domains such as physics, biology, and economics. Unlike mathematical reasoning, which assumes a predefined formulation, modeling requires open-ended problem analysis, abstraction, and principled formalization. While Large Language Models (LLMs) have shown strong reasoning capabilities, they fall short in rigorous model construction, limiting their utility in real-world problem-solving. To this end, we formalize the task of LLM-powered real-world mathematical modeling, where agents must analyze problems, construct domain-appropriate formulations, and generate complete end-to-end solutions. We introduce MM-Bench, a curated benchmark of 111 problems from the Mathematical Contest in Modeling (MCM/ICM)~\footnote{\url{https://www.contest.comap.com/undergraduate/contests/index.html}}, spanning the years 2000 to 2025 and across ten diverse domains such as physics, biology, and economics.  
To tackle this task, we propose MM-Agent, an expert-inspired framework that decomposes mathematical modeling into four stages: open-ended problem analysis, structured model formulation, computational problem solving, and report generation.
Experiments on MM-Bench show that MM-Agent significantly outperforms baseline agents, achieving an 11.88\% improvement over human expert solutions while requiring only 15 minutes and \$0.88 per task using GPT-4o. Furthermore, under official MCM/ICM protocols, MM-Agent assisted two undergraduate teams in winning the Finalist Award (\textbf{top 2.0\% among 27,456 teams}) in MCM/ICM 2025, demonstrating its practical effectiveness as a modeling copilot. 

\noindent \includegraphics[height=1.5em]{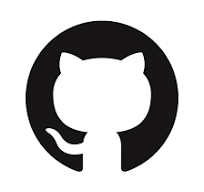} \href{https://github.com/usail-hkust/LLM-MM-Agent}{Code is available at \url{https://github.com/usail-hkust/LLM-MM-Agent} }

\noindent \includegraphics[height=1.5em]{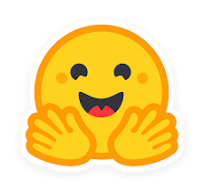} \href{https://huggingface.co/spaces/MathematicalModelingAgent/MathematicalModelingAgent}{Demo is available at \url{https://huggingface.co/spaces/MathematicalModelingAgent/MathematicalModelingAgent} }

\end{abstract}
\label{abs}
\vspace{-0.1in}



\vspace{-0.1in}
\section{Introduction}

\begin{figure}[!t]
    \centering
    \includegraphics[width=\linewidth]{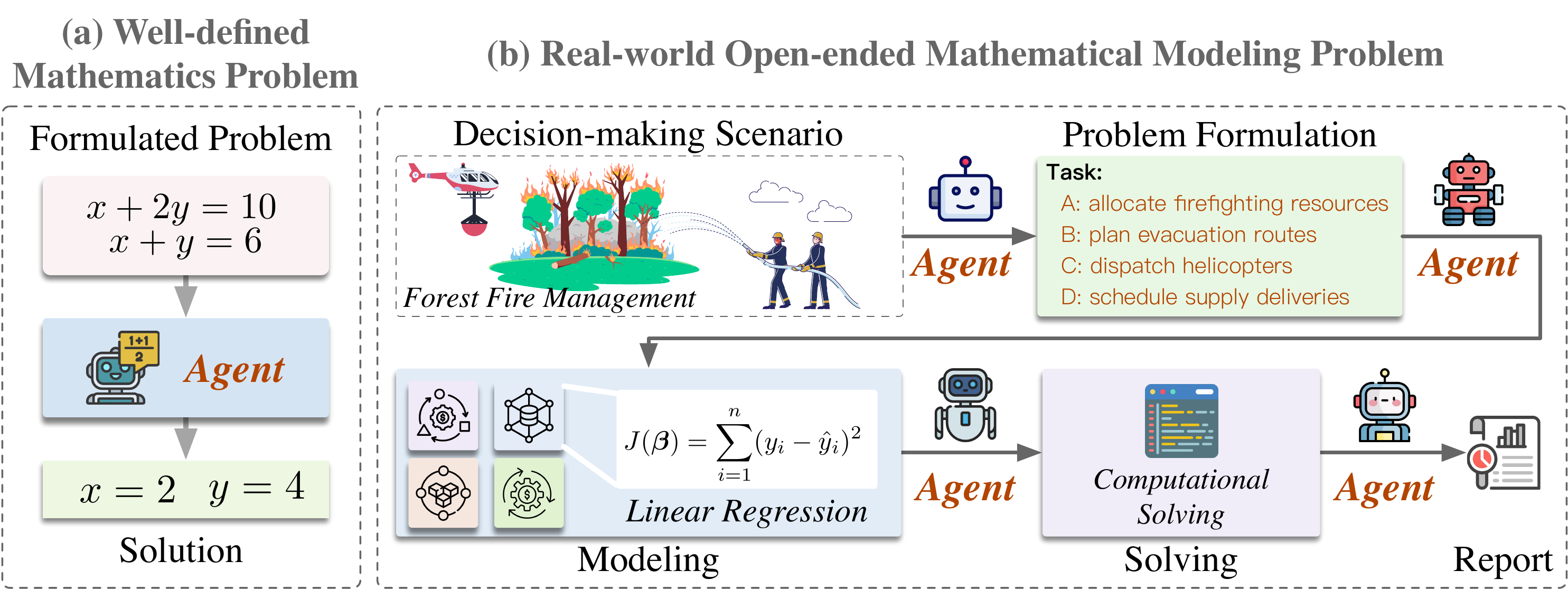}
    \caption{
    Traditional well-defined mathematics problem vs LLM-powered open-ended mathematical modeling problem.
    \textbf{Left}: A well-defined mathematical problem, where an agent solves a well-defined problem to obtain a solution. \textbf{Right}: \emph{An open-ended mathematical modeling problem}, where given an abstract application scenario or phenomenon, the agent first needs to formulate the mathematical problem before solving it and providing an end-to-end solution. }
    \label{fig:difference}
\vspace{-0.2in}
\end{figure}

Mathematical modeling serves as a cornerstone methodology for formulating, analyzing, and solving complex real-world problems, underpinning scientific discovery and technological advancement across applied mathematics, natural sciences, engineering, and the social sciences. In practice, this process often begins by identifying the core problem, abstracting it into a mathematical form, constructing appropriate models, and solving them to generate actionable insights. It enables the transformation of ill-posed challenges, such as epidemic control, energy forecasting, and supply chain management, into mathematical systems that support analysis, prediction, and decision-making through abstraction, theoretical formulation, empirical validation, and iterative refinement~\cite{bender2000introduction, meerschaert2013mathematical}. Unlike mathematical reasoning, which starts from fixed formulations, mathematical modeling demands open-ended problem abstraction, assumption design, and domain-grounded interpretation, making it context-sensitive and hard to automate, as shown in Figure~\ref{fig:difference}. Recent advances in Large Language Models (LLMs) offer new opportunities to automate parts of this workflow, showing promise in symbolic reasoning, scientific problem-solving, and numerical computation~\cite{nature_TrinhWLHL24, yang2025optibench, starace2025paperbench}. Developing LLM-based modeling agents could unlock scalable, efficient solutions across disciplines. However, current agents often fail to capture essential modeling principles, such as abstraction, constraints, and assumptions, leading to oversimplified and scientifically invalid outputs. As shown in Table~\ref{tab:main_exps}, they frequently omit key assumptions, producing models with limited real-world validity.

To address this gap, we formally define the task of LLM-powered real-world mathematical modeling, which requires agents to translate complex real-world problems or phenomena into structured and executable modeling pipelines, culminating in complete analytical reports.  To enable systematic evaluation, we introduce MM-Bench, a new benchmark constructed from 111 real-world problems adapted from MCM/ICM, spanning the years 2000 to 2025. MM-Bench covers ten application domains (e.g.,  physics, biology, and economics) and eight modeling task types (e.g., decision-making, prediction, and evaluation). Each sample includes rich contextual components (e.g., textual descriptions, task goals, dataset information, and variable definitions) and requires agents to conduct problem interpretation, model formulation, and numerical reasoning in an integrated, end-to-end fashion.   A detailed breakdown of task types and domain distribution is provided in Appendix~B.

To address this task, we propose MM-Agent, an end-to-end solution for open-ended real-world modeling problems. Inspired by expert workflows, MM-Agent systematically analyzes unstructured problem descriptions, formulates structured mathematical models, derives solutions, and generates analytical reports.
Among these stages, the modeling step poses the greatest challenge, as it requires abstracting complex scenarios into mathematically coherent formulations grounded in both problem context and solution feasibility. To address this, we introduce the Hierarchical Mathematical Modeling Library (HMML): a tri-level knowledge hierarchy encompassing domains, subdomains, and method nodes. HMML encodes 98 high-level modeling schemas that enable both problem-aware and solution-aware retrieval of modeling strategies, supporting abstraction and method selection.  Specifically, MM-Agent first analyzes the problem and decomposes it into subtasks. It then retrieves suitable methods from HMML and refines its modeling plans via an actor-critic mechanism. To solve the models, the agent autonomously generates and iteratively improves code using the MLE-Solver for efficient, accurate execution. Finally, it compiles a structured report summarizing the modeling approach, experimental results, and key insights.

Our contribution can be summarized as follows:
(1) We develop \textbf{MM-Bench}, a benchmark comprising 111 real-world mathematical modeling problems across 8 problem types and 10 domains, designed to evaluate the mathematical modeling capabilities of LLM agents. This benchmark has been carefully created based on real-world competitions. (2) To enhance the mathematical modeling capabilities of LLM agents, we construct the \textbf{HMML}, a three-tiered structure that organizes and retrieves modeling methods through broad domains (e.g., optimization, simulation), specific subdomains (e.g., linear programming, Monte Carlo simulations), and method nodes representing techniques, core ideas, and applications, enabling precise task-specific retrieval.  (3) We introduce \textbf{MM-Agent},  an autonomous agent framework to create mathematical representations of real-world scenarios for making predictions or providing insights. (4) We conduct comprehensive experiments on the proposed benchmark and demonstrate that MM-Agent effectively solves mathematical modeling tasks, outperforming baseline approaches, with an average cost of \$0.88 and \$0.56 per task on GPT-4o~\cite{gpt-4o} and DeepSeek R1~\cite{DBLP:journals/corr/abs-2501-12948}, respectively, and achieving \textbf{an 11.88\% gain over human expert solutions}. Furthermore, following official MCM/ICM protocols, MM-Agent helped two undergraduate teams win the Finalist Award (\textbf{top 2.0\% out of 27,456 teams}) in MCM/ICM 2025.

\vspace{-0.15in}
\section{Related Works}
\vspace{-0.15in}

\textbf{LLM Agents.} Recent advances have led to the development of LLM-based agents that incorporate structured planning, reasoning, and interaction capabilities. By leveraging mechanisms such as memory augmentation, reflective reasoning, and tool usage, these agents enhance task decomposition, iterative refinement, and adaptive problem-solving~\cite{DBLP:conf/iclr/YaoZYDSN023, DBLP:conf/iclr/ChenSZ0YCYLHQQC24, DBLP:conf/nips/LiHIKG23, DBLP:conf/acl/QianLLCDL0CSCXL24, DBLP:journals/corr/abs-2308-08155}. As a result, LLM-based agents have been successfully applied in diverse areas, including software engineering~\cite{DBLP:conf/iclr/JimenezYWYPPN24, DBLP:journals/corr/abs-2407-16741, DBLP:conf/nips/YangJWLYNP24, DBLP:journals/corr/abs-2409-02977}, game playing~\cite{DBLP:journals/corr/abs-2411-00114, DBLP:conf/nips/FengLWTYSM0W23, DBLP:journals/tmlr/WangX0MXZFA24, DBLP:conf/acl/DongZPZ024, DBLP:conf/icml/Xu000W24, DBLP:journals/corr/abs-2404-02039}, human interaction modeling~\cite{DBLP:conf/uist/ParkOCMLB23, DBLP:journals/corr/abs-2411-10109}, cybersecurity~\cite{DBLP:journals/corr/abs-2409-16165, DBLP:journals/corr/abs-2402-06664, DBLP:journals/corr/abs-2408-01605}, robotics~\cite{DBLP:conf/corl/IchterBCFHHHIIJ22, DBLP:journals/corr/abs-2410-24164, DBLP:conf/rss/BrohanBCCDFGHHH23, DBLP:conf/corl/ZitkovichYXXXXW23, DBLP:conf/corl/KimZSDKFK24}, data science~\cite{DS-Agent, DBLP:journals/corr/abs-2402-18679, DBLP:conf/nips/CaoLWCFGXZHMXXZ24, DBLP:journals/corr/abs-2305-13657}, medical diagnosis~\cite{DBLP:journals/corr/abs-2312-00164, DBLP:journals/corr/abs-2405-07960, DBLP:journals/corr/abs-2401-05654, qiu2024llm}, web automation~\cite{DBLP:conf/nips/DengGZCSWSS23, DBLP:conf/iclr/GurFHSMEF24, DBLP:conf/acl/HeYM0D0L024, DBLP:journals/corr/abs-2408-07199}, and scientific research~\cite{yamada2025ai, agent_lab, DBLP:journals/corr/abs-2408-06292, researchagent, DBLP:journals/corr/abs-2502-14499}.

\textbf{LLMs for Autonomous Research.} Automated scientific workflows have enabled the integration of LLMs across various research stages, such as literature review, idea generation, experimental design, and scientific writing. Some studies focus on general research tasks~\cite{researchagent}, while others explore specific domains like AI~\cite{yamada2025ai, agent_lab, AI_ResearchAgent, DBLP:journals/corr/abs-2408-06292}, biomedical discovery~\cite{gao2024empowering}, chemical experiments~\cite{darvish2025organa}, and traffic research~\cite{DBLP:journals/corr/abs-2409-16876}.
For instance, Agent Laboratory~\cite{agent_lab} is an autonomous LLM-based framework designed to expedite AI research by managing key stages such as literature review, experimentation, and report generation. Similarly, ORGANA~\cite{darvish2025organa} is a robotic system that automates chemical experiments by integrating decision-making and perception tools. It collaborates with chemists via LLMs to define objectives and generate detailed experiment logs.

\textbf{LLMs for Mathematics.}  LLMs has advanced mathematical problem-solving through curated datasets and improved reasoning strategies~\cite{yang2025optibench}. The  math~\cite{DBLP:conf/nips/HendrycksBKABTS21,DBLP:journals/corr/abs-2110-14168}  benchmarks have become key resources for evaluating mathematical competence, especially when paired with prompting techniques such as COT~\cite{DBLP:conf/nips/Wei0SBIXCLZ22}. In formal mathematics, LLMs have been fine-tuned on theorem-proving datasets like MiniF2F~\cite{DBLP:conf/iclr/ZhengHP22} and integrated with proof assistants such as Lean~\cite{DBLP:conf/iclr/HanRWAP22} and Coq~\cite{DBLP:conf/icml/YangD19}. Program-aided reasoning further enhances LLM performance by allowing models to generate and execute code for verification~\cite{DBLP:journals/tmlr/ChenM0C23}. While LLMs perform well on well-defined mathematical tasks~\cite{cobbe2021training, XiaoZWXWHFZZS024, AhmadiTeshniziG24, ramamonjison2023nl4opt} with clear symbolic goals, mathematical modeling remains an open-ended challenge that requires translating real-world scenarios into formal representations, often without a single correct solution.

To the best of our knowledge, MM-agent is the first work to explore the application of LLMs to real-world mathematical modeling problems. To facilitate autonomous mathematical modeling, we develop an automated pipeline encompassing problem analysis, mathematical modeling, computational solving, and solution reporting.

\vspace{-0.2in}
\section{Building LLM Agent for Real-World Mathematical Modeling Problems}
\vspace{-0.1in}
Section~\ref{sec:MM-bench} introduces the task of real-world mathematical modeling and presents the construction of MM-Bench, \emph{the first benchmark designed to enable systematic evaluation of LLM-based modeling agents on open-ended tasks}. To further support model construction, we also introduce the Hierarchical Mathematical Modeling Library (HMML), which encodes a tri-level knowledge hierarchy spanning domains, subdomains, and method nodes to facilitate structured method selection and abstraction in Section~\ref{sec:HMML}. In Section~\ref{sec:MM-Agent}, we present MM-Agent, an expert-inspired framework that decomposes the modeling process into four key stages: open-ended problem analysis, structured model formulation, computational problem solving, and report generation.

\vspace{-0.15in}
\subsection{MM-Bench}~\label{sec:MM-bench}
\vspace{-0.2in}

\textbf{Benchmark Construction.}  Real-world mathematical modeling competitions, such as MCM/ICM, challenge undergraduate students worldwide to transform complex real-world phenomena (e.g., risk management, biological dynamics) into mathematical frameworks for prediction, optimization, and decision-making\cite{bender2000introduction, meerschaert2013mathematical}. Drawing participation from a large and diverse pool of teams across many countries and regions~\cite{comap}, these prestigious contests require participants to collaboratively interpret open-ended problems, conduct in-depth analyses, and develop comprehensive solutions~\cite{comap_rule}. These contests offer a natural benchmark for evaluating the problem-solving capabilities of LLM agents in complex, real-world scenarios. We collect all competition problems from the MCM and ICM contests held from 2000 to 2025~\footnote{\url{https://www.contest.comap.com/undergraduate/contests/mcm/previous-contests.php}}, which include both the problem descriptions and associated attachments, such as datasets. We then use GPT-4o to extract the following elements from the competition problems: the \textit{background information} describing the context of the problem, the \textit{problem requirements} outlining the tasks to be completed, the \textit{dataset path} indicating the location of the dataset, the \textit{dataset description} providing details about the dataset, and the \textit{variable description} explaining the attributes within the dataset. For policy-oriented or decision-focused tasks, datasets may not be provided, as these problems typically emphasize qualitative reasoning or scenario-based analysis. Finally, we manually review and correct errors in the extracted information, resulting in the creation of the Mathematical Modeling Benchmark, named MM-Bench. The resulting MM-Bench consists of 10 domains, 8 task types (\eg decision-making, prediction, evaluation \etal ), and a total of 111 problem samples. Detailed statistical information is provided in Section~\ref{sec:appendix_data} of the Appendix.

\textbf{Task Formation.}
MM-Bench evaluates the performance of agents on real-world mathematical modeling problems, which involve translating real-world phenomena into simplified mathematical forms to analyze, interpret, and predict system behavior and outcomes. Given a mathematical problem $\mathcal{F}$, the agent accesses all relevant content $\mathbf{f} \in \mathcal{F}$ (e.g., background information, problem requirements, dataset path, dataset description, and variable description) to generate a final solution report.

\textbf{Evaluation.} Since real-world mathematical modeling problems are open-ended and often lack standard solutions, we reference official modeling evaluation criteria to assess agent performance~\cite{comap_rule}. Specifically, we evaluate the final solution report along four key dimensions: (1) \textit{Analysis Evaluation.} Examines problem definition clarity, identification of key components, and the logical coherence between sub-tasks and overarching objectives.
(2) \textit{Modeling
Rigorousness.} Focuses on rigor and rationality, evaluating whether the assumptions are clearly stated and justified, and whether the chosen methods, metrics, and model structure accurately and scientifically represent the real-world problem.
(3) \textit{Practicality and Scientificity.} Evaluates the practicality and scientific validity of the model, ensuring that it is realistically applicable, provides valuable insights for decision-making, and adheres to scientific principles. This stage also verifies whether the model is theoretically sound and considers all relevant scientific factors to ensure its validity. (4) \textit{Result and Bias Analysis.}  Measures the clarity, interpretability, and analytical depth of results, with attention to identifying and mitigating data or modeling biases to ensure robustness and transparency. We conduct both LLM-based and expert-human evaluations to ensure a comprehensive and reliable assessment. For further details, please refer to Section~\ref{sec:appendix_exp_setup} in the Appendix.

\vspace{-0.2in}
\subsection{Hierarchical  Mathematical Modeling Library Construction}~\label{sec:HMML}
\vspace{-0.2in}

\begin{wrapfigure}{r}{0.6\textwidth}
    \vspace{-10pt}
    \centering
    
    \includegraphics[width=0.6\textwidth]{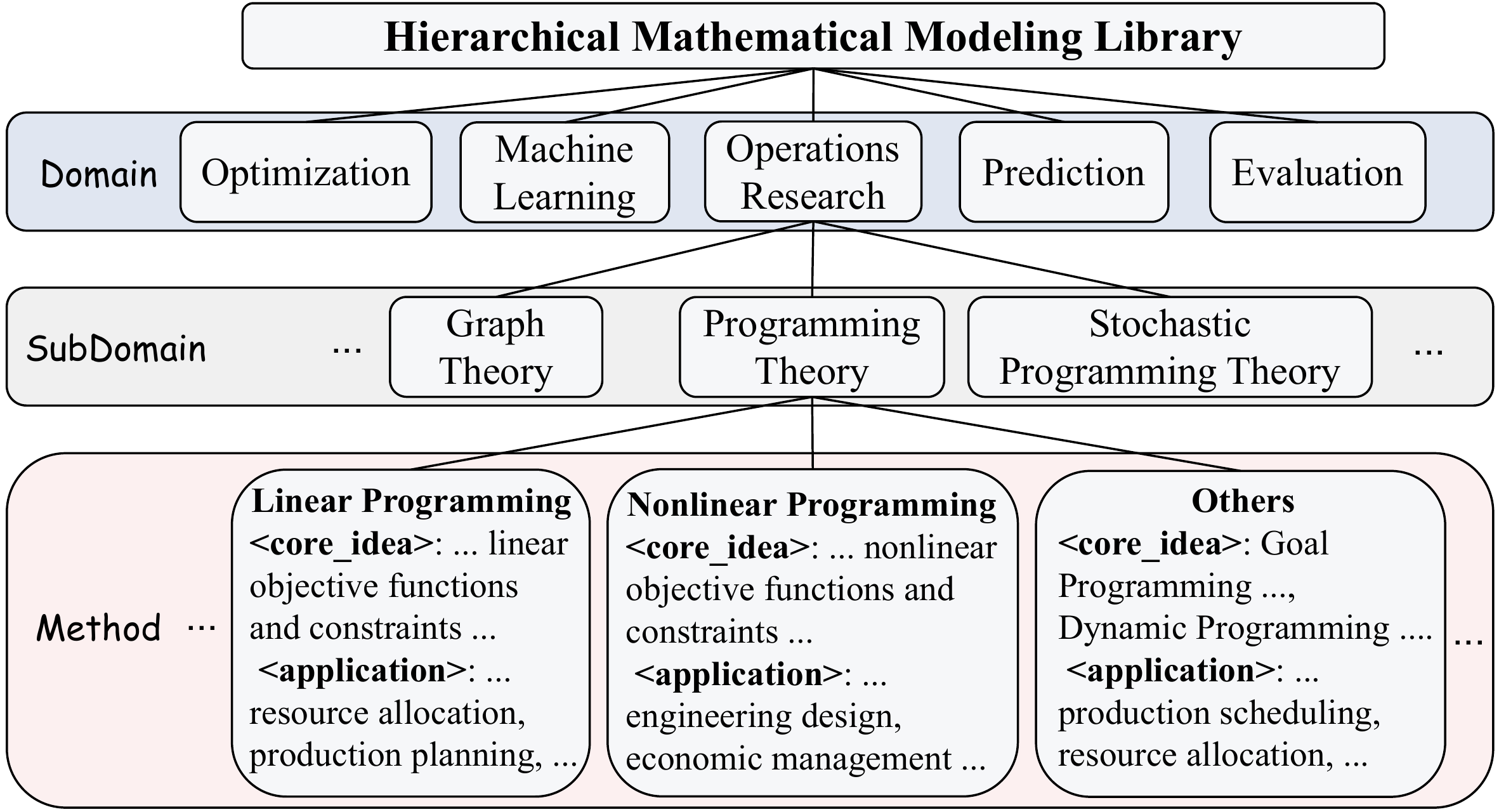}
    \caption{
        The structure of HMML is organized in three levels: modeling domains, subdomains, and method nodes.
    }
    \label{fig:HMML}
    \vspace{-10pt}
\end{wrapfigure}

To enhance the mathematical modeling capabilities of LLM agents, we construct the Hierarchical Mathematical Modeling Library (HMML), a three-tiered structure that organizes and retrieves modeling methods through broad domains (e.g., optimization, simulation), specific subdomains (e.g., linear programming, Monte Carlo simulations), and method nodes representing techniques, core ideas, and applications, enabling precise task-specific retrieval. As shown in Figure~\ref{fig:HMML}, HMML adopts a tree structure. At the top level, the library consists of modeling domains
$
\mathcal{T} = \{ \mathcal{T}^{(1)}, \mathcal{T}^{(2)}, \cdots, \mathcal{T}^{(n)} \}
$.
Each modeling domain $\mathcal{T}^{(i)}$ is subdivided into multiple subdomains:
$
\mathcal{T}^{(i)} = \{ \mathcal{T}^{(i,1)}, \mathcal{T}^{(i,2)}, \cdots, \mathcal{T}^{(i,k)} \}
$.
Within each subdomain $\mathcal{T}^{(i,j)}$, specific method nodes $\mathcal{N}^{(i,j,l)}$ are represented as tuples:
$
\mathcal{N}^{(i,j,l)} = \{\text{modeling method}, \text{core idea}, \text{application}\}
$.
Here, the modeling method provides a high-level overview, the core idea explains the underlying principles, and the application describes the method's typical use cases (e.g., resource allocation, production scheduling).
For example, in the domain of Operations Research ($\mathcal{T}^{(1)}$), the subdomain Programming Theory ($\mathcal{T}^{(1,1)}$) includes the method node $\mathcal{N}^{(1,1,1)}$, which involves Linear Programming. The core idea is optimizing linear objectives with constraints, applied to problems like production resource scheduling.
The HMML includes modeling domains such as Operations Research, Optimization, Machine Learning, Prediction, and Evaluation, with 17 subdomains and about 98 modeling methods, such as Linear Programming, Ant Colony Optimization, Expectation Maximization, Analytic Hierarchy Process, and Kolmogorov-Smirnov Test. For further details, please refer to Section~\ref{sec:HMML_appendix} in the Appendix.

\subsection{MM-Agent}~\label{sec:MM-Agent}
\vspace{-0.2in}


\begin{figure}[!t]
    \centering
    \includegraphics[width=\linewidth]{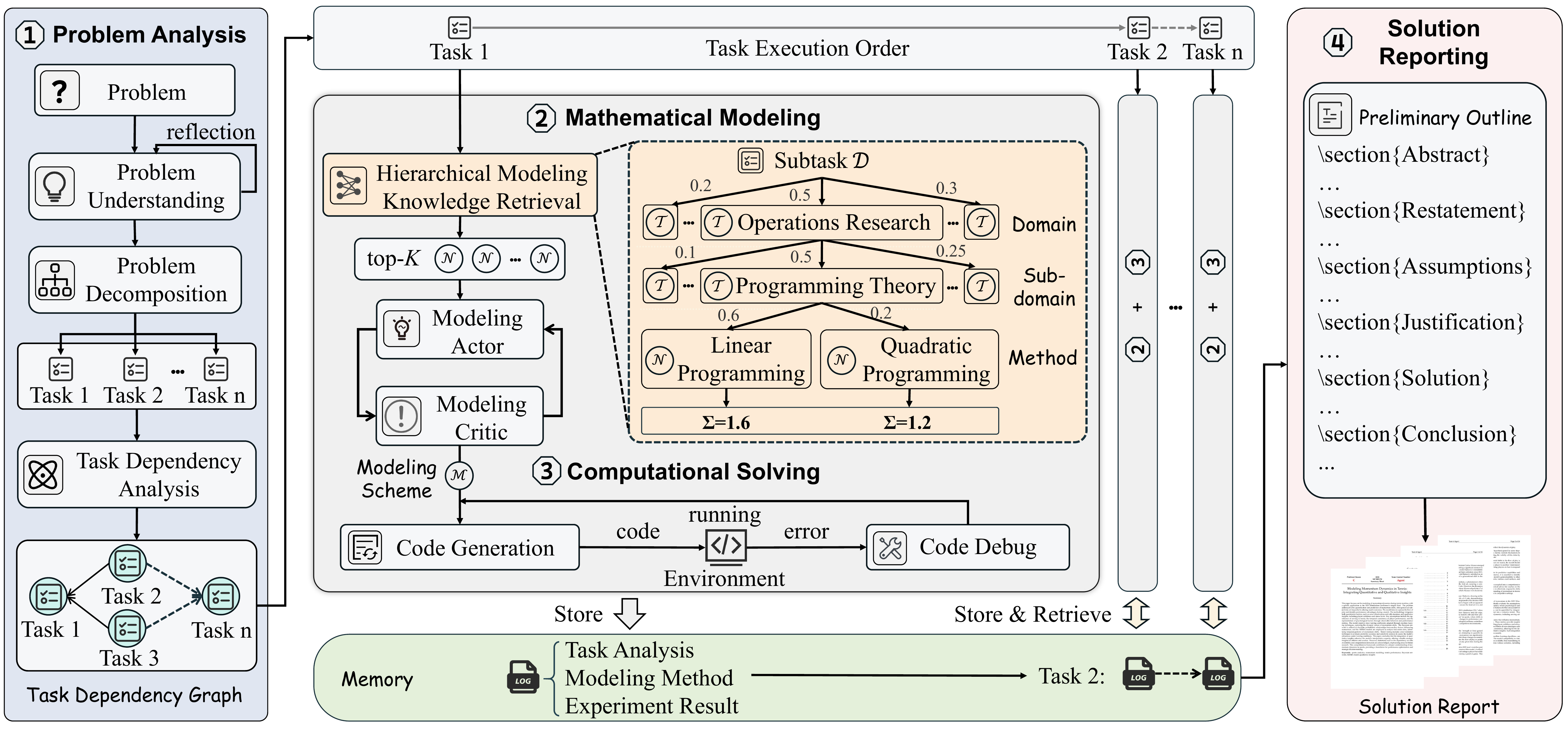}
    \caption{Overview of the MM-Agent framework. The workflow consists of four sequential phases: Problem Analysis, Mathematical Modeling, Computational Solving, and Solution Reporting. In the Problem Analysis phase, MM-Agent decomposes the input problem into structured subtasks. In Mathematical Modeling, it constructs formal mathematical representations for each subtask. During Computational Solving, MM-Agent applies appropriate computational methods to derive solutions. Finally, in Solution Reporting, it synthesizes the results into a comprehensive report, clearly summarizing the solutions and associated insights.}
    \label{fig:overall_framework}
\vspace{-0.1in}
\end{figure}

This section introduces MM-Agent, an LLM-based multi-agent system designed to automate mathematical modeling tasks. Its workflow consists of four key phases: Problem Analysis, Mathematical Modeling, Computational Solving, and Solution Reporting. MM-Agent begins by analyzing the given problem and breaking it into subtasks. It then constructs formal mathematical models for each subtask, conducts experiments, and generates a solution. Finally, MM-Agent produces a comprehensive report summarizing the solution and results. The overall framework is illustrated in Figure~\ref{fig:overall_framework}.

\subsubsection{Problem Analysis}
This section presents the problem analysis phase of MM-Agent, which transforms complex real-world problems into mathematical modeling tasks. The process involves abstracting key elements (e.g., background, requirements) and analyzing relationships (e.g., variable dependencies) to identify suitable modeling methods. The problem analysis phase consists of three steps: \textit{problem understanding}, \textit{problem decomposition}, and \textit{task dependency analysis}.

\textbf{Problem Undersanding.} Given a mathematical modeling problem $\mathcal{F}$, we consider an LLM $\mathbf{y} = \pi_{\theta}(\mathbf{x}; \mathbf{x}_I)$, where $\pi_{\theta}(\cdot)$ denotes a language model parameterized by $\theta$, which autoregressively generates output tokens $\mathbf{y}$ from an input sequence $\mathbf{x}$ under the guidance of an instruction prompt $\mathbf{x}_I$. The prompt $\mathbf{x}_I$ encodes task-relevant context such as background information, problem requirements, and dataset descriptions. Conditioned on this input, the analyst agent performs a structured analysis to identify the problem type, core concepts, assumptions, objectives, and other essential factors. Specifically, this process is represented as  $\mathcal{U} _p= \pi_{\theta}(\mathcal{F} ; \mathbf{x}_{u})$, where $\mathbf{x}_{u}$ represents the profile prompt used for problem undersanding, and $\mathcal{U} _p$ is the analysis result. To deepen the understanding of the problem, the analyst agent adopts self-reflection to iteratively refine its analysis.

\textbf{Problem Decomposition.}  After understanding the problem, the coordinator agent decomposes it into a set of subtasks to address its multiple objectives. This process is represented as $\textbf{D} = \pi_{\theta}( \mathcal{F} ,\mathcal{U} _p; \mathbf{x}_{d} ) $, where $\mathbf{x}_{d}$ represents the profile prompt used for problem decomposition, $ \textbf{D} = \{\mathcal{D} _1,\mathcal{D} _2,\cdots ,\mathcal{D} _n\}$ denotes the set of subtasks, and each $\mathcal{D}_i$ corresponds to an individual subtask. Each subtask is associated with a specific objective or component of the problem. As illustrated in~\figurename~\ref{fig:problem_analysis} in Appendix~\ref{sec:case_study},  for the problem of predicting momentum in tennis matches, the agent decomposes the problem into four key subtasks: Momentum Quantification, Differentiation between Momentum and Randomness, Momentum Prediction, and Model Generalization.  

\textbf{Task Dependency Analysis.}  Since individual tasks are not independent, dependencies exist among them (\eg a model prediction task may rely on the analysis results from a data analysis task). To capture these dependencies and optimally address problem requirements, the task coordinator agent first conducts a comprehensive analysis to identify interdependencies among tasks. This process is formulated as $\mathbf{U} = \pi_{\theta} ( \mathbf{D} ; \textbf{x}_{t}) $, where $\mathbf{U}= \{u_1, \cdots, u_n\}$ represents the task-specific dependency analysis with $u_i$ denoting the detailed analysis of task $\mathcal{D}_i$. The instruction prompt $\mathbf{x}_{t}$ directs the LLM to analyze dependencies among tasks. Subsequently, the task coordinator agent further leverages task analysis results to construct sequential subtasks $\{ \mathcal{D}_1, \mathcal{D}_2,\cdots, \mathcal{D}_n\}$ with depenceny graph represented by $\mathcal{G} =\left( \mathcal{V} ,\mathcal{E} \right)$, where $\mathcal{V} =\left\{ \mathcal{D} _1,\mathcal{D} _2,\cdots ,\mathcal{D} _n \right\} $ represents the set of nodes (tasks), and $\mathcal{E} =\left\{ \left( \mathcal{D} _i,\mathcal{D} _j \right) |\mathcal{D} _i,\mathcal{D} _j\in \mathcal{V} \right\} $ denotes the directed edges indicating task dependencies. The sequential subtasks are executed in order, with the outcomes of historical modeling processes stored in memory modules represented as $\mathcal{H} = \{(\mathcal{D}_1, \mathcal{Q}_1), \cdots, (\mathcal{D}_n, \mathcal{Q}_n)\}$, where $\mathcal{Q}_i = \{\mathcal{M}_i, \mathcal{C}_i, \mathcal{O}_i\}$ denotes the intermediate outputs of subtask $\mathcal{D}_i$. Specifically, $\mathcal{Q}_i$ contains the mathematical modeling scheme $\mathcal{M}_i$, computational code $\mathcal{C}_i$, and experimental results $\mathcal{O}_i$. The coordination agent leverages a dependency graph to manage relationships among subtasks and utilizes the memory module to facilitate information transfer and communication between tasks.

\vspace{-0.0in}

\subsubsection{Mathematical Modeling}
\vspace{-0.0in}
To efficiently automate solving  mathematical modeling, we propose the Hierarchical Actor-Critic Modeling Optimization. The specific mathematical modeling process for each subtask $\mathcal{D}_i$ involves hierarchical method retrieval from the HMML, followed by actor-critic iterative optimization.


\textbf{Hierarchical Modeling Knowledge Retrieval.} Given a subtask $\mathcal{D}_{i}$ and a hierarchical modeling library $\mathcal{T}$, a Depth-First Search (DFS) traversal is initiated from the root node $\mathcal{T}^{(i)}$, to compute the similarity between the subtask and modeling methods. The similarity measure is defined as $\text{Sim}(\mathcal{D}, \mathcal{N}) = \frac{ \mathbf{e}_{\mathcal{D}} \cdot \mathbf{e}_{\mathcal{N}}}{  \left \| \mathbf{e}_{\mathcal{D}} \right \|  \cdot  \left \| \mathbf{e}_{\mathcal{N}} \right \| }$, where $\mathbf{e}_{\mathcal{D}}$ and $\mathbf{e}_{\mathcal{N}}$ represent embeddings of  the subtask $\mathcal{D}$ and the mathematical modeling method node $\mathcal{N}$, respectively. In practice, we adopt the embedding model mGTE~\cite{zhang2024mgte} to generate these embeddings. After traversing the entire hierarchical tree of methods, each method node's final score is updated by combining its own similarity with the similarity of its parent node, computed as $S(\mathcal{D}, \mathcal{N}) = \omega \cdot \text{Sim}(\mathcal{D}, \mathcal{N}) + (1-\omega) \cdot \text{Sim}(\mathcal{D}, \mathcal{N}_{\text{parent}})$, where $S$ denotes the scoring function, $\omega$ is a hyperparameter, and $\mathcal{N}_{\text{parent}}$ represents the parent (subdomain) of method node $\mathcal{N}$. Finally, the top-$K$ method nodes with the highest scores, denoted as $\mathcal{N}_{\text{top-}K} = \{\mathcal{N}_{(1)}, \cdots, \mathcal{N}_{(K)}\}$, are selected and returned to the modeling agent.

\textbf{Actor-Critic Iterative Modeling Optimization.} While retrieved mathematical modeling knowledge offers foundational methods and ideas, it often lacks the depth needed to address specific problem nuances (e.g., dealing with nonlinear constraints, optimizing multiple conflicting objectives, etc.). To overcome these limitations, we introduce an actor-critic iterative optimization framework that progressively refines the modeling scheme, enabling it to effectively manage complex constraints and enhance overall solution quality. 

Given a problem $\mathcal{D}_i$, the task coordinator agent retrieves the relevant dependent resource $\mathcal{R}_i$ from memory modules $\mathcal{H}$ based on the task dependency graph $\mathcal{G}$. Using the retrieved resource $\mathcal{R}_i$ and the method set $\mathcal{N}_{\text{top-}K} $ obtained from the retrieval step, the actor modeling agent generates an initial modeling scheme: $\mathcal{M}_i^{(0)} = \pi_{\theta}(\mathcal{D}_i, \mathcal{N}_{\text{top-}K} , \mathcal{R}_i; \mathbf{x}_{a})$, where $\mathbf{x}_{a}$ is the modeling prompt.
Subsequently, the critic agent evaluates the quality of the current modeling scheme $\mathcal{M}_i^{(t)}$ and provides targeted feedback: $\mathcal{F}_i^{(t)} = \pi_{\theta}(\mathcal{D}_i, \mathcal{M}_i^{(t)}, \mathcal{R}_i; \mathbf{x}_{c})$, where $\mathbf{x}_{c}$ is the critic feedback prompt. Upon receiving feedback $\mathcal{F}_i^{(t)}$, the actor modeling agent refines the scheme by integrating the critic's suggestions and corrections by $\mathcal{M}_i^{(t+1)} = \pi_{\theta}(\mathcal{M}_i^{(t)}, \mathcal{F}_i^{(t)};\mathbf{x}_{r})$, where $\mathbf{x}_r$ is the mathematical modeling refine prompt. This iterative procedure continues until the maximum number of iterations $n_{r}$ is reached.

\vspace{-0.0cm}
\subsubsection{Computational Solving and Solution Reporting}
This section describes the computational solving and solution reporting phase of MM-Agent, which focuses on solving the mathematical model and generating a comprehensive solution report. The agent autonomously writes code to conduct computational experiments using the MLE-Solver~\cite{chan2025mlebench}, which iteratively generates, tests, and refines code to ensure efficient and accurate execution. Upon completion of the experiments, the agent formulates a structured solution report, summarizing the modeling approach, experimental results, and key findings.

\textbf{Code Generation and Execution.} Given the mathematical modeling scheme $\mathcal{M}_i$, the modeling programmer agent generates the corresponding code as follows: $\mathcal{C}_i = \pi_{\theta}( \mathcal{D}_i, \mathcal{M}_i; \mathbf{x}_g )$, where $\mathbf{x}_g$ represents the instruction prompt used to direct the LLM to generate the computational code, and $\mathcal{C}_i$ denotes the mathematical modeling code for task $\mathcal{D}_i$. After code generation, the program is compiled to check for runtime errors. If it compiles successfully, the experimental results $\mathcal{O}_i$ are returned. If the code fails to compile, the agent attempts to repair it over $n_c$ iterations by analyzing the last error message and making the necessary corrections. Upon task completion, the task coordinator agent updates the agent’s memory: $\mathcal{H} \gets  \mathcal{H} \cup \{\mathcal{D} _i,\mathcal{Q} _i \}$. In practice, for policy-related modeling problems, where the goal is to provide insights and recommendations based on existing knowledge or models, the modeling agent directly offers these insights without generating code.

\textbf{Preliminary Report Outline.}
 After all tasks have been completed, reporting agent compiles a comprehensive summary of the problem-solving process. The first step  is to construct a structured outline for the mathematical modeling report. This outline establishes the framework of the report, organizing it into eight key sections: abstract, problem restatement, model assumptions, justification of assumptions, notation and definitions, problem analysis, solution, and conclusion. To ensure clarity and coherence, the outline integrates proper LaTeX formatting, facilitating seamless compilation and further refinements. By structuring the content systematically, it provides a solid foundation for an in-depth and well-organized final report.

\textbf{Solution Report.} Once the outline is established, the reporting agent employs specialized commands to progressively refine the report, drawing on the task coordinator agent’s memory $\mathcal{H}$. Prior to incorporating any revisions, the system compiles the LaTeX code to ensure that it functions correctly, preserving the integrity of the document. Through a series of iterative edits, the agent guarantees that the report meets the necessary standards for quality, coherence, and academic rigor.

\vspace{-0.0cm}
\section{Experiments}
\vspace{-0.0cm}
\subsection{Experimental Setup}
\vspace{-0.0cm}
\textbf{Baselines.} We evaluate MM-Agent against both human-authored solutions and SOTA LLM agents. As no prior work directly targets mathematical modeling problems, we repurpose existing autonomous research agents for comparison. The baselines include: (1) Human Team: Award-winning solutions (Honorable Mention or above) from real-world modeling competitions, serving as a strong human benchmark; (2) DS-Agent~\cite{DS-Agent}: An LLM agent for automated data science, adapted with its core case-based reasoning framework for modeling tasks; (3) ResearchAgent~\cite{AI_ResearchAgent}: Originally designed to automate experimentation loops for machine learning tasks, adapted with its core framework for modeling problems; and (4) Agent Laboratory~\cite{agent_lab}: A scientific discovery framework that guides agents through literature review, experimentation, and report writing. We extend it to search arXiv for relevant modeling methods and assemble them into problem-solving pipelines.

\textbf{Experimental Implementation.} We select a subset of mathematical modeling problems from the past five years (2021–2025) as our test set, ensuring diversity across problem types and domains to support a representative evaluation. This subset consists of 32 problems in total. To mitigate potential data leakage from LLM pretraining, we evaluate problems from 2021–2024 separately from those in 2025.
The LLM agents used in this evaluation include GPT-4o and Deepseek-R1 as base models. For the evaluation, we adopt both GPT-4o-based automatic scoring and human expert review, using a unified 1-to-10 scale. The selected human experts have previously earned at least an Honorable Mention in mathematical modeling competitions. Additional experimental details are provided in Appendix~\ref{sec:appendix_exp_setup}. To evaluate annotation quality, we measure inter-annotator agreement, including both human–human and model–human agreements, as detailed in Appendix~\ref{sec:annotation_quality}.

\begin{table}[t]
\vspace{-0.0in}
\caption{Experimental results on the 2021–2024 and 2025 mathematical modeling competitions. AE, MR, PS, and RBA denote \textit{Analysis Evaluation}, \textit{Modeling Rigorousness}, \textit{Practicality and Scientificity}, and \textit{Result and Bias Analysis}, respectively.}~\label{tab:main_exps}
\centering
\resizebox{\textwidth}{!}{
\begin{tabular}{l|ccccc|ccccc}
\hline
\multirow{2}{*}{\textbf{Methods}} & \multicolumn{5}{c|}{\textbf{2021–2024}} & \multicolumn{5}{c}{\textbf{2025}} \\
\cline{2-11}
 & \textbf{AE $\uparrow$} & \textbf{MR $\uparrow$} & \textbf{PS $\uparrow$} & \textbf{RBA $\uparrow$} & \textbf{Overall $\uparrow$}
 & \textbf{AE $\uparrow$} & \textbf{MR $\uparrow$} & \textbf{PS $\uparrow$} & \textbf{RBA $\uparrow$} & \textbf{Overall $\uparrow$} \\
\hline
\multicolumn{11}{c}{\textbf{Human}} \\
\hline
\rowcolor{gray!15} Human Team & 9.04 & 6.20 & 8.79 & 7.62 & 7.91 & 9.25 & 7.42 & 8.92 & 6.50 & 8.02 \\
\hline
\multicolumn{11}{c}{\textbf{GPT-4o}} \\
\hline
GPT-4o & 7.62 & 3.86 & 8.48 & 5.17 & 6.28 & 7.67 & 3.67 & 8.90 & 5.75 & 6.50 \\
DS-Agent & 8.18 & 7.08 & 8.72 & 7.47 & 7.86 & 8.25 & \textbf{7.33} & 8.92 & 7.10 & 7.90 \\
ResearchAgent & 7.97 & 6.80 & 8.82 & 7.37 & 7.74 & 8.00 & 7.30 & 8.60 & 7.00 & 7.73 \\
Agent Laboratory & 8.56 & 6.35 & 8.63 & 5.56 & 7.28 & 8.75 & 5.58 & 8.58 & 5.33 & 7.13 \\
\textbf{MM-Agent} & \textbf{9.15} & \textbf{7.28} & \textbf{9.00} & \textbf{8.44} & \textbf{8.85} & \textbf{8.86} & 7.21 & \textbf{9.00} & \textbf{8.43} & \textbf{8.38} \\
\hline
\multicolumn{11}{c}{\textbf{DeepSeek-R1-671B}} \\
\hline
DeepSeek-R1 & 7.23 & 4.79 & 8.69 & 4.50 & 6.30 & 7.42 & 4.25 & 8.50 & 5.25 & 6.35 \\
DS-Agent & 8.25 & 6.88 & 8.74 & 7.19 & 7.77 & 7.92 & 6.33 & 9.00 & 7.60 & 7.71 \\
ResearchAgent & 8.13 & 7.04 & 8.77 & 6.92 & 7.72 & 8.00 & 6.75 & 8.83 & 7.58 & 7.79 \\
Agent Laboratory & 8.65 & 5.96 & 8.70 & 5.91 & 7.31 & 8.83 & 5.50 & 8.83 & 5.58 & 7.19 \\
\textbf{MM-Agent} & \textbf{9.54} & \textbf{8.25} & \textbf{9.06} & \textbf{8.54} & \textbf{8.85} & \textbf{9.50} & \textbf{8.33} & \textbf{9.25} & \textbf{8.58} & \textbf{8.92} \\
\hline
\end{tabular}
}
\end{table}

\vspace{-0.0in}
\subsection{Experimental Results}~\label{sec_exp_results}
\textbf{Main Experiments.}
Table~\ref{tab:main_exps} shows that MM-Agent achieves state-of-the-art (SOTA) performance across all evaluation dimensions.
(1) Directly applying foundational models (GPT-4o or DeepSeek-R1-671B) without agent-level orchestration results in significantly weaker performance, particularly in MR and RBA. This gap underscores the inadequacy of LLMs in handling the open-ended, structured reasoning required for real-world modeling tasks and highlights the necessity of structured agent-based workflows.
(2) MM-Agent consistently outperforms all baseline agents, achieving the highest overall scores under both GPT-4o and DeepSeek-R1-671B backbones.
(3) Agents built on DeepSeek-R1-671B surpass their GPT-4o counterparts, with MM-Agent demonstrating marked gains in Modeling Rigorousness and Result and Bias Analysis, suggesting stronger reasoning capabilities in the larger model.
(4) Human teams remain strong competitors, outperforming all LLM-based agents except MM-Agent on most metrics, underscoring both the complexity of the task and MM-Agent’s near-human modeling proficiency.
(5) The 2025 results closely mirror those from 2021–2024, indicating strong temporal consistency. This robustness mitigates concerns about potential data leakage (e.g., memorized solutions) and further supports the conclusion that MM-Agent performs genuine modeling rather than overfitting.
In addition to benchmark results, we developed a publicly available modeling copilot system~\footnote{\url{https://huggingface.co/spaces/MathematicalModelingAgent/MathematicalModelingAgent}} based on MM-Agent, aligned with official MCM/ICM protocols and LLM usage guidelines. This system assisted two undergraduate teams in securing the Finalist Award (\textbf{top 2.0\% among 27,456 teams}) in the 2025 MCM/ICM competition. This real-world validation illustrates MM-Agent’s practical effectiveness as a modeling copilot, capable of supporting human users in high-stakes, open-ended scientific tasks.

\vspace{-0.0in}
\subsection{Ablation Study and Further Analysis}
To better understand the design and practical utility of MM-Agent, we present a three-part analysis. First, we conduct an ablation study to quantify the impact of each core module on modeling performance. Second, we evaluate token usage, cost, and runtime to assess deployment efficiency. Finally, we test MM-Agent on well-defined, formulated mathematical problems to examine its generalization beyond open-ended modeling. 

\begin{wrapfigure}{r}{0.6\textwidth}
    \vspace{-1pt}
    \centering
    \subfloat[GPT-4o]{%
        \includegraphics[width=0.28\textwidth]{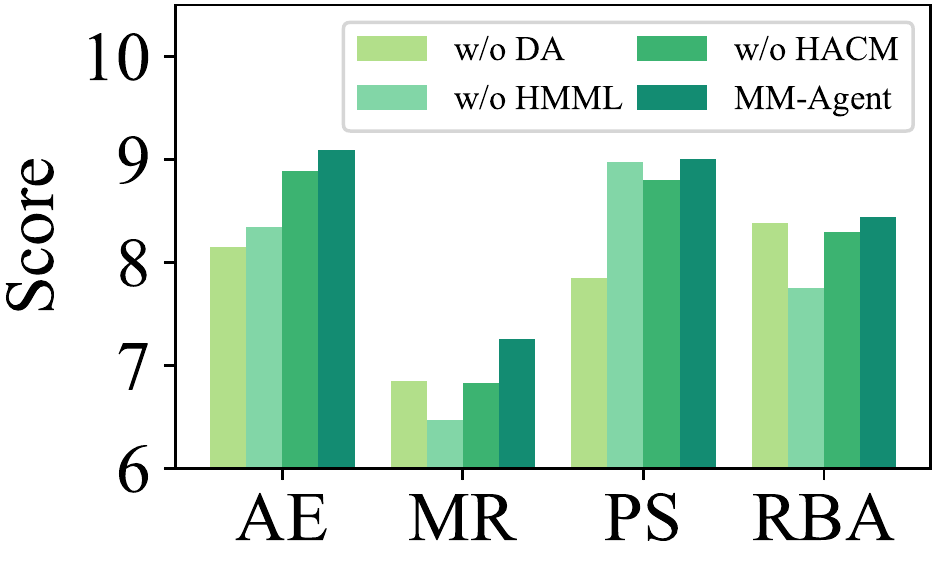}
    }%
    \hspace{0.01\textwidth}
    \subfloat[DeepSeek-R1-671B]{%
        \includegraphics[width=0.28\textwidth]{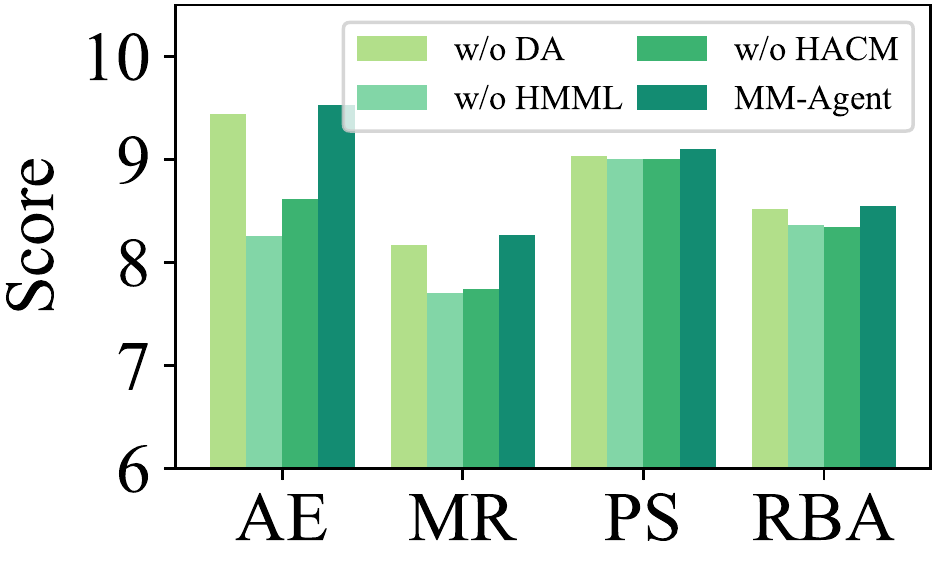}
    }
    \caption{Ablation study of the effect of the problem
analysis and mathematical modeling. }
    \label{fig:ablation_study}
    \vspace{-5pt}
\end{wrapfigure}

\textbf{Contribution of Key Components.}
We perform an ablation study to assess the impact of three core modules in MM-Agent: \textit{task dependency analysis} (DA), the \textit{Hierarchical Mathematical Modeling Library} (HMML), and \textit{hierarchical actor-critic modeling} (HACM). Specifically, we (1) replace DA with a naive task parser (\textbf{w/o DA}), (2) substitute HMML with a flat retrieval library lacking hierarchical structure (\textbf{w/o HMML}), and (3) remove HACM to disable iterative self-refinement (\textbf{w/o HACM}). These variants allow us to evaluate each module's contribution to structured problem understanding and modeling performance. As shown in Figure~\ref{fig:ablation_study}, MM-Agent consistently outperforms all ablated variants across GPT-4o and DeepSeek-R1-671B backbones under four evaluation metrics. Removing DA significantly reduces MR, indicating that deep task comprehension is essential for rigorous formulation. The absence of HACM leads to sharp declines in AE and PS, highlighting its critical role in constructing coherent, scientifically sound models. Notably, removing HMML causes clear drops in AE and RBA, underscoring the importance of structured retrieval in aligning modeling strategies with both problem context and solution needs. Unlike flat libraries that treat all methods equally, HMML encodes 98 high-level modeling schemas organized hierarchically by problem type, abstraction level, and solution paradigm. This enables problem-aware and solution-aware retrieval that better supports abstraction, constraint reasoning, and method selection, key capabilities for effective modeling.

\begin{wraptable}{r}{0.48\textwidth}
\centering
\vspace{-10pt}
\caption{Experimental results on average token consumption, cost, and runtime.}
\resizebox{0.48\textwidth}{!}{
\begin{tabular}{cccc}
\toprule
\multicolumn{1}{c|}{Methods}         & Token  & Cost(\$) & Runtime(s) \\ \midrule
\multicolumn{4}{c}{GPT-4o}                                            \\ \hline
\multicolumn{1}{c|}{DS-Agent}        & 198,186 & 0.77     & 1,044       \\
\multicolumn{1}{c|}{ResearchAgent}   & 170,732 & 0.67     & 459        \\
\multicolumn{1}{c|}{Agent Laboratory} & 746,159 & 2.14     & 1,015       \\
\multicolumn{1}{c|}{MM-Agent}        & 240,877 & 0.88     & 906        \\ \hline
\multicolumn{4}{c}{DeepSeek-R1-671B}                                  \\ \hline
\multicolumn{1}{c|}{DS-Agent}        & 341,432 & 0.46     & 7,035       \\
\multicolumn{1}{c|}{ResearchAgent}   & 222,030 & 0.28     & 4,816       \\
\multicolumn{1}{c|}{Agent Laboratory} & 974,423 & 0.89     & 11,331      \\
\multicolumn{1}{c|}{MM-Agent}        & 530,363 & 0.56     & 7,529       \\  \bottomrule
\end{tabular}}
\vspace{-10pt}
\label{tab:cost}
\end{wraptable}

\textbf{Cost Efficiency Analysis.}
We assess the cost efficiency of MM-Agent in solving real-world mathematical modeling problems, focusing on token usage, monetary cost, and runtime. All evaluations are conducted via official APIs provided by model vendors.
As shown in \tablename~\ref{tab:cost}, MM-Agent matches the performance of DS-Agent and ResearchAgent with comparable computational cost and runtime. Compared to Agent Laboratory, it achieves higher performance while substantially reducing both cost and execution time, highlighting its scalability and practical viability. Further results on additional models and a detailed case study are included in Appendix~\ref{sec:appendix_other_exps}.

\textbf{Experiments on Well-defined Mathematical Optimization Problems.}
To complement MM-Bench's open-ended focus, we evaluate MM-Agent on well-defined mathematical optimization tasks, including both linear and nonlinear programming. In this setting, the agent receives complete problem specifications (e.g., variables, objective function, and constraints ) and directly outputs the numerical solution.
Since these tasks have known ground truth answers, accuracy serves as a direct performance metric. We conduct experiments on the OPTIBENCH dataset~\cite{yang2025optibench}, with detailed results provided in Table~\ref{tab:optibench} (Appendix~\ref{sec:mathopt}). MM-Agent consistently outperforms GPT-4o across all subtasks in a zero-shot setting, demonstrating robust generalization to formulated optimization problems.

\vspace{-0.1in}
\section{Conclusion}
\vspace{-0.1in}

In this work, we introduce MM-Bench, a benchmark for evaluating LLM-based agents in real-world mathematical modeling. By assessing agents across diverse domains and problems, we expose key challenges in bridging real-world phenomena with mathematical formulations. Our findings reveal that existing LLM agents often overlook essential modeling principles, such as abstraction, constraints, and assumptions, resulting in oversimplified and scientifically invalid outputs. To address these issues, we propose MM-Agent, an autonomous pipeline that systematically handles problem analysis, model formulation, solution development, and result interpretation. Comprehensive experiments show that MM-Agent significantly outperforms existing LLM agents, though challenges remain in higher-order reasoning and interdisciplinary problem-solving. We hope our benchmark and framework lay a foundation for future progress in LLM-driven mathematical modeling.

{\small
\bibliographystyle{unsrt}
\bibliography{ref}
}

\newpage
\appendix

\newcommand{\appendixtitle}{%
    \begin{center}
        \vspace*{2cm}
        {\LARGE\bfseries
        MM-Agent: LLM as Agents for Real-world Mathematical Modeling Problem \\ Supplementary Material}
        \vspace{1cm}
    \end{center}
}

\appendixtitle

\tableofcontents

\newpage

\section{Broader impacts}
Mathematical modeling serves as a cornerstone methodology for formulating, analyzing, and solving complex real-world problems, underpinning scientific discovery and technological advancement across applied mathematics, natural sciences, engineering, and the social sciences. By automating this process, our work on LLM-powered mathematical modeling agents (MM-Agents) has the potential to substantially broaden access to high-quality modeling expertise, accelerate research in data-scarce or expert-limited domains, and support decision-making in high-stakes environments such as epidemiology, sustainability, and infrastructure planning.

\textbf{MM-Agents lower the barrier to mathematical modeling for diverse real-world applications.}  MM-Agents can democratize the modeling process by enabling non-experts, such as students, practitioners, or policymakers, to explore complex systems through structured analytical reasoning. This may significantly enhance STEM education, interdisciplinary collaboration, and rapid response in time-sensitive domains like disaster management or urban systems. Moreover, by encoding expert-level workflows and structured domain knowledge, MM-Agents provide a foundation for scalable, reusable modeling across diverse fields, potentially catalyzing scientific discovery in areas with limited modeling resources.

\textbf{Contamination \& Plagiarism.}
Because our dataset includes publicly available mathematical modeling competitions, there is a possibility that LLMs may have previously encountered solution reports (e.g., from websites like Arxiv). This introduces a potential risk of contamination, meaning models might memorize solutions or gain insights that artificially boost their performance on MM-bench beyond actual capabilities. To mitigate this risk, we selected competition problems from the most recent year (2025), ensuring the evaluated LLMs had not been trained on these specific solutions. Our experiments (Section~\ref{sec_exp_results}) detected no systematic contamination effects in GPT-4o or Deepseek-R1. Furthermore, the distribution of results from the 2025 competition aligns consistently with those from previous years. Nevertheless, we cannot guarantee that future models will remain unaffected. To proactively manage potential contamination risks, we recommend regularly updating the MM-bench with new mathematical modeling problems.

\textbf{Judge Bias and Accessibility.}  Although we evaluate annotation quality using inter-annotator agreement metrics, bias still exists in both human and LLM annotators, particularly due to the inherent subjectivity in evaluating different modeling solutions. In future work, we aim to develop a mathematical modeling judge, which could provide more structured, transparent, and consistent evaluations, thereby mitigating annotator bias.
Additionally, running LLM agents on MM-Bench is computationally intensive. In our experiments, GPT-4o and DeepSeek-R1 consumed approximately 0.53 million and 0.24 million tokens, respectively.

\textbf{Misuse.} The MM-agent offers substantial opportunities to streamline real-world mathematical modeling tasks across diverse areas such as engineering optimization, environmental resource management, and epidemic forecasting. By automating complex and labor-intensive processes, this technology enables researchers to focus more effectively on conceptual innovation and experimental design. Nevertheless, the agent's robust automation capabilities also introduce significant ethical concerns. Reduced barriers to entry may inadvertently promote the generation of low-quality or misleading scientific outputs. Furthermore, entirely AI-generated reports could potentially be misused in mathematical modeling competitions. To address these ethical challenges and safeguard academic integrity, it is imperative to transparently disclose any AI assistance involved in competition submissions.

\section{Statistics of MM-Bench}~\label{sec:appendix_data}

\begin{figure*}[ht]
    \centering
    \subfloat[Data Domain]{%
        \includegraphics[width=0.4\linewidth]{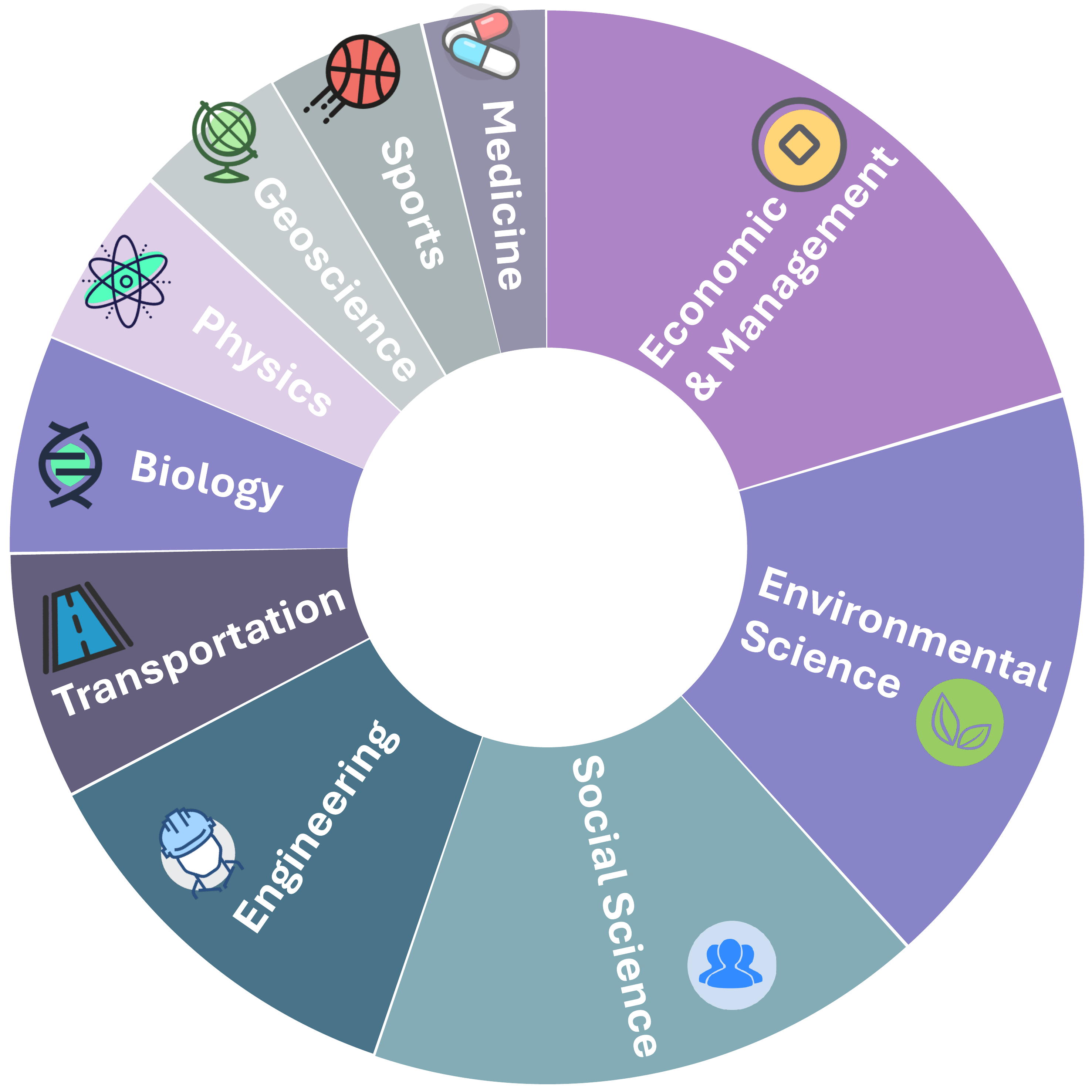}
    }
    \hfill
    \subfloat[Problem Type]{%
        \includegraphics[width=0.4\linewidth]{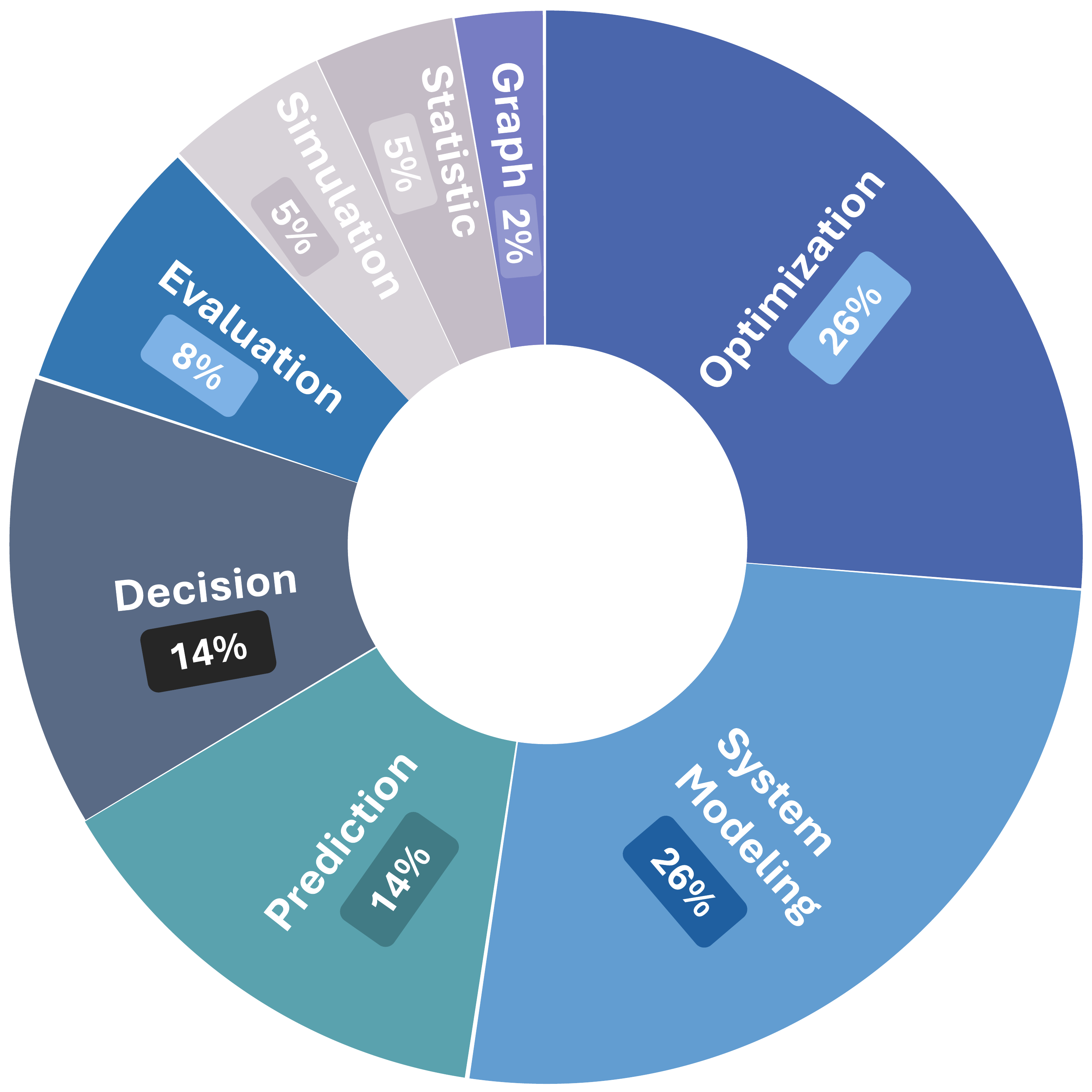}
    }
    \caption{Illustrations of problem domain and types.}
    \label{fig:data_types_domains}
\end{figure*}

MM-Bench consists of 10 domains, 8 task types (\eg decision, prediction, evaluation \etal ), and a total of 111 problem samples, all sourced from undergraduate-level Mathematical Modeling Contests (MCM and ICM). Each sample in MM-Bench is based on a mathematical modeling competition problem and includes \textit{background information} that describes the context of the problem, the \textit{problem requirements} outlining the tasks to be completed, the \textit{dataset path} indicating the location of the dataset, \textit{dataset description} providing details about the dataset, and \textit{variable description} explaining the attributes within the dataset. For policy-oriented or decision-focused tasks, datasets may not be provided, as these problems often emphasize qualitative reasoning or scenario-based analysis. The statistical information can be seen in Figure~\ref{fig:data_types_domains}.

\section{Hierarchical  Mathematical Modeling Library Construction}~\label{sec:HMML_appendix}
To enhance LLM agents' mathematical modeling capabilities, we introduce the Hierarchical Mathematical Modeling Library (HMML), a three-level structured hierarchy designed for efficient, targeted method retrieval. Unlike conventional flat libraries, HMML explicitly captures method heterogeneity by categorizing them into distinct modeling domains (top layer), associated subdomains (middle layer), and specific method nodes (bottom layer). This structured design streamlines retrieval through progressively refined searches guided by high-level reasoning schemas tailored specifically to mathematical modeling tasks. Specifically, HMML adopts a tree structure comprising three abstraction layers, as illustrated in~\figurename~\ref{fig:HMML}. The top layer represents distinct mathematical modeling domains, the second layer corresponds to their respective subdomains, and the third layer includes specific method nodes. Formally, the hierarchical structure of HMML is represented as follows: at the highest level, the mathematical modeling domains are denoted as $\mathcal{T} =\{ \mathcal{T}^{(1)}, \mathcal{T}^{(2)}, \cdots,\mathcal{T}^{(n)} \}$. Each modeling domain subtree $\mathcal{T}^{(i)}$ is further subdivided into multiple subdomains: $\mathcal{T}^{(i)} =\{ \mathcal{T}^{(i,1)}, \mathcal{T}^{(i,2)}, \cdots,\mathcal{T}^{(i,k)} \}$. Within each subdomain $\mathcal{T}^{(i,j)}$, specific method nodes $\mathcal{N}^{(i,j,l)}$ are structured explicitly as tuples: $\mathcal{N}^{(i,j,l)} =\{\text{modeling method}, \text{core idea}, \text{application}\}$. Here, $\textit{modeling method}$ provides a high-level introduction to the mathematical modeling approach, $\textit{core idea}$ describes the fundamental principles underpinning the modeling method, and $\textit{application}$ indicates typical scenarios and delineates their application scope, such as resource allocation optimization and production scheduling.  For example, in the domain of operations research ($\mathcal{T}^{(1)} = \text{Operations Research}$), the subdomain of programming theory ($\mathcal{T}^{(1,1)} = \text{Programming Theory}$) includes the specific method node $\mathcal{N}^{(1,1,1)}$, which involves the modeling method of linear programming, with the core idea of optimization using linear objectives and constraints, and its application in production resource scheduling. The final mathematical modeling library features five domains (e.g., Operations Research, Optimization, Machine Learning, Prediction and Evaluation), with 17 subdomains (e.g., Programming Theory, Graph Theory, Clustering, Statistics, etc.), encompassing approximately 98 modeling methods (e.g., Linear Programming, Ant Colony Optimization, Expectation Maximization, Analytic Hierarchy Process, Kolmogorov-Smirnov Test). 

\section{Experiments Setup}~\label{sec:appendix_exp_setup}

\textbf{Evalaution.} Since real-world mathematical modeling problems are open-ended and lack standard answers, we follow official Mathematical Modeling evaluation standards to assess the final solution report. We evaluate the agent’s mathematical modeling capabilities across four key aspects:
(1) \textit{Analysis Evaluation.} Assesses the clarity of problem definition, the identification of key components, and the coherence of sub-tasks in relation to the overall goal.
(2) \textit{Modeling
Rigorousness.} Focuses on rigor and rationality, evaluating whether the assumptions are clearly stated and justified, and whether the chosen methods, metrics, and model structure accurately and scientifically represent the real-world problem.
(3) \textit{Practicality and Scientificity.} Evaluates the practicality and scientific validity of the model, ensuring that it is realistically applicable, provides valuable insights for decision-making, and adheres to scientific principles. This stage also verifies whether the model is theoretically sound and considers all relevant scientific factors to ensure its validity.
(4) \textit{Result and Bias Analysis.} Assesses the clarity, interpretability, and thoroughness of the results and analysis. Additionally, it evaluates how well potential biases, such as data or model bias, are identified, analyzed, and mitigated to enhance robustness and acknowledge model limitations. 

 \textbf{Baselines.} We compare MM-Agent with both human team competition solutions and existing LLM-based agents. Since there is no prior work specifically addressing mathematical modeling problems, we adopt other LLMs agents designed for autonomous research to tackle these problems. Specifically, our baselines include: (1) Human Team, using original solutions from real-world mathematical modeling competitions, where teams obtained at least an Honorable Mention. These award-winning solutions serve as a benchmark reference; (2) LLM, where a LLM is directly used to generate mathematical modeling solutions; (3) DS-Agent, a specialized LLM agent for automating data science tasks. We adapt its core design, based on case-based reasoning, to address mathematical modeling problems; (4) ResearchAgent, an LLM-based agent designed to automate research workflows and generate research ideas. We integrate it with a machine learning agent to enhance its capabilities for mathematical modeling; and (5) Agent Laboratory, an LLM-based framework designed to accelerate scientific discovery by guiding the research process through stages of literature review, experimentation, and report writing. For Agent Laboratory, the agent searches arXiv for related papers to identify mathematical modeling methods, which are then used to construct its pipeline for solving modeling problems. 

\section{Other Experiments}~\label{sec:appendix_other_exps}

\subsection{Experiments on Other Models}

The experimental results in Table~\ref{tab:main_exps} demonstrate that MM-Agent consistently outperforms the baseline agents, DS-Agent and ResearchAgent, across all evaluation metrics in the 2021–2025 mathematical modeling competitions on Qwen2.5-72B.   Since the capabilities of Qwen2.5-72B are weaker than those of GPT-4o and DeepSeek-R1-671B, the agent's laboratory workflow is unable to run efficiently. MM-Agent achieves the highest overall score of 8.37, excelling in Analysis Evaluation (8.61), Modeling Rigorousness (7.59), and Practicality and Scientificity (8.89), reflecting its superior ability in handling complex problems with scientific rigor and practical relevance. The stability of MM-Agent’s performance across the competition years further supports the robustness of its approach, ensuring that its success is not a result of overfitting, but rather a reflection of its effective modeling capabilities.
\begin{table}[ht]
\caption{Experiment results on the 2021–2025 mathematical modeling competitions on Qwen2.5-72B. AE, MR, PS, and RBA denote \textit{Analysis Evaluation}, \textit{Modeling Rigorousness}, \textit{Practicality and Scientificity}, and \textit{Result and Bias Analysis}, respectively.}~\label{tab:main_exps_Qwen}
\centering
\resizebox{0.6\textwidth}{!}{
\begin{tabular}{l|ccccc}
\hline
\multirow{2}{*}{\textbf{Methods}} & \multicolumn{5}{c}{\textbf{2021–2025}} \\
\cline{2-6}
 & \textbf{AE $\uparrow$} & \textbf{MR $\uparrow$} & \textbf{PS $\uparrow$} & \textbf{RBA $\uparrow$} & \textbf{Overall $\uparrow$} \\
\hline
Qwen2.5-72B & 7.58& 3.71 & 8.35 & 5.72 & 6.34 \\
DS-Agent & 8.33 & 7.08 & 8.53 & 7.48 & 7.86 \\
ResearchAgent & 8.17 & 6.94 & 8.73 & 7.63 &  7.87\\
\hline
\textbf{MM-Agent} & \textbf{8.61} & \textbf{7.59} & \textbf{8.89} & \textbf{8.37} & \textbf{8.37} \\
\hline
\end{tabular}
}
\end{table}

\begin{table}[h]
\centering
\caption{Experimental results on average token consumption, cost, and runtime using Qwen-2.5 72B.}
\begin{tabular}{cccc}
\toprule
\multicolumn{1}{c|}{Methods}         & Token  & Cost(\$) & Runtime(s) \\ \midrule
\multicolumn{1}{c|}{DS-Agent}        & 264973 & 0.21     & 1757       \\
\multicolumn{1}{c|}{ResearchAgent}   & 313688 & 0.24     & 2577       \\
\multicolumn{1}{c|}{MM-Agent}        & 455610 & 0.34     & 2691       \\ \bottomrule
\end{tabular}
\label{tab:cost_qwen}
\end{table}
\footnotetext[1]{The execution failed due to the demanding requirements on the instruction-following capabilities of LLMs.}

\begin{table}[]
\caption{Experiment on the well-defined optimization problem under zero-shot setting.}
\centering
\begin{tabular}{c|cccc|c|c}
\toprule
\multirow{2}{*}{Model} & \multicolumn{2}{c}{Linear}                                   & \multicolumn{2}{c|}{Nonlinear}                                & \multicolumn{1}{c|}{\multirow{2}{*}{All}} & \multicolumn{1}{c}{\multirow{2}{*}{Code Pass}} \\
                       & \multicolumn{1}{c}{w/o Table} & \multicolumn{1}{c}{w/ Table} & \multicolumn{1}{c}{w/o Table} & \multicolumn{1}{c|}{w/ Table} & \multicolumn{1}{c|}{}                     & \multicolumn{1}{c}{}                           \\ \midrule
GPT-4o                 & 77.5\%                       & 68.8\%                      & 48.1\%                       & 40.0\%                       & 66.8\%                                   & 89.6\%                                        \\
MM-Agent               & \textbf{79.5\%}                       & \textbf{70.0\%}                         & \textbf{50.4\%}                       & \textbf{42.0\%}                         & \textbf{68.8\% }                                 & \textbf{99.3\%}                                        \\ \bottomrule
\end{tabular}
\label{tab:optibench}
\end{table}

\subsection{Experiments on Well-defined Math Optimization Problems}~\label{sec:mathopt}

To further evaluate the capabilities of our MM-Agent, we extended the experiments to well-defined mathematical optimization problems, including both linear and nonlinear programming. In this setting, the agent is provided with the variables, objective function, and constraints, and is required to solve the optimization problem by directly producing the numerical solution.
Since these problems have well-defined ground-truth solutions, accuracy can be directly used as the metric to evaluate the performance of MM-Agent. Specifically, we conduct experiments on the widely used dataset OPTIBENCH~\cite{yang2025optibench}, and the results are shown in Table~\ref{tab:optibench}. As these optimization problems do not require task decomposition, we appropriately modify the agent's configuration to align with the structure of this task. As shown in Table~\ref{tab:optibench}, MM-Agent consistently outperforms GPT-4o across all subtasks under the zero-shot setting. In linear programming, MM-Agent achieves 79.5\% accuracy without tabular input and 70.0\% with table support, surpassing GPT-4o by 2.0\% and 1.2\%, respectively. Similar trends are observed in nonlinear optimization, where MM-Agent improves performance by 2.3\% (w/o table) and 2.0\% (w/ table). Notably, MM-Agent also achieves a higher overall accuracy (68.8\%) and code pass rate (99.3\%), indicating better robustness and code reliability. These results highlight MM-Agent's enhanced reasoning capability and robustness in solving well-structured mathematical tasks.

\subsection{Human Evaluation Results}

\begin{wrapfigure}{r}{0.52\textwidth}
    \vspace{-4pt}
    \centering
\includegraphics[width=0.45\textwidth]{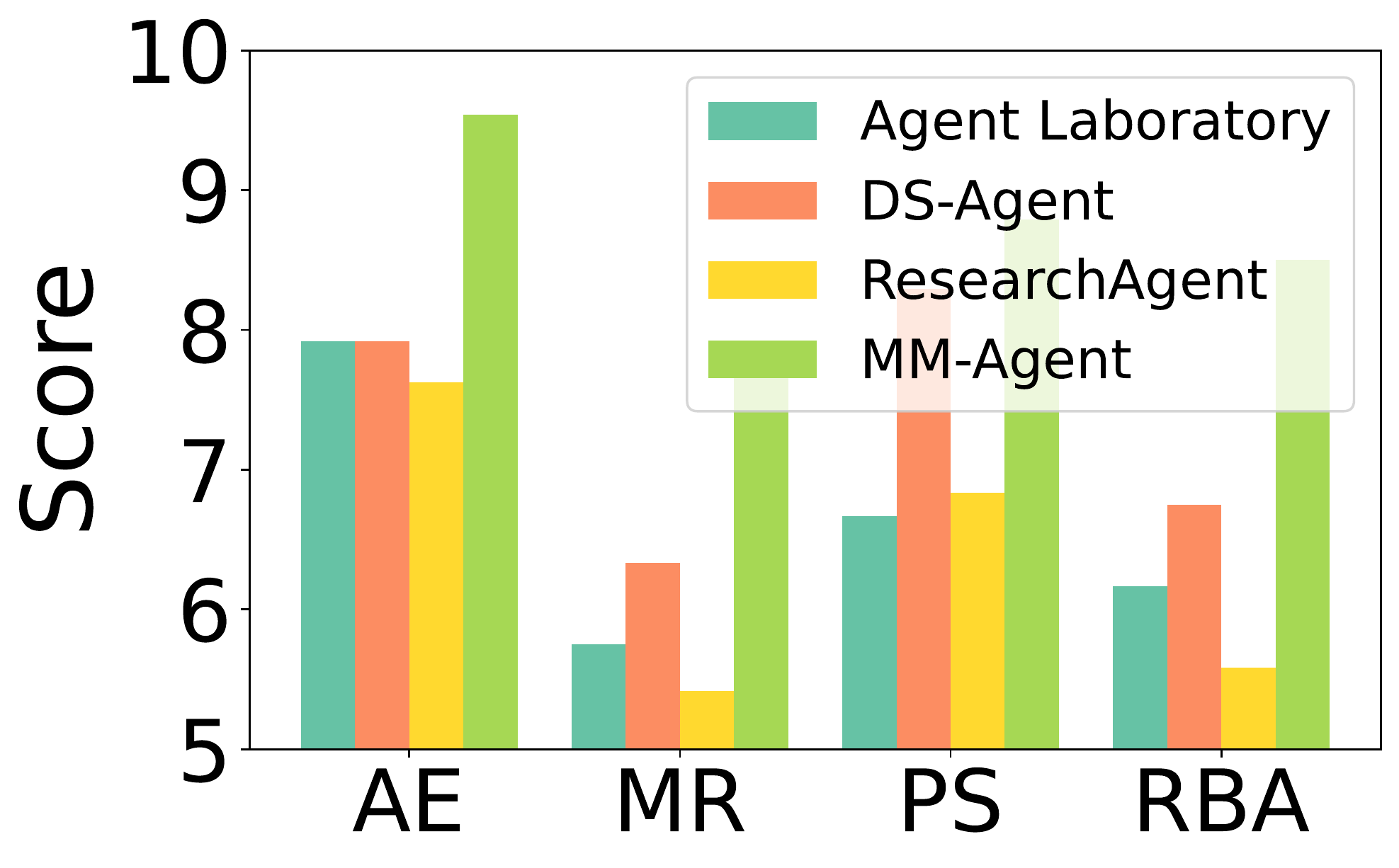}
    \caption{Human Evaluation Results.  }
    \label{fig:human_evaluate}
    \vspace{-5pt}
\end{wrapfigure}
The experiments evaluated by human experts are summarized in Figure~\ref{fig:human_evaluate}. We report average scores across four key dimensions: \textit{Analysis Evaluation (AE)}, \textit{Modeling Rigorousness (MR)}, \textit{Practicality and Scientificity (PS)}, and \textit{Result and Bias Analysis (RBA)}. As shown in the figure, the MM-Agent consistently achieves the highest performance across all dimensions, particularly excelling in AE and RBA. DS-Agent and Agent Laboratory exhibit comparable performance in AE and PS, though the former shows slight superiority in RBA. Notably, ResearchAgent performs competitively in PS but lags behind in MR and RBA, indicating weaker modeling rigor and bias awareness. These results demonstrate the superior overall performance and robustness of MM-Agent under expert evaluation, particularly in producing well-analyzed and unbiased modeling outputs.

\subsection{Annotation Quality}~\label{sec:annotation_quality}
To evaluate annotation quality, we measure inter-annotator agreement on the MM-2025 datasets. We first rank the scores provided by each human and LLM annotator, and then compute the Pearson correlation coefficient between the ranked scores of each annotator pair. The results are summarized in Table~\ref{tab:agreement_summary}.
\textbf{Human-Human Agreements}: The agreement between two human annotators varies across evaluation categories. We observe consistently high agreement in the four metrics—Analysis Evaluation (AE), Modeling Rigorousness (MR), Practicality and Scientificity (PS), and Result and Bias Analysis (RBA)—indicating strong overall consistency among annotators. 
\textbf{Model-Human Agreements}: To further assess annotation reliability, we compare model-generated scores with human evaluations on the same subset. The model demonstrates reasonable alignment with human assessments, particularly in MR and PS,  as reported in Table~\ref{tab:agreement_summary}. 
While RBA and AE exhibit relatively lower agreement compared to the other metrics, this does not necessarily indicate a shortcoming in the quality of problem and result analysis produced by MM-Agent, as demonstrated in Figure~\ref{fig:human_evaluate} and Table~\ref{tab:main_exps}. Instead, we interpret this as reflecting the inherent subjectivity and variability in how such explanations are evaluated by different annotators, a phenomenon also discussed by~\cite{researchagent}

\begin{table}[h]
\centering
\caption{Results of agreements between two human annotation results and between human and model evaluation results.}
\label{tab:agreement_summary}
\resizebox{0.65\textwidth}{!}{
\begin{tabular}{lcccc}
\toprule
\textbf{Categories} & \textbf{AE} & \textbf{MR} & \textbf{PS} & \textbf{RBA} \\
\midrule
Human and Human & 0.7475 & 0.4813 & 0.7890 & 0.7625 \\
Model and Human & 0.5068 & 0.7130 & 0.7860 & 0.5692 \\
\bottomrule
\end{tabular}
}
\end{table}

\begin{figure}[!t]
    \centering
    \includegraphics[width=\linewidth]{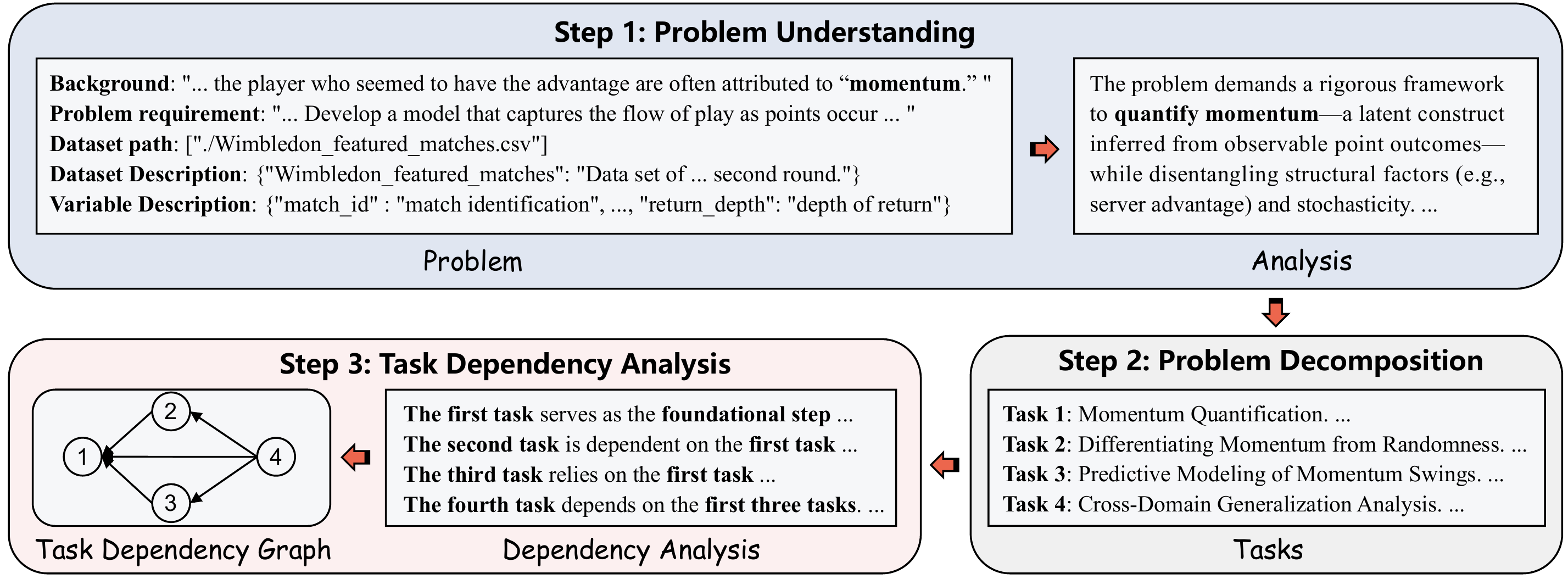}
    \caption{The workflow of the problem analysis phase in MM-Agent. Mathematical modeling tasks often involve interdependent objectives and subtasks. MM-Agent addresses this complexity by decomposing the problem into structured subtasks.}
    \label{fig:problem_analysis}
\end{figure}

\begin{figure}[!t]
    \centering
    \includegraphics[width=\linewidth]{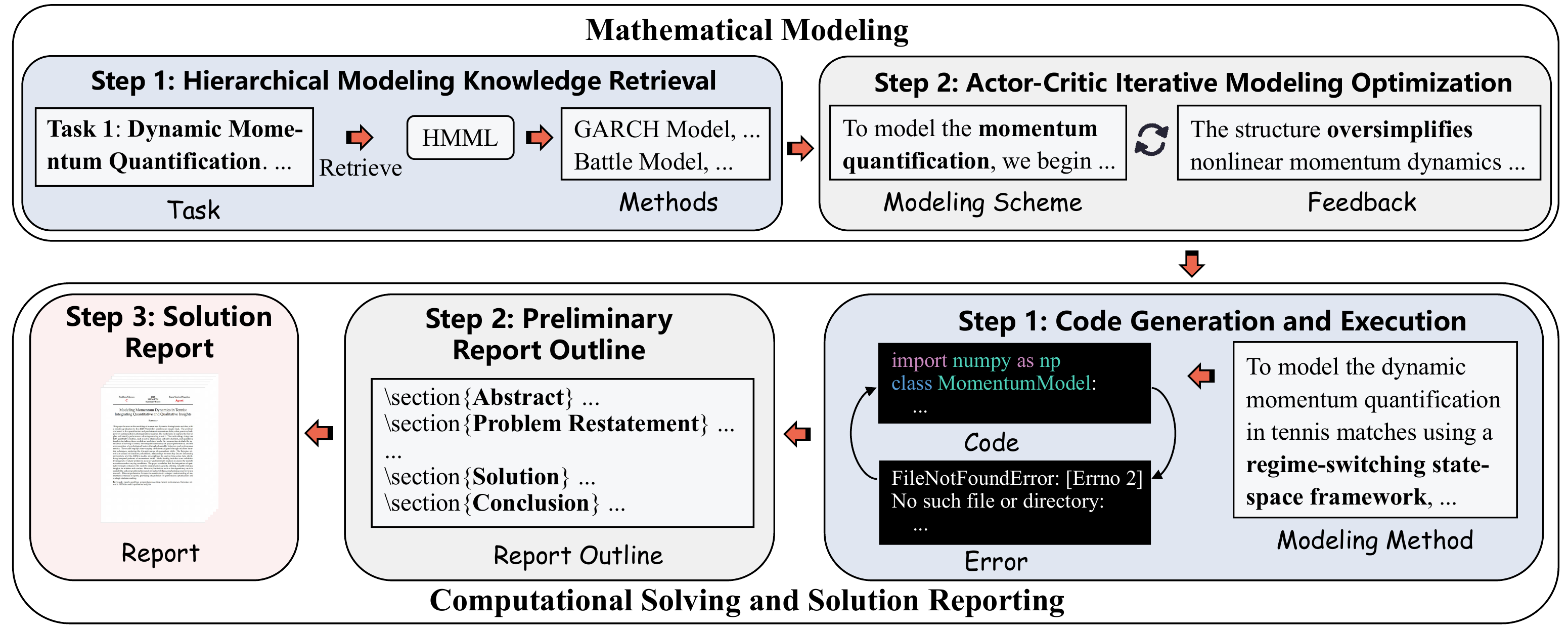}
    \caption{The workflow of the mathematical modeling phase and computational solving and solution reporting phase of MM-Agent.}
    \label{fig:model_and_report}
\end{figure}

\begin{figure}[!t]
  \centering
  \begin{subfigure}{0.19\textwidth}
    \includegraphics[width=\linewidth]{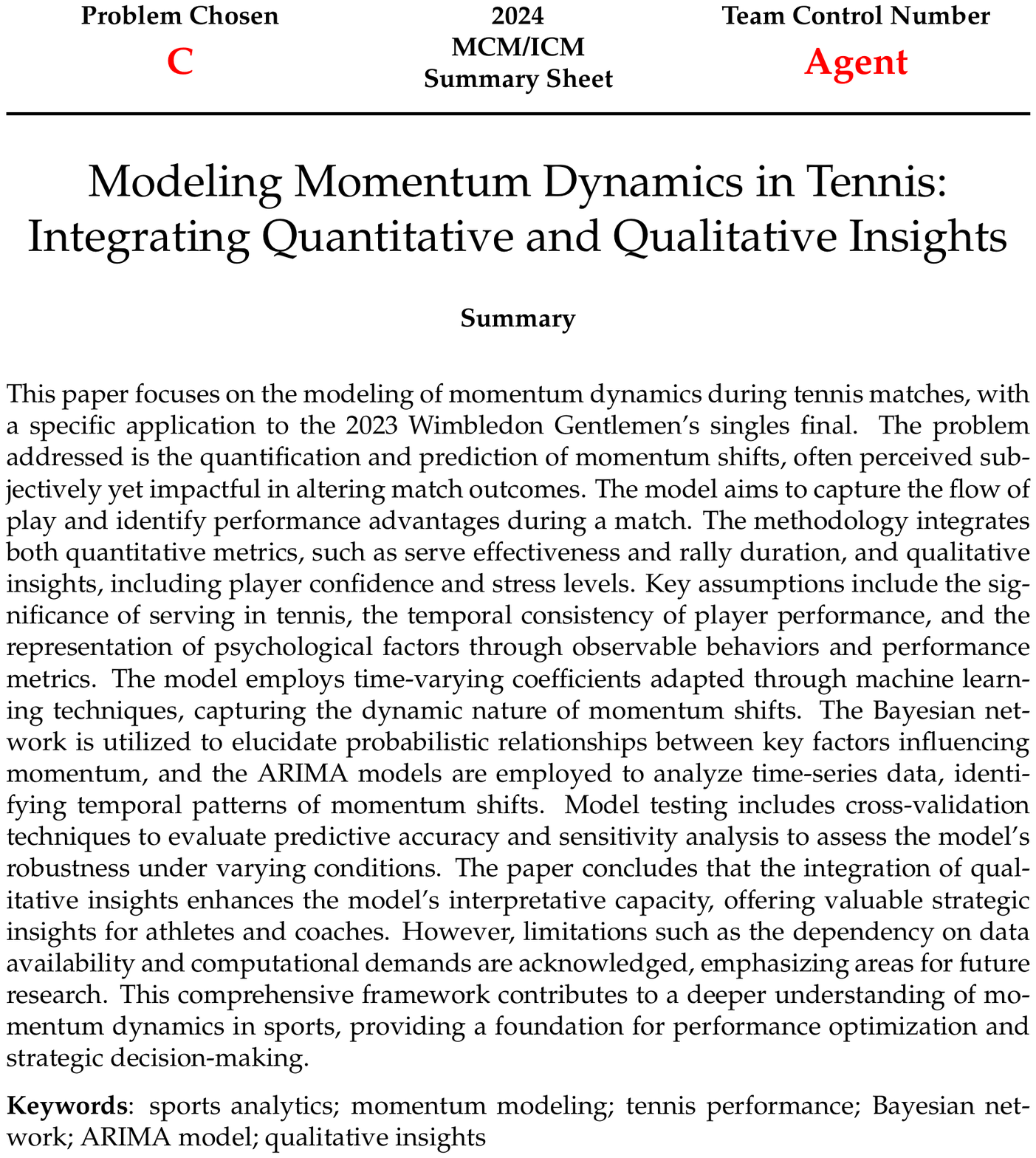}
  \end{subfigure}
  \begin{subfigure}{0.19\textwidth}
    \includegraphics[width=\linewidth]{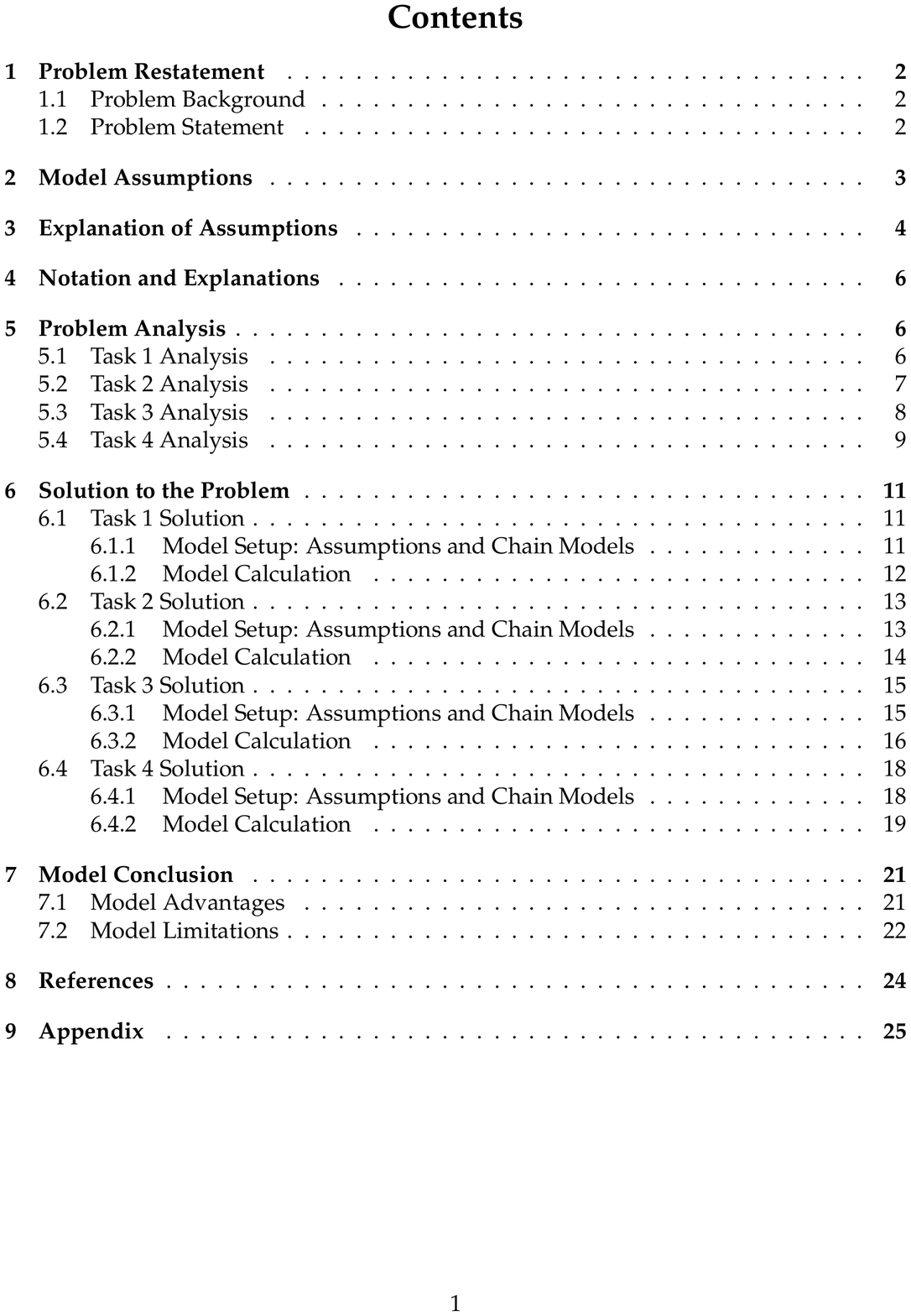}
  \end{subfigure}
  \begin{subfigure}{0.19\textwidth}
    \includegraphics[width=\linewidth]{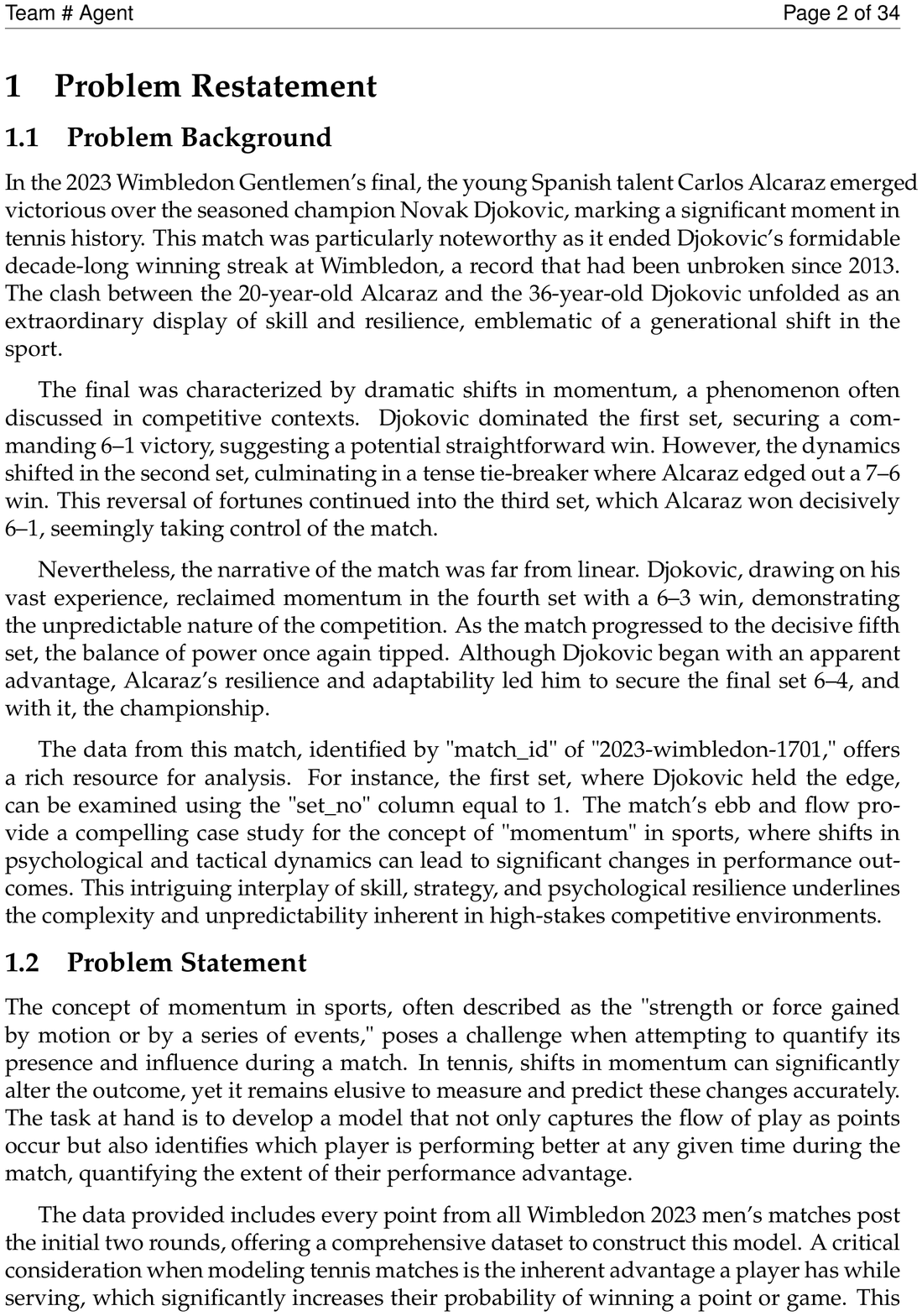}
  \end{subfigure}
  \begin{subfigure}{0.19\textwidth}
    \includegraphics[width=\linewidth]{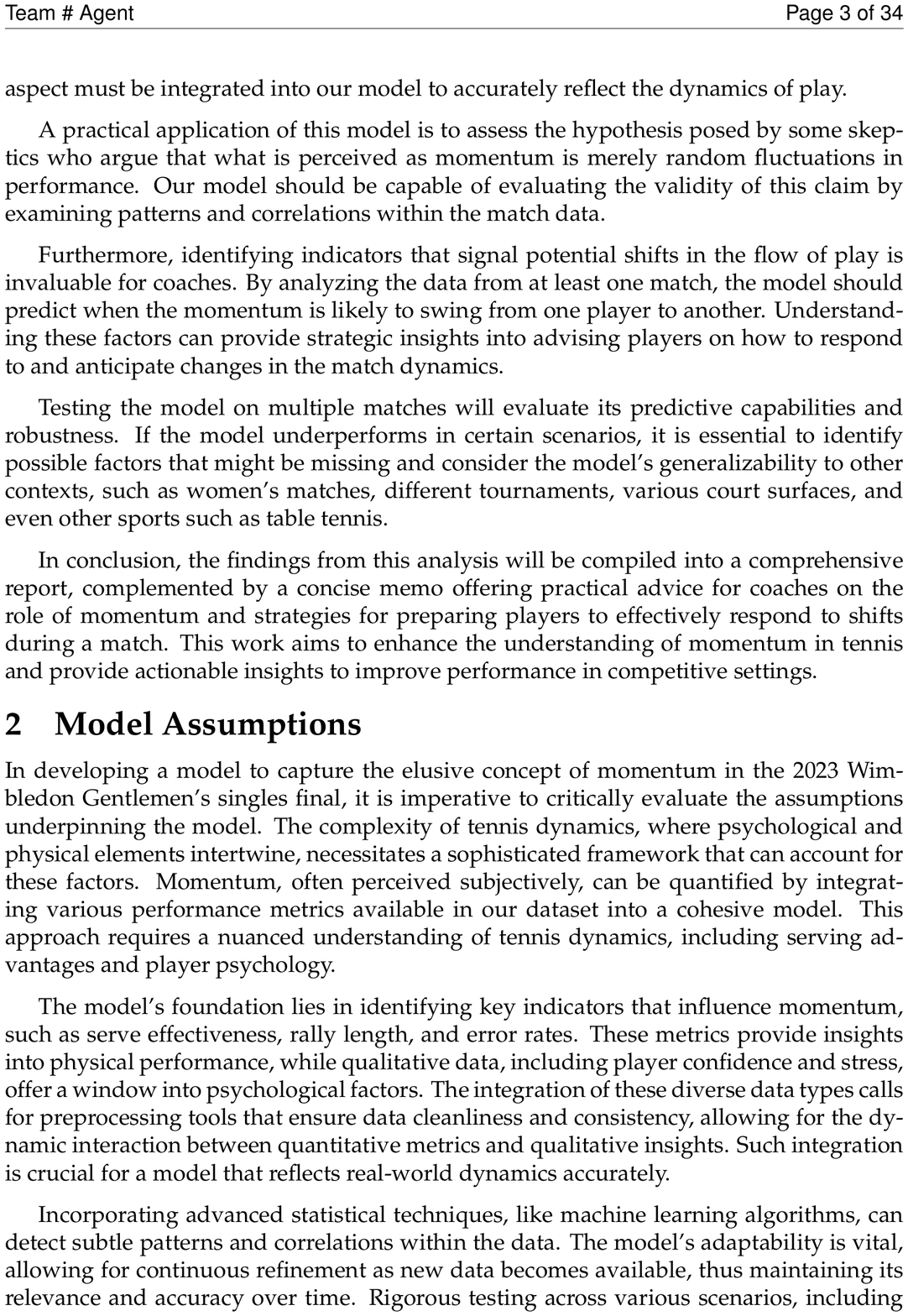}
  \end{subfigure}
  \begin{subfigure}{0.19\textwidth}
    \includegraphics[width=\linewidth]{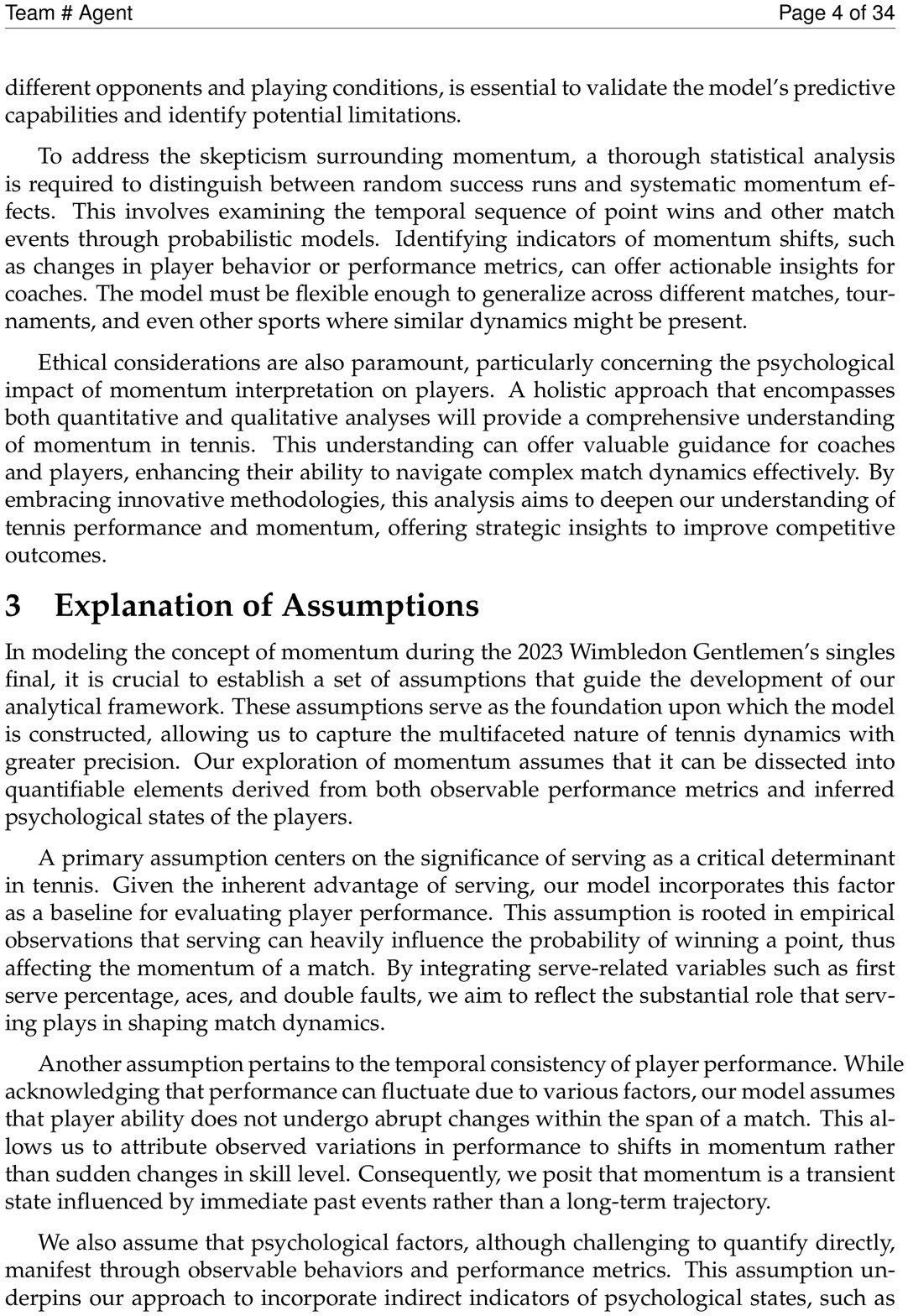}
  \end{subfigure}
  \begin{subfigure}{0.19\textwidth}
    \includegraphics[width=\linewidth]{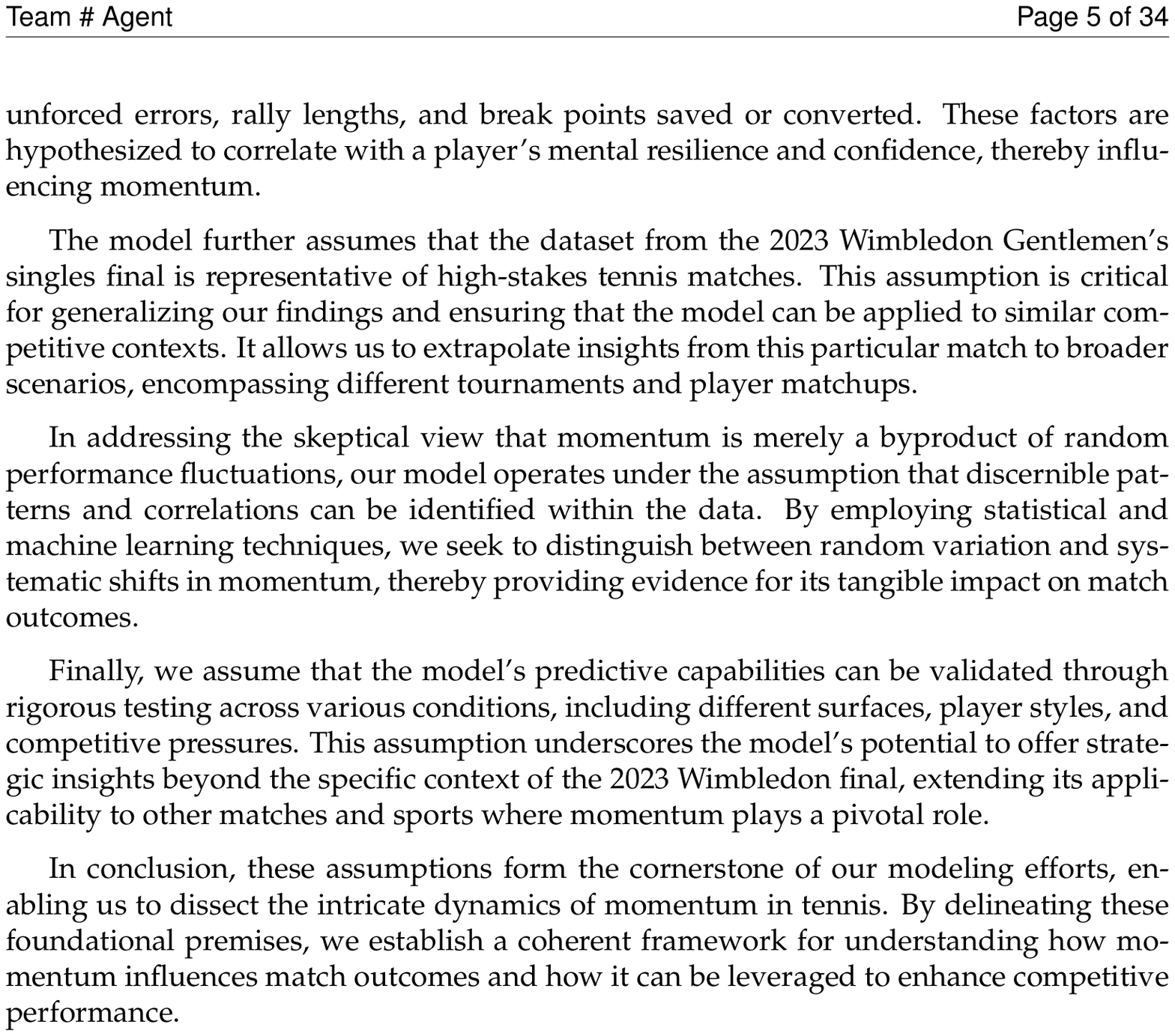}
  \end{subfigure}
  \begin{subfigure}{0.19\textwidth}
    \includegraphics[width=\linewidth]{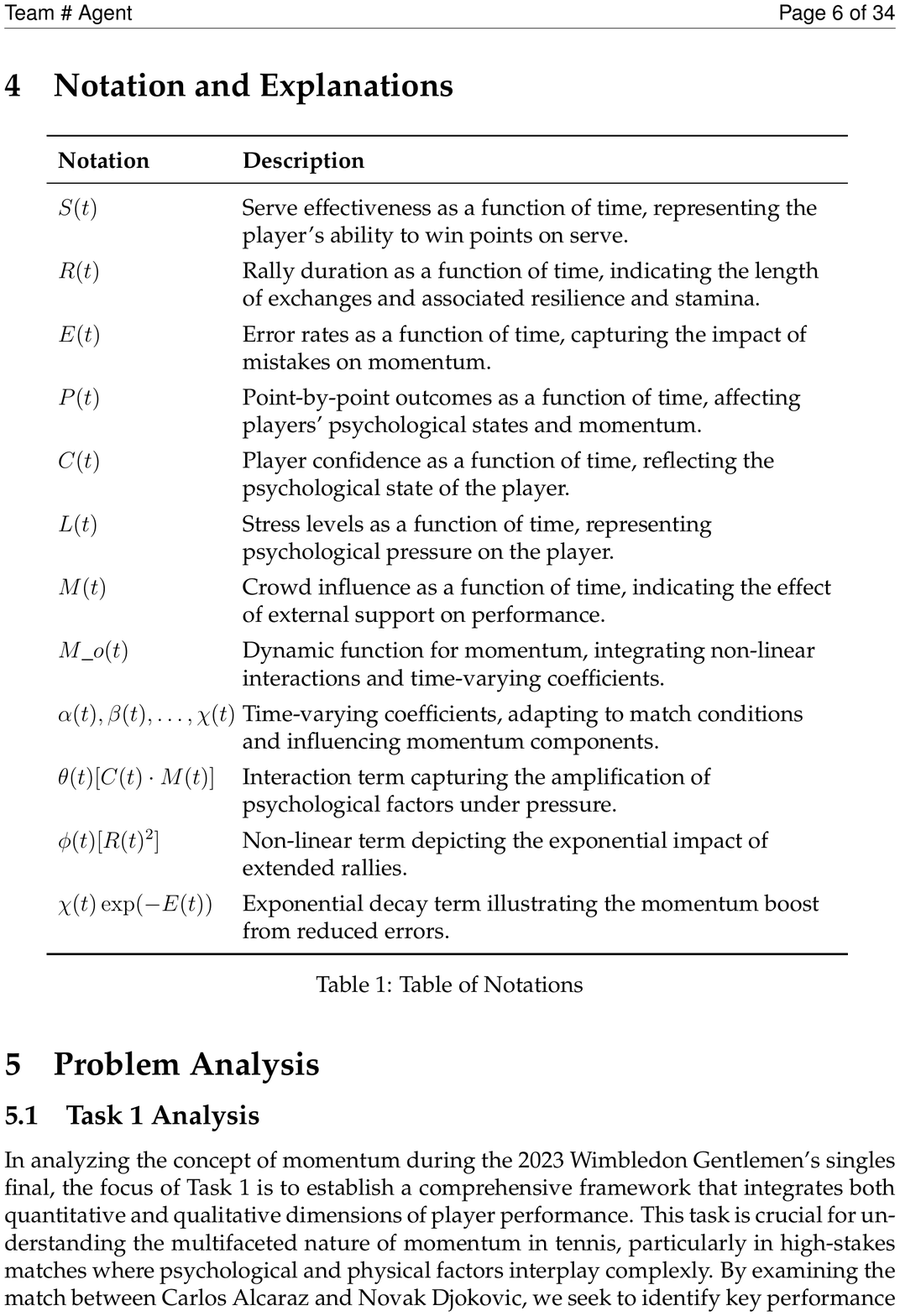}
  \end{subfigure}
  \begin{subfigure}{0.19\textwidth}
    \includegraphics[width=\linewidth]{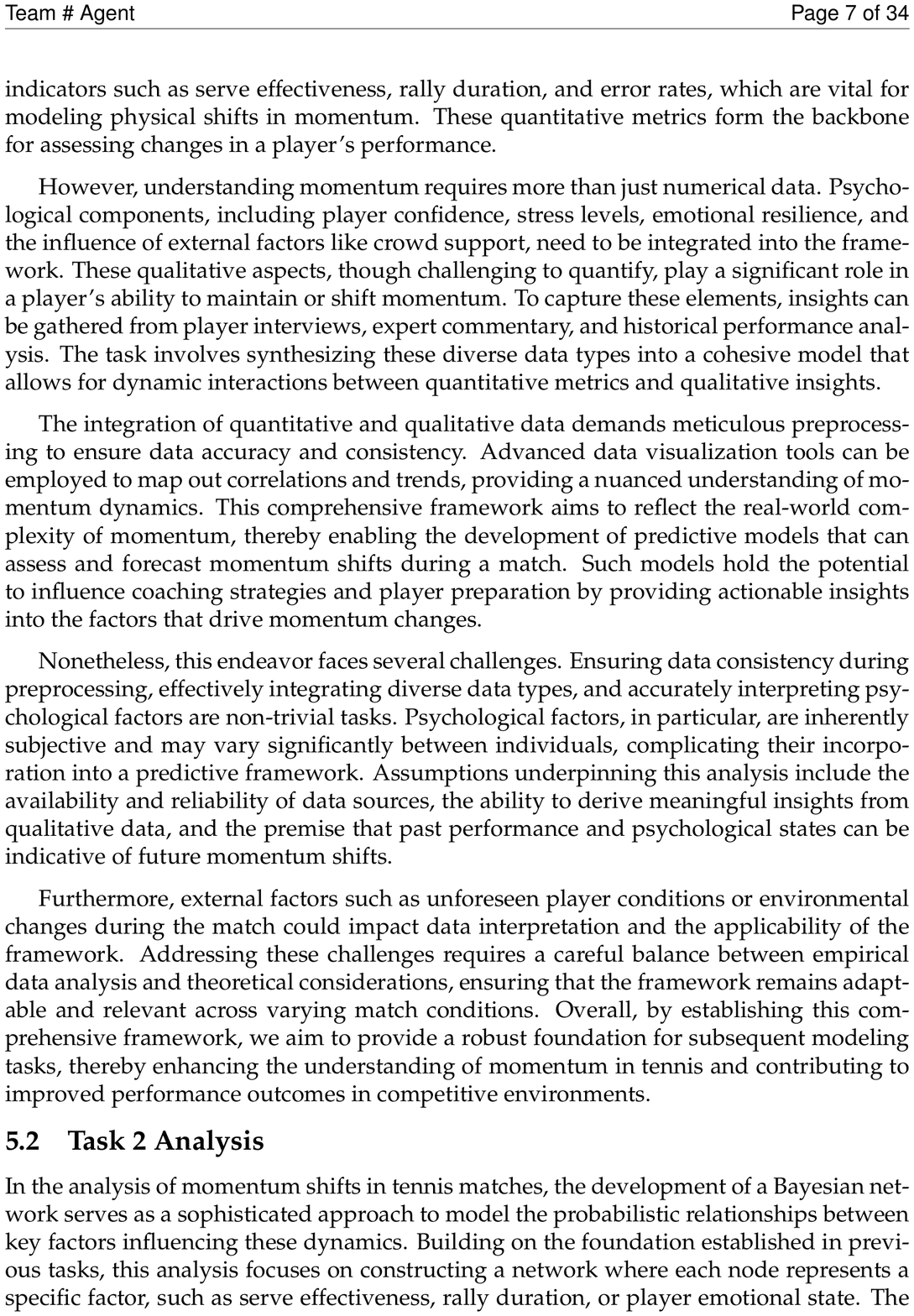}
  \end{subfigure}
  \begin{subfigure}{0.19\textwidth}
    \includegraphics[width=\linewidth]{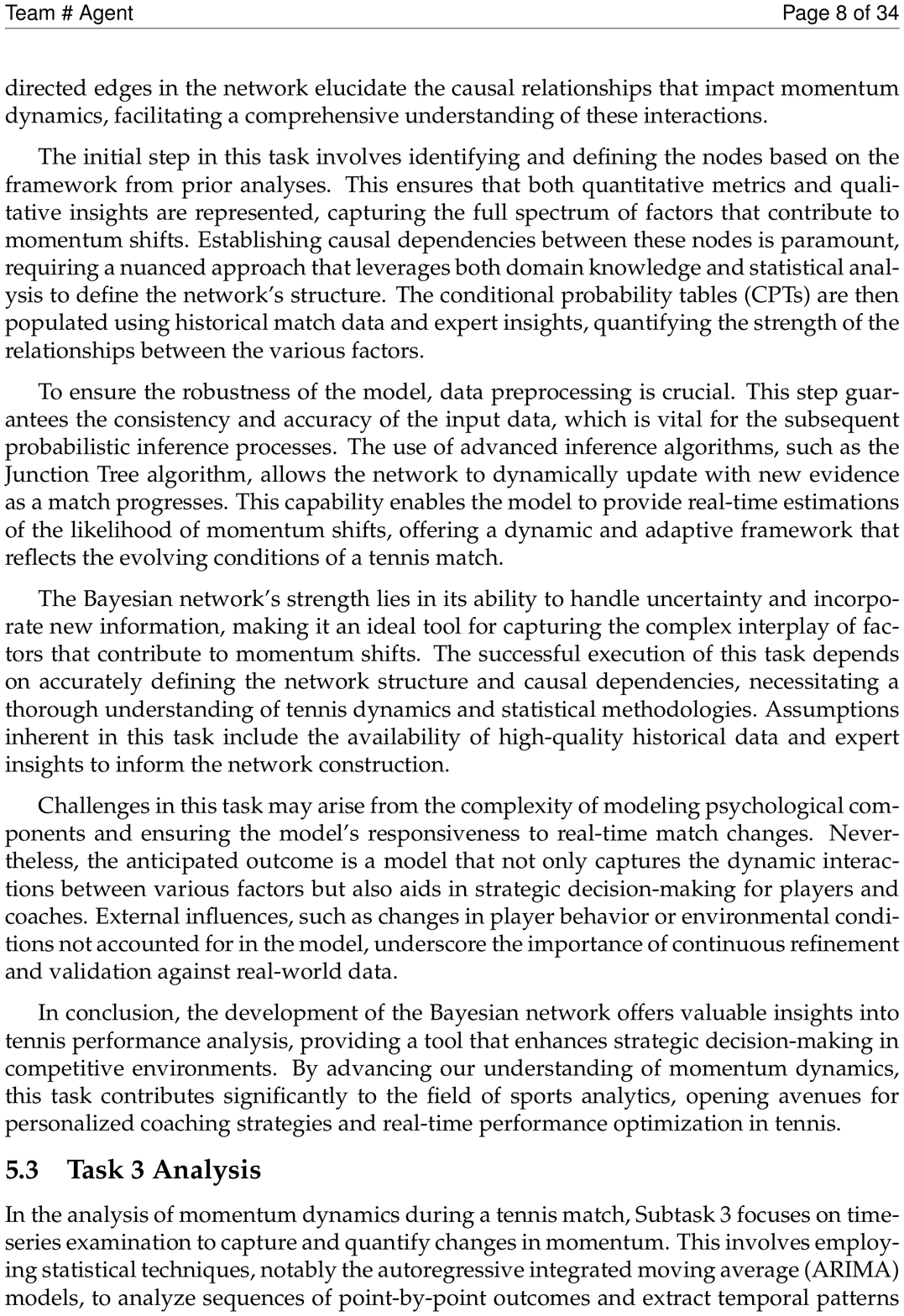}
  \end{subfigure}
  \begin{subfigure}{0.19\textwidth}
    \includegraphics[width=\linewidth]{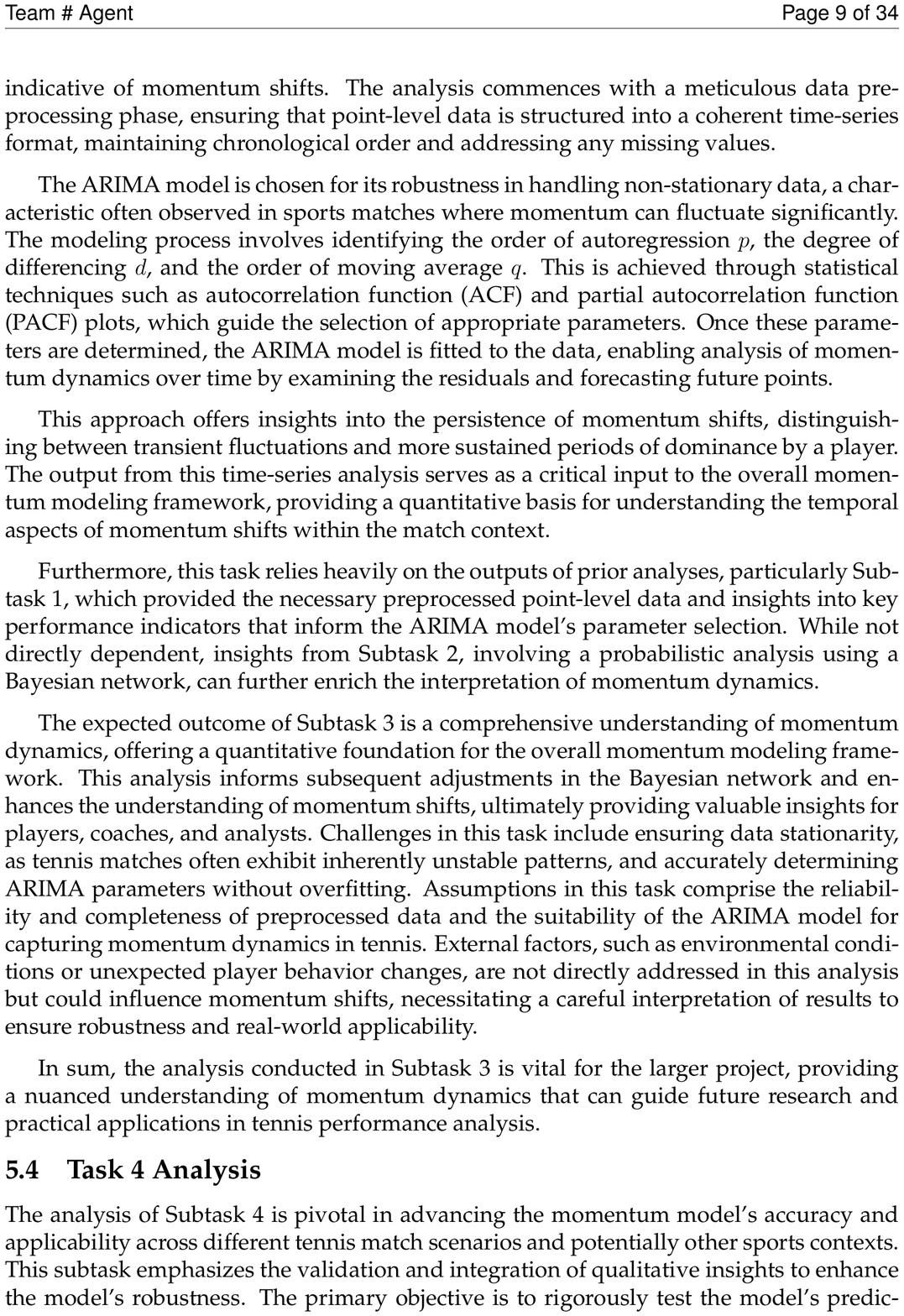}
  \end{subfigure}
  \begin{subfigure}{0.19\textwidth}
    \includegraphics[width=\linewidth]{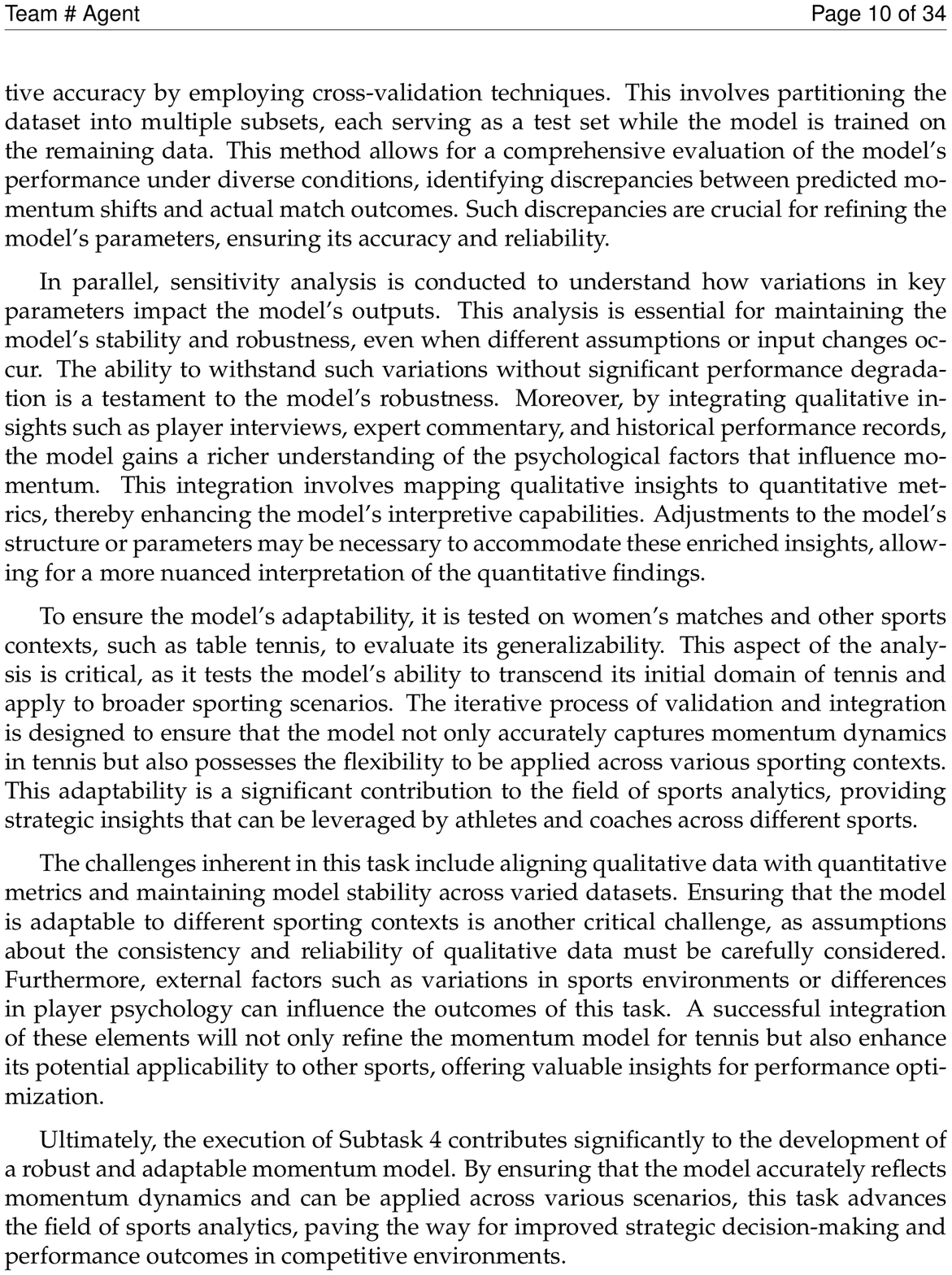}
  \end{subfigure}
  \begin{subfigure}{0.19\textwidth}
    \includegraphics[width=\linewidth]{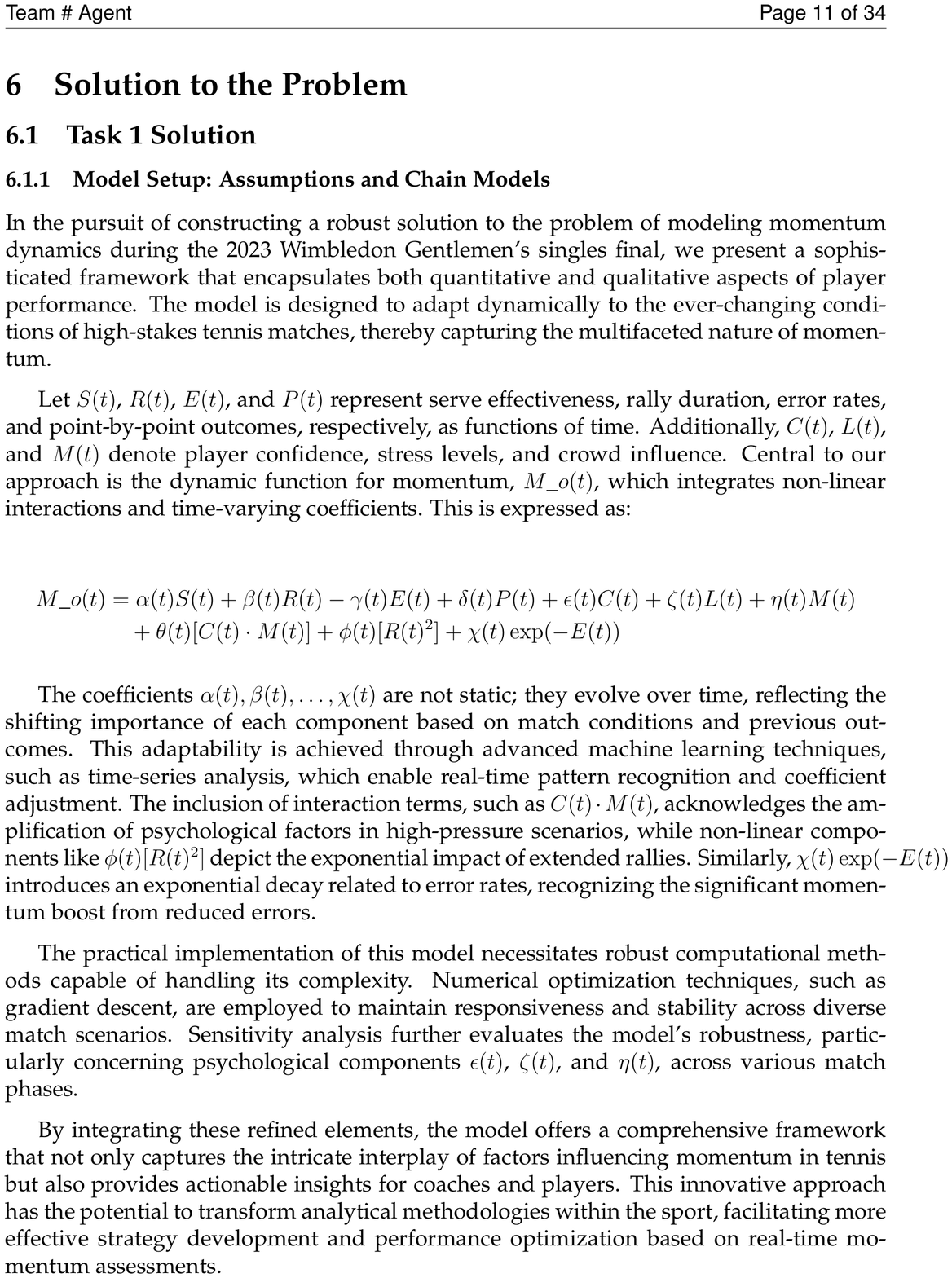}
  \end{subfigure}
  \begin{subfigure}{0.19\textwidth}
    \includegraphics[width=\linewidth]{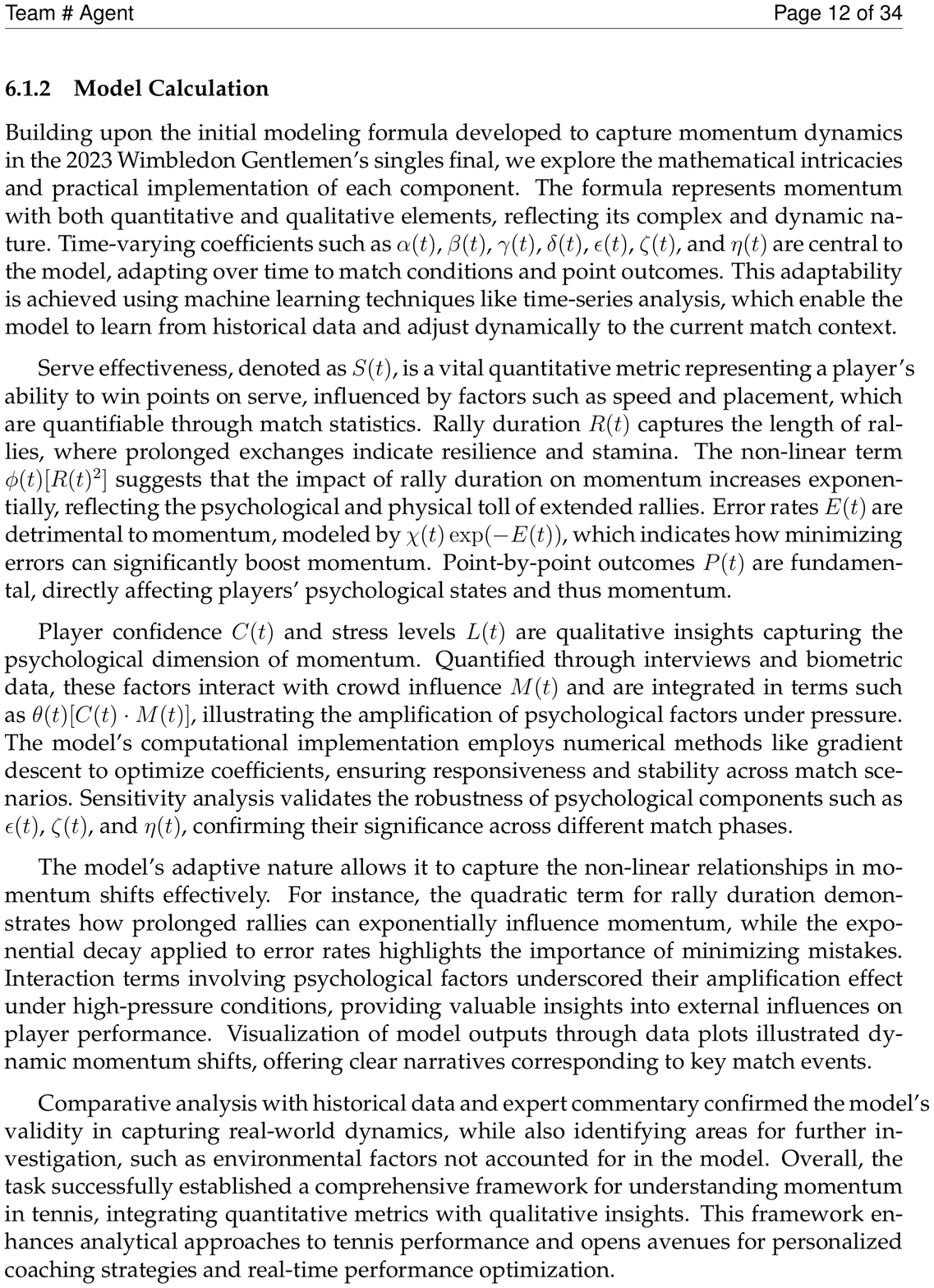}
  \end{subfigure}
  \begin{subfigure}{0.19\textwidth}
    \includegraphics[width=\linewidth]{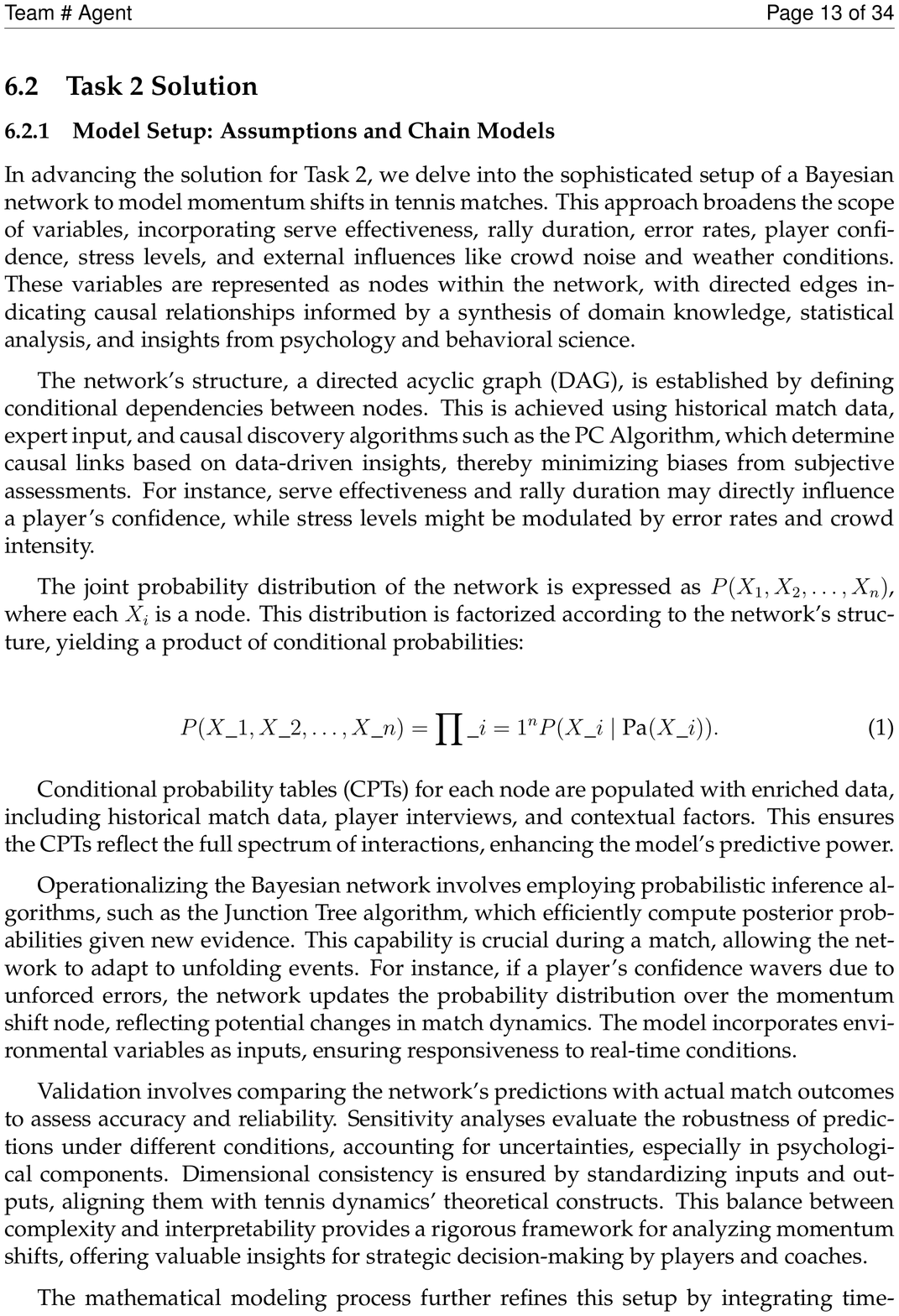}
  \end{subfigure}
  \begin{subfigure}{0.19\textwidth}
    \includegraphics[width=\linewidth]{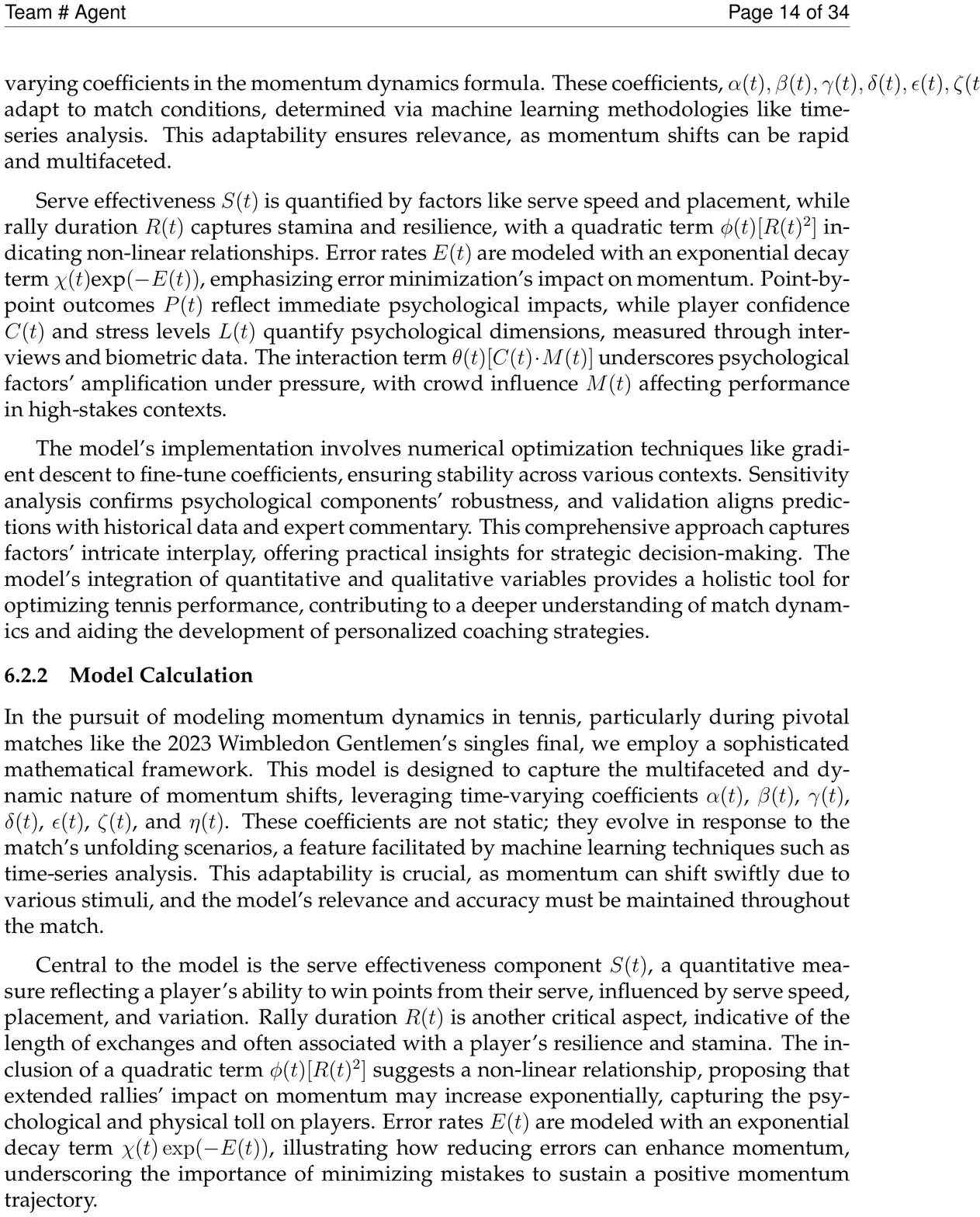}
  \end{subfigure}
  \begin{subfigure}{0.19\textwidth}
    \includegraphics[width=\linewidth]{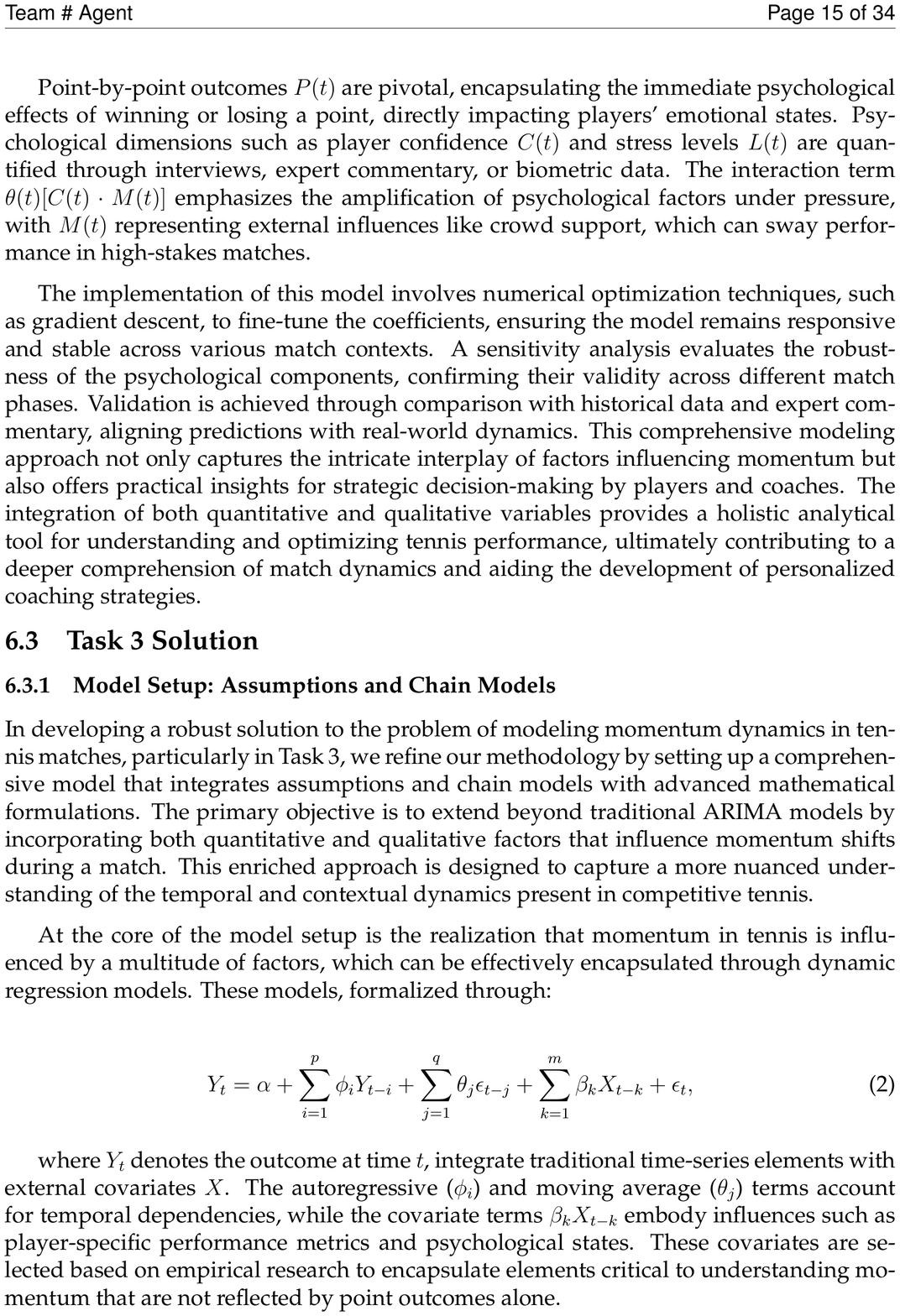}
  \end{subfigure}
  \begin{subfigure}{0.19\textwidth}
    \includegraphics[width=\linewidth]{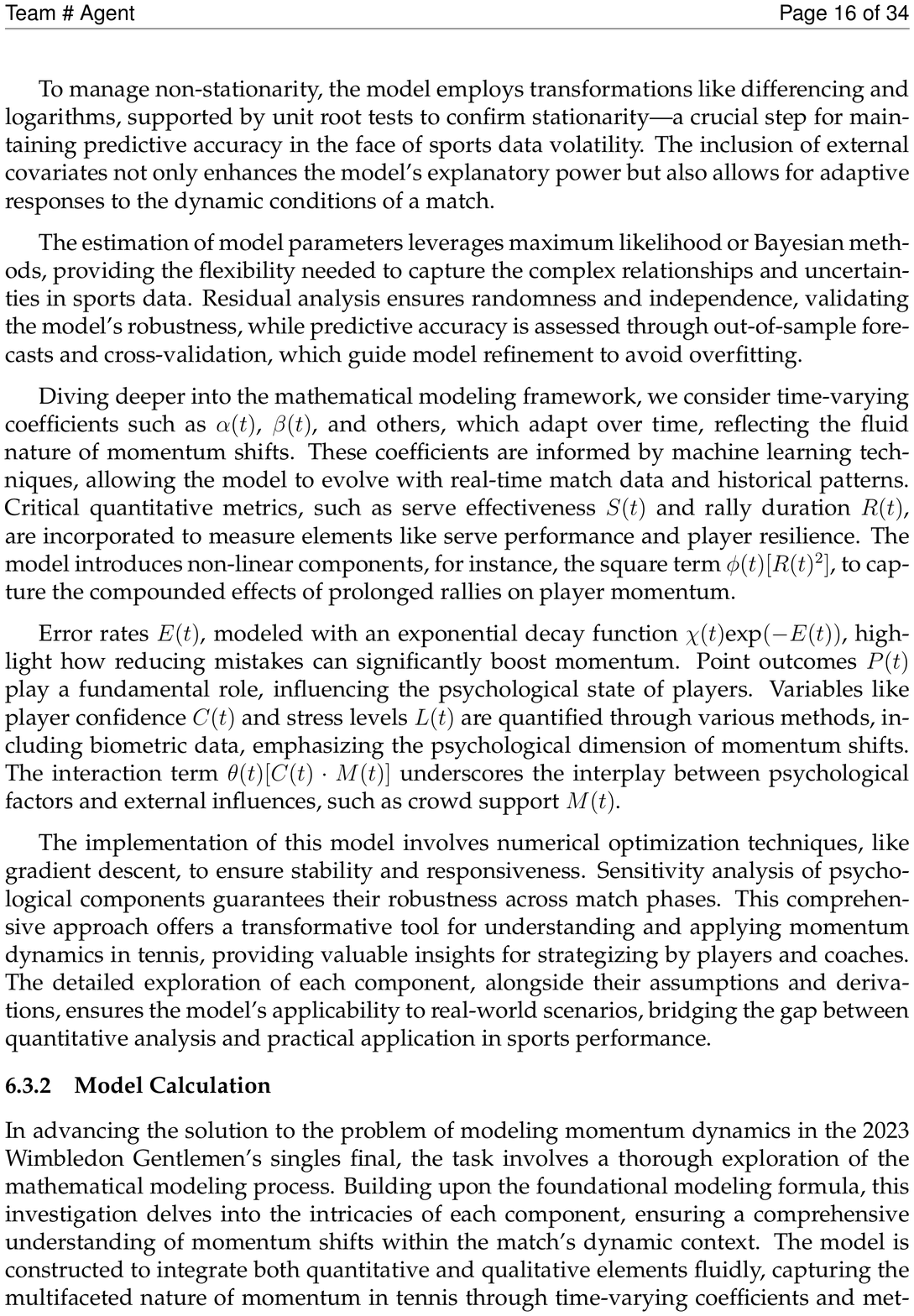}
  \end{subfigure}
  \begin{subfigure}{0.19\textwidth}
    \includegraphics[width=\linewidth]{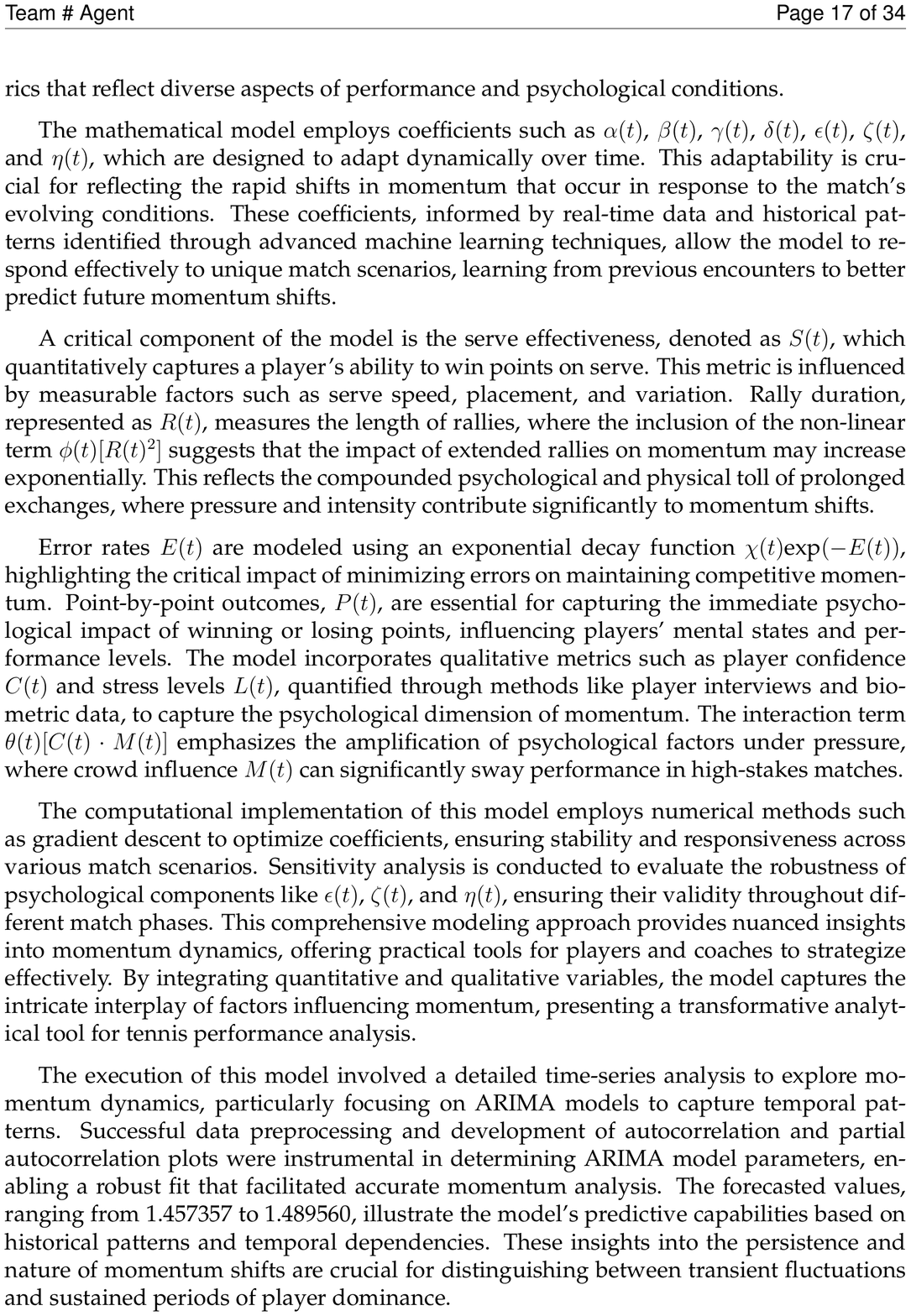}
  \end{subfigure}
  \begin{subfigure}{0.19\textwidth}
    \includegraphics[width=\linewidth]{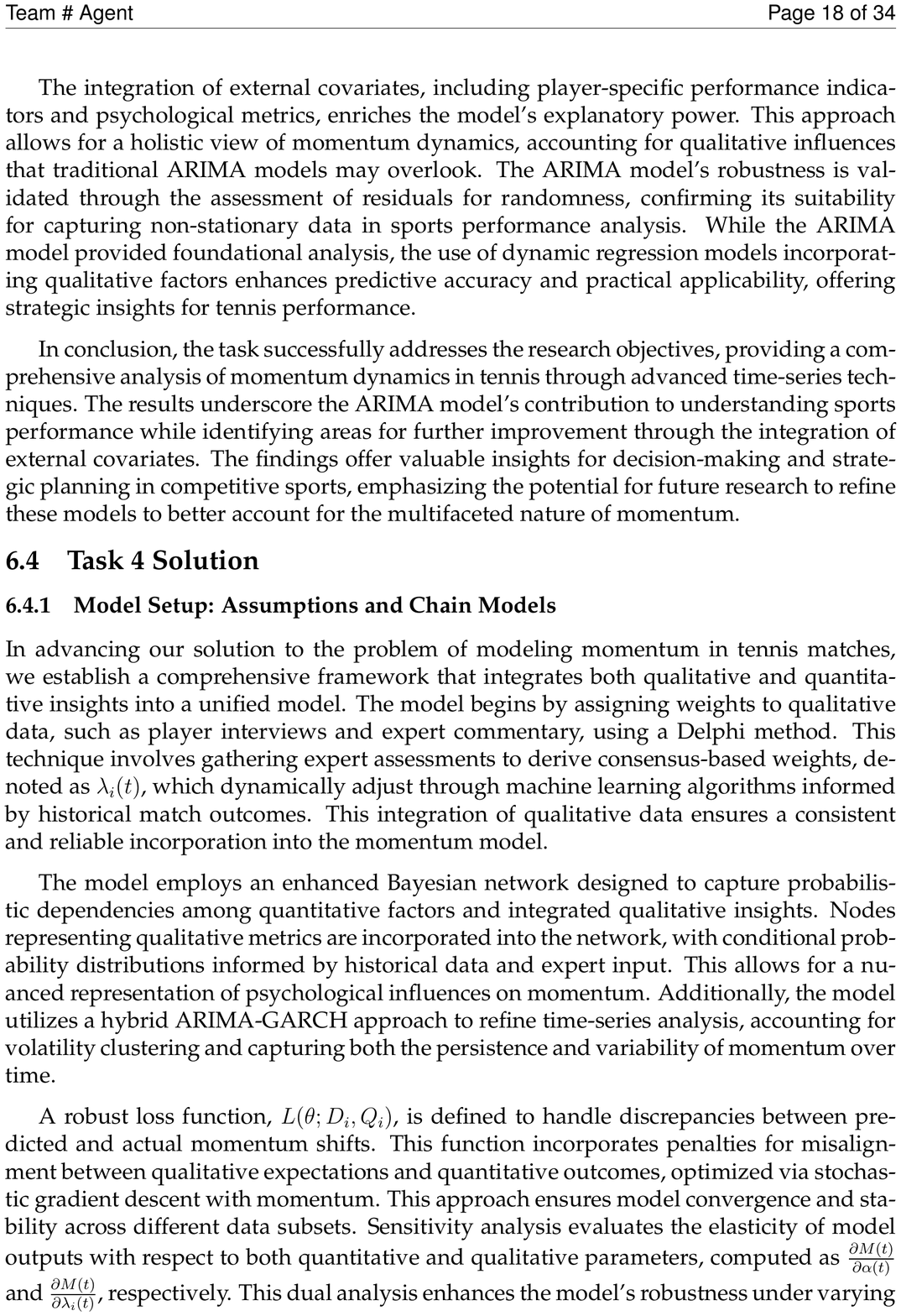}
  \end{subfigure}
  \begin{subfigure}{0.19\textwidth}
    \includegraphics[width=\linewidth]{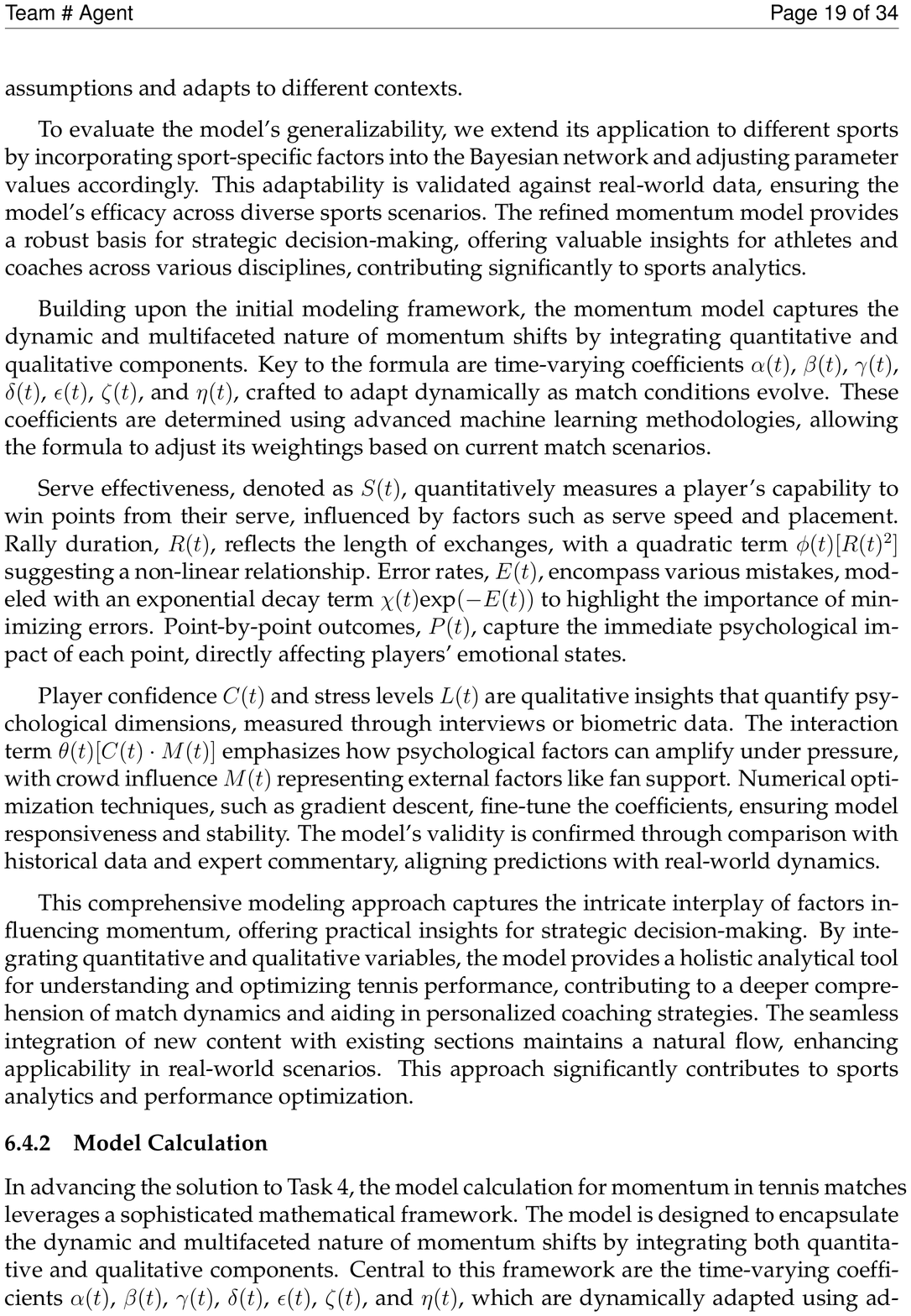}
  \end{subfigure}
  \begin{subfigure}{0.19\textwidth}
    \includegraphics[width=\linewidth]{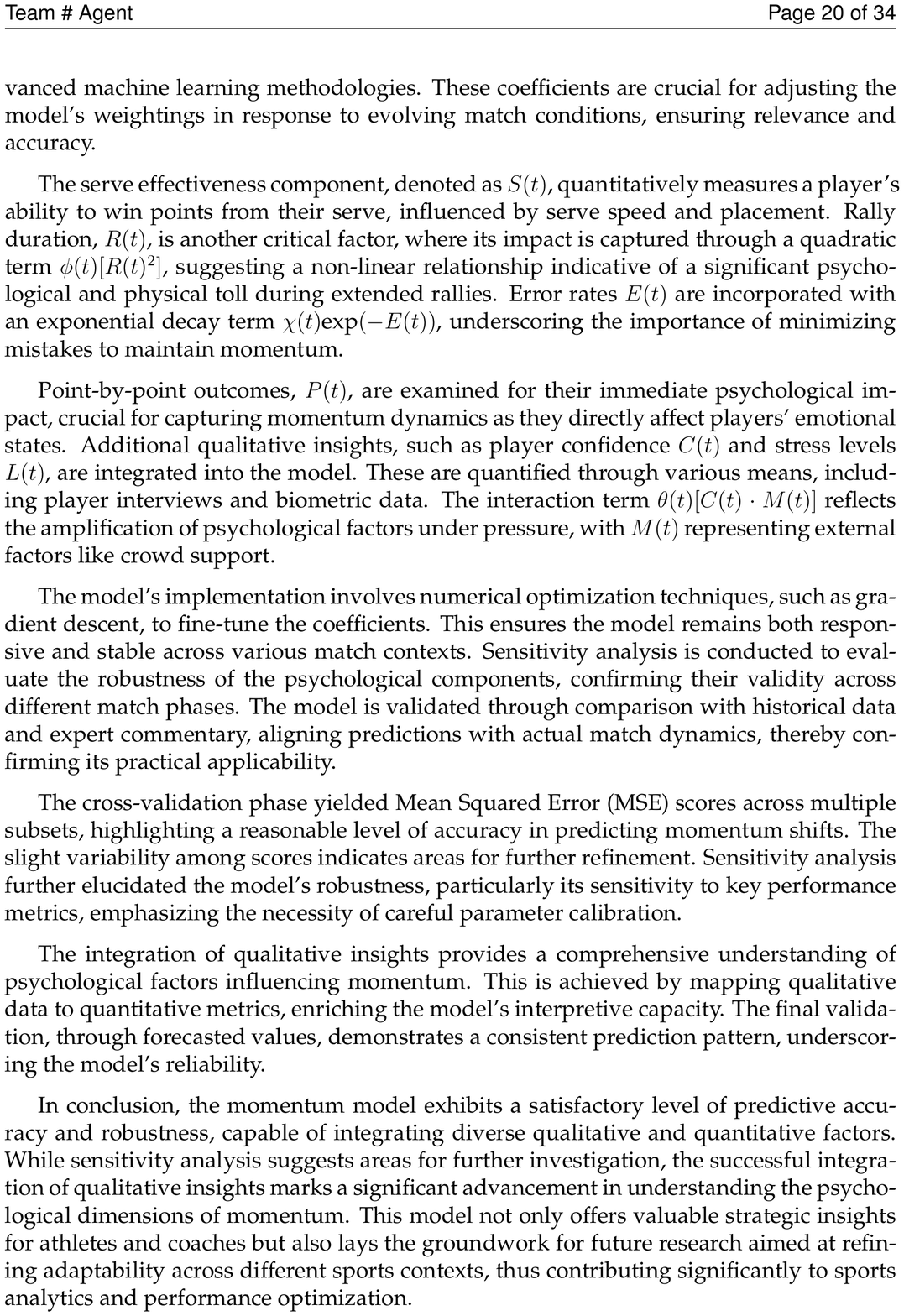}
  \end{subfigure}
  \begin{subfigure}{0.19\textwidth}
    \includegraphics[width=\linewidth]{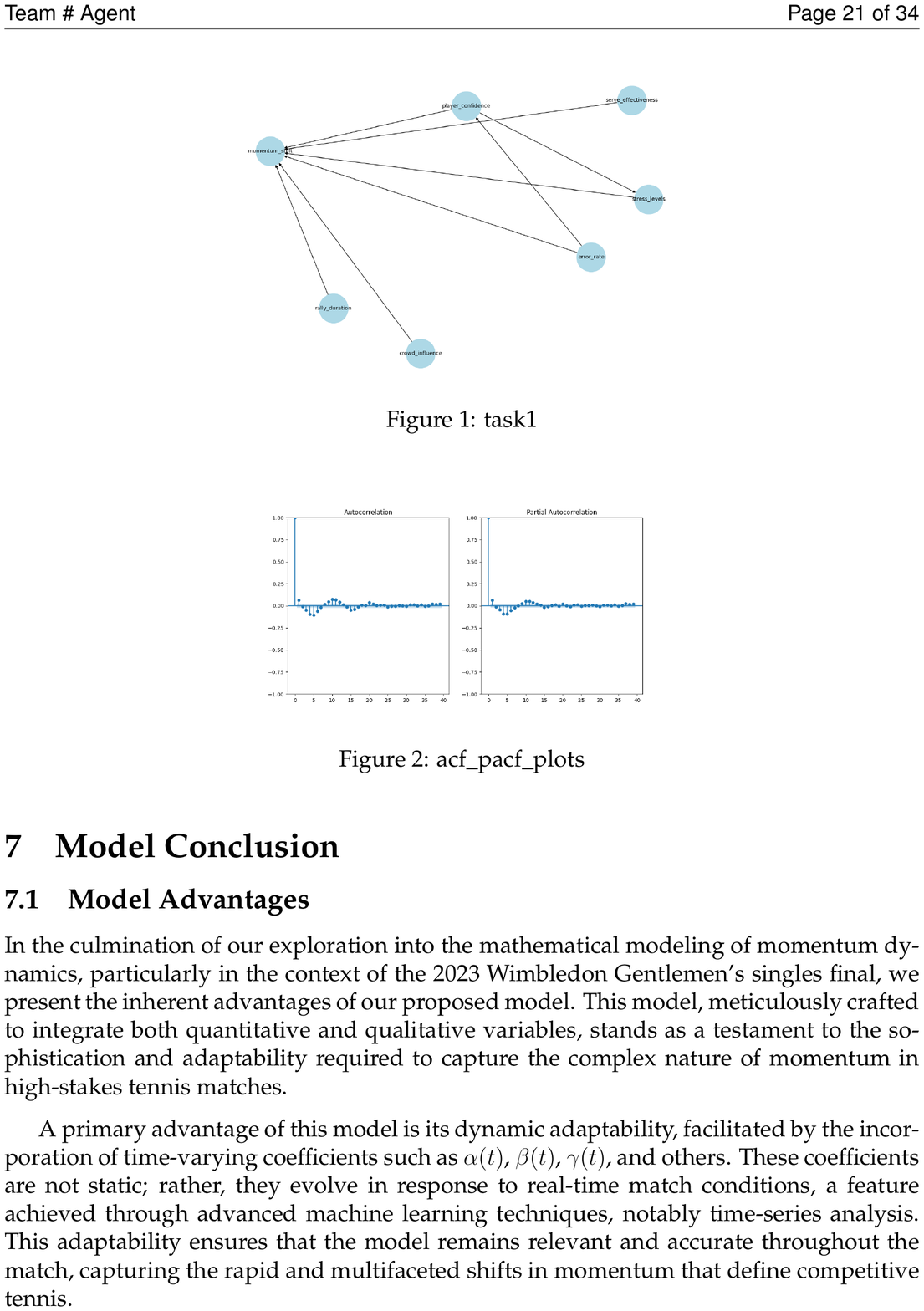}
  \end{subfigure}
  \begin{subfigure}{0.19\textwidth}
    \includegraphics[width=\linewidth]{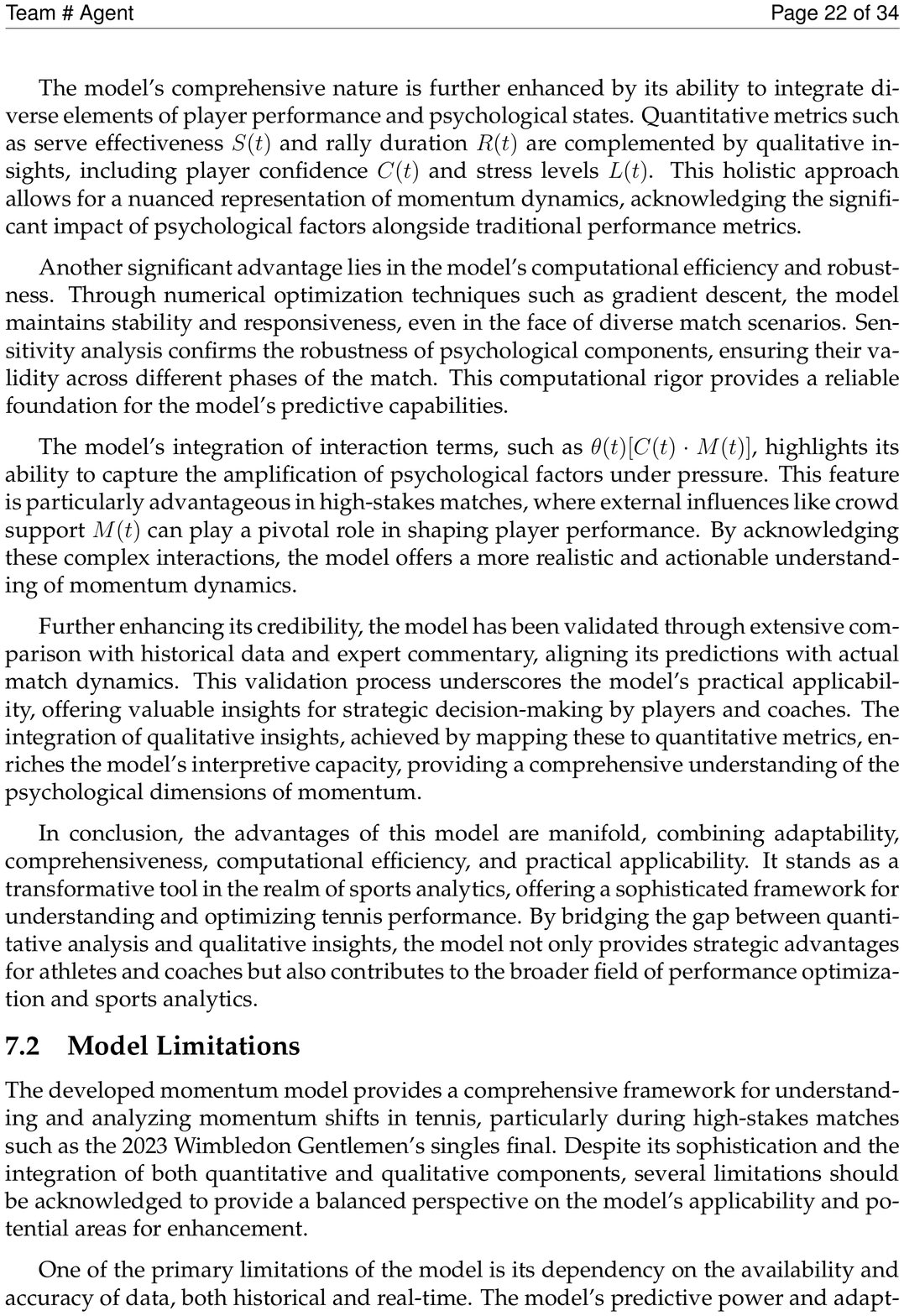}
  \end{subfigure}
  \begin{subfigure}{0.19\textwidth}
    \includegraphics[width=\linewidth]{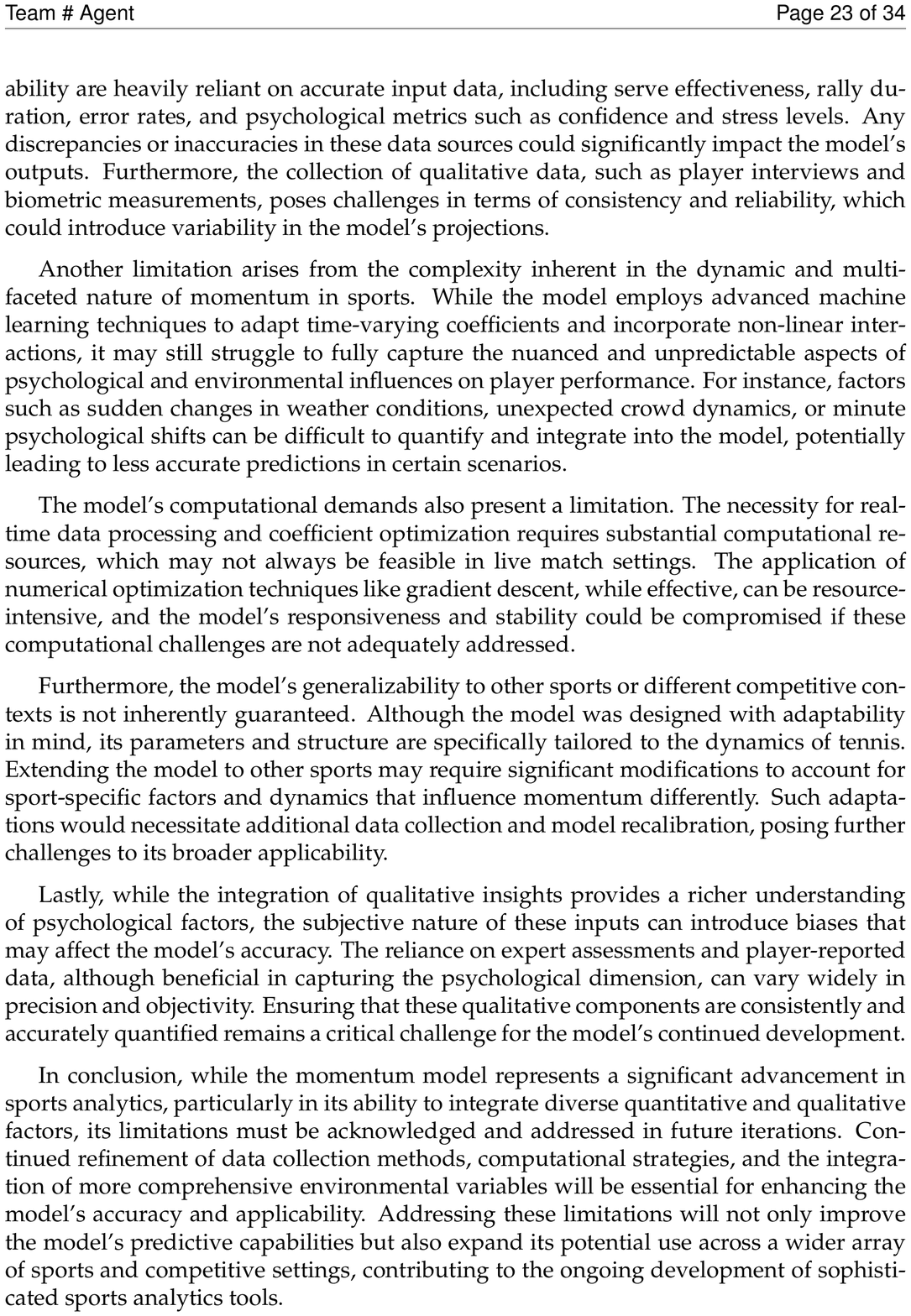}
  \end{subfigure}
  \begin{subfigure}{0.19\textwidth}
    \includegraphics[width=\linewidth]{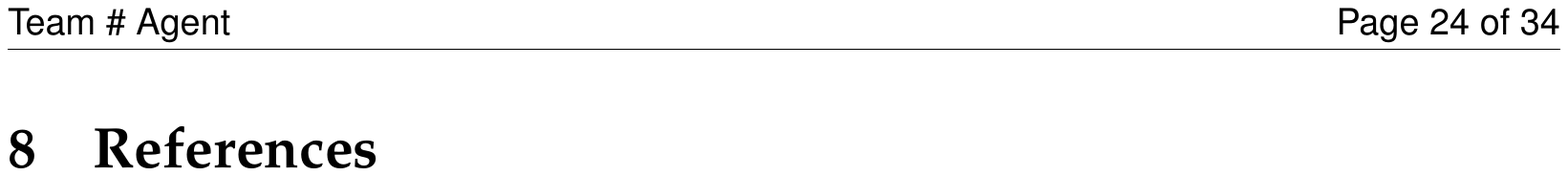}
  \end{subfigure}
  \caption{An example report generated by MM-Agent for Problem C of the 2024 MCM.}
  \label{fig:example}
\end{figure}

\subsection{Case Study}~\label{sec:case_study}
This section provides detailed descriptions of the case study shown in Figures~\ref{fig:problem_analysis} and~\ref{fig:model_and_report}, illustrating how MM-Agent performs end-to-end problem-solving on a real-world mathematical modeling task from the MCM competition. 

In the problem analysis phase (Figure~\ref{fig:problem_analysis}), MM-Agent begins with Step 1: Problem Understanding. It extracts essential elements such as the task background, dataset path, and variable descriptions. For example, the agent identifies that the modeling task involves quantifying "momentum" in tennis matches using a dataset named \texttt{Wimbledon\_featured\_matches.csv}. The agent interprets the modeling goal as constructing a framework that infers momentum from point-level outcomes while accounting for severe advantages and inherent stochasticity. In Step 2: Problem Decomposition, MM-Agent breaks down the overall objective into four coherent subtasks: (1) Momentum Quantification, (2) Differentiating Momentum from Randomness, (3) Predictive Modeling of Momentum Swings, and (4) Cross-Domain Generalization Analysis. This decomposition transforms the open-ended problem into actionable components. Step 3 involves Task Dependency Analysis, where MM-Agent constructs a dependency graph capturing the logical and computational relationships between subtasks. For instance, Task 1 is recognized as foundational, while the remaining tasks build on its outcomes. This structure ensures that the agent follows a semantically grounded modeling order.

In the modeling and reporting phase (Figure~\ref{fig:model_and_report}), Step 1 is Hierarchical Modeling Knowledge Retrieval. The agent retrieves candidate methods from the Hierarchical Mathematical Modeling Library (HMML) based on the task description, including HMMs, GARCH models, and battle models for momentum estimation. Step 2 features an Actor-Critic Iterative Optimization mechanism. The actor module proposes an initial modeling scheme, such as a Hidden Markov Model for momentum quantification. The critic then evaluates the model and returns structured feedback, noting that the current approach oversimplifies nonlinear momentum dynamics, prompting revisions toward more suitable alternatives like regime-switching models. Step 3 is Code Generation and Execution. MM-Agent generates Python code for the selected modeling pipeline. If execution fails due to errors (e.g., \texttt{FileNotFoundError}), the agent diagnoses the issue and refines the code, continuing the process until a functioning version is achieved.

The final stage involves constructing the solution report. The agent first drafts a preliminary outline, including sections like \texttt{Abstract}, \texttt{Problem Restatement}, \texttt{Solution}, and \texttt{Conclusion}. It then generates a complete report in human-readable LaTeX format, reflecting the entire modeling workflow. The final solution report can be seen in Figure~\ref{fig:example}.

\section{Prompts used for MM-Bench and MM-Agent}~\label{sec:appendix_prompts}
This appendix presents the full set of prompts used in the construction of MM-Bench and the implementation of MM-Agent. These prompts are designed to support the automated, modular, and rigorous execution of real-world mathematical modeling workflows using large language models. Each prompt encapsulates a specific functional objective within the overall agent pipeline, including problem understanding, task decomposition, model formulation, code generation, result interpretation, and solution synthesis. The prompts are carefully structured to align with academic writing and reasoning standards, support multi-agent collaboration, and enable traceable, reproducible modeling. We include both instruction-level and response-format specifications to ensure clarity and operational consistency. Together, these prompts form the foundation of our benchmark and agent framework, enabling end-to-end mathematical modeling automation.

\begin{tcolorbox}[colframe=blue!20!gray, colback=blue!5!white, coltitle=white, fonttitle=\bfseries, title=Analysis Evaluation Prompt, breakable]
Your task is to evaluate the rationality and overall coherence of the problem decomposition into sub-problems by the modeler, given the backgroud and problem requirement  in mathematical modeling.\\
\\
**Background**:\\
\{background\}\\
\\
**Problem Requirements**:\\
\{requirements\}\\
\\
Below is the modeler's task analysis:\\
**Task Analysis**:\\
\{all\_task\_analyses\}\\
\\
**Evaluation Criteria**:\\
\#\#\# 1. Problem Analysis and Understanding\\
\\
\#\#\#\# 1.1 Problem Definition and Goals\\
Ensure the model definition is clear, the analysis is accurate, and the goals are explicit.\\
- Is the scope and goal of the problem clearly defined?\\
- Are the key components of the problem effectively identified?\\
- Are the actual goals that the model aims to solve clearly stated?\\
\\
**Scoring Criteria**:\\
1-2 = Completely unclear; 3-4 = Not clear enough; 5-6 = Basically clear; 7-8 = Clear; 9-10 = Completely clear.\\
\\
\#\#\#\# 1.2 Relevant Scope and Coverage\\
Ensure that the core part of the problem is not deviated from, and whether each sub-task is interrelated and completely covers the actual goals.\\
- Do the sub-tasks have dependencies?\\
- Are all sub-tasks and steps directly related and support the final goal?\\
- Are there any key parts missing or deviations from the actual goals?\\
\\
**Scoring Criteria**:\\
1-2 = Completely deviated from the goal; 3-4 = Partially deviated; 5-6 = Basically covered; 7-8 = Mostly covered; 9-10 = Completely covered.\\
\\
**Output Format**: Please put your evaluation reasons and scores in the tags <reason> your\_reason </reason>, and <score> your\_score </score>.\\
Example:\\
\#\#\# 1.1 Problem Definition and Goals: \textbackslash n\textbackslash n**Evaluation:**\textbackslash n\textbackslash nThe modeler has provided a clear definition of the problem and its goals. However, there are some areas that need further clarification, such as the specific metrics used to measure success and the assumptions made during the analysis. Overall, the problem definition is mostly clear but could benefit from additional detail.\\
**Score:**\textbackslash n<reason> The problem definition is mostly clear but lacks some details </reason>  \textbackslash n<score> 7 </score>  \\
\#\#\# 1.2 Relevant Scope and Coverage: \textbackslash n\textbackslash n**Evaluation:**\textbackslash n\textbackslash nThe sub-tasks are well-defined and cover the main aspects of the problem. There is a logical flow between the tasks, and each task supports the overall goal. However, some sub-tasks could be more detailed to ensure complete coverage of the problem.\\
**Score:**\textbackslash n<reason> The sub-tasks are well-defined but could be more detailed </reason>  \textbackslash n<score> 8 </score>\\
\\
Please objectively and detailedly evaluate the problem analysis and understanding according to the above evaluation criteria, and give the final score and reason.\\
\#\#\# 1.1 Problem Definition and Goals:
\end{tcolorbox}
\noindent\begin{minipage}{\linewidth}
\captionof{figure}{The prompt used for evaluating Analysis of Agent.}
\label{prompt:AEP}
\end{minipage}

\begin{tcolorbox}[colframe=blue!20!gray, colback=blue!5!white, coltitle=white, fonttitle=\bfseries, title=Modeling Rigorousness Evaluation Prompt, breakable]
Your task is to evaluate the rigor and rationality of the modeling given the backgroud and problem requirement in mathematical modeling, particularly focusing on the assumptions and rationality.\\
\\
**Background**:\\
\{background\}\\
\\
**Problem Requirements**:\\
\{requirements\}\\
\\
Below is the modeler's modeling analysis:\\
**Modeling Analysis**:\\
\{all\_task\_analyses\}\\
\\
**Evaluation Criteria**:\\
\#\#\# 2. Rigor and Rationality of Modeling\\
\\
\#\#\#\# 2.1 Assumptions\\
Clear and explicit. These assumptions are the foundation of the model and need to be rigorously justified.\\
- Are the model assumptions clearly explained?\\
- Are the assumptions reasonable and consistent with the background of the actual problem?\\
- Is the rationality and impact of the assumptions considered?\\
\\
**Scoring Criteria**:\\
1-2 = Completely unreasonable; 3-4 = Partially reasonable; 5-6 = Average; 7-8 = Reasonable; 9-10 = Very reasonable.\\
\\
\#\#\#\# 2.2 Rationality\\
The rationality of the model is key to evaluation. Evaluation criteria can include: whether an appropriate model is chosen, whether the model can realistically reflect the problem, etc.\\
- Has the model chosen appropriate methods and metrics?\\
- Does the structure of the model scientifically reflect the actual problem?\\
\\
**Scoring Criteria**:\\
1-2 = Completely unreasonable; 3-4 = Partially reasonable; 5-6 = Average; 7-8 = Reasonable; 9-10 = Very reasonable.\\
\\
**Output Format**:\\
Example:\\
\#\#\# 2.1 Assumptions\textbackslash n\textbackslash n**Evaluation:**\textbackslash n\textbackslash nThe assumptions are crucial for model building, but the modeling analysis does not describe the assumptions in sufficient detail. The rationality and impact of the assumptions are not fully justified, lacking detailed explanations of data sources, data distribution, and competition characteristics. For example, the assumption about "serve advantage" is mentioned but not detailed on how it is quantified and integrated into the model. Additionally, the assumptions are not clearly explained, making the foundation of the model less robust.\\
**Score:**\textbackslash n<reason> The model assumptions are not clear enough and lack sufficient explanation of their sources and impacts </reason>  \textbackslash n<score> 3 </score>  \\
\#\#\# 2.2 Rationality \textbackslash n\textbackslash n**Evaluation:**\textbackslash n\textbackslash nThe rationality of the model is average. The modeler chose to evaluate player performance based on match data (such as points won, games won, and sets won), which is reasonable to some extent. However, the specific modeling methods and metrics are not detailed. For example, how to quantify "performance score", how to handle time series data, and whether psychological factors in the competition are considered. Although some possible methods (such as time series analysis, regression, or classification) are mentioned, their specific applications and reasons for selection are not deeply explained. The structure of the model may have certain limitations in reflecting the actual problem.\\
**Score:**\textbackslash n<reason> The rationality of the model is average, with methods and metrics not detailed, and the model structure has limitations </reason>  \textbackslash n<score> 5 </score>\\
\\
Please objectively and detailedly evaluate the rigor and rationality of the modeling according to the above evaluation criteria, and give the final score and reason.\\
\#\#\# 2.1 Assumptions\textbackslash n\textbackslash n**Evaluation:
\end{tcolorbox}
\noindent\begin{minipage}{\linewidth}
\captionof{figure}{The prompt used for evaluating Modeling Rigorousness of Agent.}
\label{prompt:MREP}
\end{minipage}

\begin{tcolorbox}[colframe=blue!20!gray, colback=blue!5!white, coltitle=white, fonttitle=\bfseries, title=Practicality and Scientificity Evaluation Prompt, breakable]
Your task is to evaluate the practicality and scientificity of the modeling process given the background and problem requirements in mathematical modeling, particularly focusing on whether the model can practically solve the problem and whether it adheres to scientific principles.\\
\\
**Background**:\\
\{background\}\\
\\
**Problem Requirements**:\\
\{requirements\}\\
\\
Below is the modeler's modeling process:\\
**Modeling Process**:\\
\{all\_task\_analyses\}\\
\\
**Evaluation Criteria**:\\
\#\#\# 3. Practicality and Scientificity\\
\\
\#\#\#\# 3.1 Practicality\\
- Does the modeling method match the characteristics and requirements of the problem?\\
- Does the model provide meaningful insights beyond mere data fitting? Can its output support decision-making with clear explanations and reliable predictions across different datasets?  \\
- Does the approach go beyond standard machine learning or data processing? Has it been deeply optimized or extended, potentially integrating interdisciplinary methods like mathematical or physical modeling?  \\
- Does the model introduce novel frameworks, constraints, objectives, or data representations? Does it push beyond conventional techniques to propose new theoretical or computational approaches?  \\
- Is the selected modeling method appropriate for the given problem?\\
- Is the model reasonably constructed?\\
- Can the model solve the actual problem?\\
- Are the application scenarios of the model clear? Is it feasible for practical operation?\\
- Can the model's output provide useful information for decision-making or exaplaining or predcting?\\
- Does the approach go beyond basic data analysis and machine learning algorithms?\\
- Does the model demonstrate innovation or creativity in its approach to addressing the problem?\\
- Is the modeling approach tailored to the specific problem rather than using generic methods?\\
\\
**Scoring Criteria**:\\
1-2 = Completely impractical; 3-4 = Partially practical; 5-6 = Average; 7-8 = Practical; 9-10 = Very practical.\\
\\
\#\#\#\# 3.2 Scientificity\\
- Does the model adhere to scientific principles? Is there a theoretical basis?\\
- Are the assumptions and methods of the model scientifically justified?\\
- Does the model consider all scientific factors to ensure its rationality?\\
- Does the approach transcend simple data analysis to incorporate deeper mathematical or domain-specific principles?\\
- Is the approach innovative rather than a standard application of common techniques?\\
- Does the modeling process demonstrate understanding of the problem's unique characteristics?\\
\\
**Scoring Criteria**:\\
1-2 = Completely unscientific; 3-4 = Partially scientific; 5-6 = Average; 7-8 = Scientific; 9-10 = Very scientific.\\

**Output Format**:\\
Example:\\
\#\#\# 3.1 Practicality\textbackslash n\textbackslash n**Evaluation:**\textbackslash n\textbackslash n The model is somewhat practical, but it lacks several key aspects. The modeling method does not fully match the characteristics and requirements of the problem. Additionally, the model does not provide meaningful insights beyond mere data fitting, and its output lacks clear explanations and reliable predictions across different datasets. The approach does not go beyond standard machine learning or data processing, and it has not been deeply optimized or extended to integrate interdisciplinary methods like mathematical or physical modeling. Furthermore, the model does not introduce novel frameworks, constraints, objectives, or data representations, and it does not push beyond conventional techniques to propose new theoretical or computational approaches.\\
**Score:**\textbackslash n<reason> The model lacks several key aspects, including matching the problem characteristics, providing meaningful insights, and introducing novel approaches </reason> \textbackslash n<score> 6 </score>  \\
\\
\#\#\# 3.2 Scientificity\textbackslash n\textbackslash n**Evaluation:**\textbackslash n\textbackslash nThe model adheres to clear scientific principles and employs reasonable theoretical foundations. The assumptions and methods are scientifically justified, and the modeler has thoroughly explained the rationality of the assumptions. Rather than relying solely on basic data analysis techniques, the approach incorporates sophisticated mathematical principles and demonstrates innovative application of theoretical concepts to the specific domain of the problem.\\
**Score:**\textbackslash n<reason> The model adheres to scientific principles, incorporates advanced mathematical concepts, and demonstrates innovative application rather than generic approaches </reason> \textbackslash n<score> 7 </score>\\
\\
Please objectively and detailedly evaluate the practicality and scientificity of the modeling process according to the above evaluation criteria, and provide the final score and reason.\\
\#\#\# 3.1 Practicality\textbackslash n\textbackslash n**Evaluation:
\end{tcolorbox}
\noindent\begin{minipage}{\linewidth}
\captionof{figure}{The prompt used for evaluating Practicality and Scientificity of Agent.}
\label{prompt:PSEP}
\end{minipage}

\begin{tcolorbox}[colframe=blue!20!gray, colback=blue!5!white, coltitle=white, fonttitle=\bfseries, title=Result and Bias Analysis Evaluation Prompt, breakable]
Your task is to evaluate the result analysis and bias analysis of the given modeling report, particularly focusing on the rationality, interpretability of the model output, and the identification and correction of biases.\\
\\
**Background**:\\
\{background\}\\
\\
**Problem Requirements**:\\
\{requirements\}\\
\\
Below is the modeler's modeling report:\\
**Modeling Report**:\\
\{all\_task\_analyses\}\\
\\
**Evaluation Criteria**:\\
\#\#\# 4. Result Analysis and Bias Analysis\\
\\
\#\#\#\# 4.1 Result Analysis\\
- Are the model output results clear and as expected?\\
- Does the result provide sufficient analysis to explain the model's inference process?\\
- Are the model results interpretable and do they help in understanding the essence of the problem?\\
- Does the analysis provide clear conclusions and highlight the strengths and weaknesses of the model?\\
\\
**Scoring Criteria**:\\
1-2 = Completely unclear; 3-4 = Partially clear; 5-6 = Average; 7-8 = Clear; 9-10 = Very clear.\\
\\
\#\#\#\# 4.2 Bias Analysis\\
- Does the model identify and analyze potential biases?\\
- Does it consider data bias, model bias, and other factors?\\
- Does the model appropriately correct biases to reduce their impact on the results?\\
\\
**Scoring Criteria**:\\
1-2 = Completely ignored biases; 3-4 = Partially considered biases; 5-6 = Average; 7-8 = Considered biases and corrected; 9-10 = Very thorough, biases effectively corrected.\\
\\
**Output Format**:\\
Example 1:\\
\#\#\# 4.1 Result Analysis\textbackslash n\textbackslash n**Evaluation:**\textbackslash n\textbackslash nThe model output results are clear and well explain the model's inference process. The modeler has detailed the background and significance of the model results, helping to understand the core of the problem. The results show a reasonable inference path, making the entire analysis process more transparent. The analysis also provides clear conclusions and highlights the strengths and weaknesses of the model.\\
**Score:**\textbackslash n<reason> The result analysis is very clear and effectively supports decision-making </reason> \textbackslash n<score> 9 </score>  \\
\\
\#\#\# 4.2 Bias Analysis\textbackslash n\textbackslash n**Evaluation:**\textbackslash n\textbackslash nThe model effectively identifies and analyzes biases, particularly potential data biases. The modeler provides correction measures for biases and explains how these corrections affect the model results. Although there are still some biases in certain aspects of the model, overall, a comprehensive correction has been made.\\
**Score:**\textbackslash n<reason> The bias analysis is thorough, and biases have been effectively corrected </reason> \textbackslash n<score> 8 </score>\\
\\
Please objectively and detailedly evaluate the result analysis and bias analysis of the modeling according to the above evaluation criteria, and provide the final score and reason.\\
\#\#\# 4.1 Result Analysis\textbackslash n\textbackslash n**Evaluation:
\end{tcolorbox}
\noindent\begin{minipage}{\linewidth}
\captionof{figure}{The prompt used for evaluating Result and Bias Analysis of Agent.}
\label{prompt:RBAEP}
\end{minipage}

\begin{tcolorbox}[colframe=blue!20!gray, colback=blue!5!white, coltitle=white, fonttitle=\bfseries, title=Data Description Prompt, breakable]
Data Description:\\
\{data\_description\}\\
\{variable\_description\}
\\ \\
---
\\ \\
Your task is to generate a detailed summary of the dataset based on the dataset description provided. It needs to cover comprehensive information, but not explain each field one by one. Using plain text to describe in a single paragraph, without any Markdown formatting or syntax.
\end{tcolorbox}
\noindent\begin{minipage}{\linewidth}
\captionof{figure}{The prompt used to describe the dataset.}
\label{prompt:DDP}
\end{minipage}

\begin{tcolorbox}[colframe=blue!20!gray, colback=blue!5!white, coltitle=white, fonttitle=\bfseries, title=Problem Template, breakable]
Problem Background: \\
\{problem\_background\}
\\ \\
Problem Requirement: \\
\{problem\_requirement\} \\
\{addendum\} \\ \\
Dataset Path:\\
\{dataset\_path\}\\ \\
Data Description:\\
\{data\_summary\}
\end{tcolorbox}
\noindent\begin{minipage}{\linewidth}
\captionof{figure}{The Problem Template prompt.}
\label{prompt:PT}
\end{minipage}

\begin{tcolorbox}[colframe=blue!20!gray, colback=blue!5!white, coltitle=white, fonttitle=\bfseries, title=Problem Understanding Prompt, breakable]
\# Mathematical Modeling Problem: \\
\{modeling\_problem\}
\\ \\
---
\\ \\
You are tasked with analyzing a mathematical modeling problem with a focus on the underlying concepts, logical reasoning, and assumptions that inform the solution process. Begin by considering the nature of the problem in its broader context. What are the primary objectives of the model, and how do they shape the way you approach the task? Think critically about the assumptions that may be inherently embedded in the problem. What implicit beliefs or constraints have been set up, either explicitly or implicitly, within the problem’s description? Reflect on how these assumptions might influence the interpretation and application of any potential solutions. 
\\ \\
Dive deeper into the relationships and interdependencies between the different components of the problem. What are the potential hidden complexities that may arise from these interconnections? Are there any conflicts or tensions between different aspects of the problem that need to be resolved? Explore how these interdependencies might lead to unforeseen challenges and require revisiting initial assumptions or redefining the parameters of the task. 
\\ \\
Consider how the complexity of the problem may evolve across different scales or over time. Are there time-dependent factors or long-term consequences that should be accounted for, especially in terms of the stability or sustainability of the model’s outcomes? Think about how the model’s behavior might change under different scenarios, such as variations in input or changes in external conditions. Reflect on whether any simplifications or idealizations in the problem might inadvertently obscure key dynamics that are crucial for an accurate representation.
\\ \\
In your analysis, also give attention to possible alternative perspectives on the problem. Are there different ways to frame the issue that could lead to distinct modeling approaches or solution strategies? How would those alternative perspectives impact the overall approach? Additionally, evaluate the potential risks or uncertainties inherent in the problem, especially when it comes to choosing between competing modeling approaches. Consider how the outcomes might vary depending on the choices you make in constructing the model, and how you would manage such trade-offs.
\\ \\
Finally, reflect on the dynamic nature of the modeling process itself. How might your understanding of the problem evolve as you continue to explore its intricacies? Ensure that your thought process remains flexible, with a readiness to revise earlier conclusions as new insights emerge. The goal is to maintain a reflective, iterative analysis that adapts to deeper understandings of the task at hand, rather than pursuing a fixed or rigid approach.
\\ \\
\{user\_prompt\}
\\ \\
Respond as comprehensively and in as much detail as possible. Do not format your response in Markdown. Using plain text, without any Markdown formatting or syntax. Written as one or more cohesive paragraphs. Avoid structuring your answer in bullet points or numbered lists.
\end{tcolorbox}
\noindent\begin{minipage}{\linewidth}
\captionof{figure}{The prompt used in the Problem Understanding step. It guides the agent to perform a deep conceptual and contextual analysis of the modeling task, encouraging reflection on assumptions, interdependencies, temporal dynamics, uncertainties, and alternative perspectives to support rigorous and adaptive problem framing.}
\label{prompt:PUP}
\end{minipage}

\begin{tcolorbox}[colframe=blue!20!gray, colback=blue!5!white, coltitle=white, fonttitle=\bfseries, title=Problem Understanding Critique Prompt, breakable]
\# Mathematical Modeling Problem:\\
\{modeling\_problem\}
\\ \\
\# Problem Analysis:\\
\{problem\_analysis\}
\\ \\
---
\\ \\
Critically examine the analysis results of the given mathematical modeling problem, focusing on the following aspects:
\\ \\
1. Depth of Thinking: Evaluate whether the analysis demonstrates a comprehensive understanding of the underlying problem. Does it go beyond surface-level observations? Are the assumptions, limitations, and potential implications of the results carefully considered? Assess whether the analysis adequately addresses both the broader context and specific intricacies of the problem.\\
2. Novelty of Perspective: Analyze the originality of the approach taken in the analysis. Does it introduce new insights or merely rehash well-established methods or solutions? Are alternative perspectives or unconventional techniques explored, or is the analysis constrained by a narrow set of assumptions or typical approaches?\\
3. Critical Evaluation of Results: Consider the extent to which the analysis critically engages with the results. Are the conclusions drawn from the analysis well-supported by the mathematical findings, or do they overlook key uncertainties or counterexamples? Does the analysis acknowledge potential contradictions or ambiguities in the data?\\
4. Rigor and Precision: Assess the level of rigor applied in the analysis. Are the steps logically consistent and mathematically sound, or are there overlooked errors, gaps, or assumptions that undermine the conclusions? Does the analysis exhibit a clear, methodical approach, or is it characterized by vague reasoning and imprecision?\\
5. Contextual Awareness: Evaluate how well the analysis situates itself within the broader landscape of mathematical modeling in this area. Does it consider previous work or developments in the field? Is there any indication of awareness of real-world implications, practical constraints, or ethical concerns, if applicable?
\\ \\
Critique the analysis without offering any constructive suggestions—your focus should solely be on highlighting weaknesses, gaps, and limitations within the approach and its execution.
\end{tcolorbox}
\noindent\begin{minipage}{\linewidth}
\captionof{figure}{The prompt used for criticizing problem analysis in the Problem Understanding step. It prompts the agent to conduct a focused critique of the initial analysis by evaluating its depth, originality, logical rigor, and contextual awareness, helping identify gaps and limitations without providing corrective suggestions.}
\label{prompt:PUCP}
\end{minipage}

\begin{tcolorbox}[colframe=blue!20!gray, colback=blue!5!white, coltitle=white, fonttitle=\bfseries, title=Problem Understanding Improvement Prompt, breakable]
\# Mathematical Modeling Problem: \\
\{modeling\_problem\}\\
\\
\# Problem Analysis:\\
\{problem\_analysis\}\\
\\
\# Problem Analysis Critique:\\
\{problem\_analysis\_critique\}\\
\\
---
\\ \\ 
Refine and improve the existing problem analysis based on the critique provided to generate insightful analysis. 
\\ \\
Provide the improved version directly. DO NOT mention any previous analysis content and deficiencies in the improved analysis. Just refer to the above critical suggestions and directly give the new improved analysis.\\
\{user\_prompt\}\\
Respond as comprehensively and in as much detail as possible. Do not format your response in Markdown. Using plain text, without any Markdown formatting or syntax. Written as one or more cohesive paragraphs. Avoid structuring your answer in bullet points or numbered lists.
\\ \\
IMPROVED PROBLEM ANALYSIS:
\end{tcolorbox}
\noindent\begin{minipage}{\linewidth}
\captionof{figure}{The prompt used for improving problem analysis in the Problem Understanding step. It guides the agent to revise its initial analysis by incorporating critical feedback, enabling more rigorous, insightful, and context-aware problem understanding through iterative refinement.}
\label{prompt:PUIP}
\end{minipage}

\begin{tcolorbox}[colframe=blue!20!gray, colback=blue!5!white, coltitle=white, fonttitle=\bfseries, title=Task Decompose Prompt, breakable]
\# Decompose Principle:\\
\{decomposed\_principle\}\\
\\
\# Mathematical Modeling Problem:\\
\{modeling\_problem\}\\
\\
\# Problem Analysis:\\
\{problem\_analysis\}\\
\\
\# Modeling Solution:\\
\{modeling\_solution\}
\\ \\
---
\\ \\
Please decompose the given modeling solution into \{tasknum\} distinct and well-defined subtasks that collectively contribute to the overall objective. These subtasks should be clearly separated in their focus, each addressing a specific aspect of the modeling process. The goal is to break down the solution into key stages or methodologies, ensuring that all components of the solution are covered without redundancy. For each subtask, the approach or technique should be explicitly described, detailing the specific data, algorithms, or models required. The decomposition should reflect a logical and comprehensive path toward completing the task, with each part having a clear purpose and contributing to the final result.\\
\{user\_prompt\}\\
Each subtask should be described as comprehensively and in as much detail as possible within a single paragraph using plain text and seperated by '---' for each subtask. All the contents and details of the original solution need to be covered by the \{tasknum\} subtasks without omission. 
\end{tcolorbox}
\noindent\begin{minipage}{\linewidth}
\captionof{figure}{The prompt used in the Problem Decomposition step. It instructs the agent to transform a holistic modeling solution into a coherent set of structured subtasks, each with distinct objectives, methods, and contributions to the overall problem, ensuring logical coverage and methodological clarity.}
\label{prompt:TDP}
\end{minipage}

\begin{tcolorbox}[colframe=blue!20!gray, colback=blue!5!white, coltitle=white, fonttitle=\bfseries, title=Task Description Prompt, breakable]
\# Mathematical Modeling Problem:\\
\{modeling\_problem\}
\\ \\
\# Problem Analysis:\\
\{problem\_analysis\}
\\ \\
\# Modeling Solution:\\
\{modeling\_solution\}
\\ \\
\# Decomposed Subtasks:\\
\{decomposed\_subtasks\}
\\ \\
---
\\ \\
You are tasked with refining and improving the description of subtask \{task\_i\} to ensure it is more detailed, clear, and focused. Provide a precise and comprehensive explanation of the task, specifically elaborating on its scope, goals, and methodology without venturing into other subtasks. Make sure the description includes clear and concise language that defines the necessary steps, techniques, or approaches required for this subtask. If applicable, specify the data inputs, tools, or models to be used, but do not introduce analysis, results, or discussions related to other components of the modeling process. The goal is to enhance the clarity, depth, and precision of this subtask description, ensuring it is fully understood on its own without needing further explanation.\\
The description of subtask \{task\_i\} should be as comprehensive and in as much detail as possible within a single paragraph using plain text.
\end{tcolorbox}
\noindent\begin{minipage}{\linewidth}
\captionof{figure}{The prompt used for refining and improving task descriptions. It guides the agent to produce a precise, self-contained, and detailed explanation of a specific subtask, clarifying its scope, objectives, methodology, and required resources while avoiding overlap with other components.}
\label{prompt:TDeP}
\end{minipage}

\begin{tcolorbox}[colframe=blue!20!gray, colback=blue!5!white, coltitle=white, fonttitle=\bfseries, title=Task Dependency Analysis Prompt, breakable]
Understanding the dependencies among different tasks in a mathematical modeling process is crucial for ensuring a coherent, logically structured, and efficient solution. Given a mathematical modeling problem and its solution decomposition into \{tasknum\} subtasks, analyze the interdependencies among these subtasks.  
\\ \\
\#\# Input Information:\\
- **Mathematical Modeling Problem:** \{modeling\_problem\}\\
- **Problem Analysis:** \{problem\_analysis\}\\
- **Modeling Solution:** \{modeling\_solution\}\\
- **Decomposed Tasks:** \{task\_descriptions\}
\\ \\
\#\# Task Dependency Analysis Instructions:\\
1. **Identify Task Dependencies:** For each task, determine which preceding tasks provide necessary input, data, or conditions for its execution. Clearly outline how earlier tasks influence or constrain later ones.\\
2. **Describe Dependency Types:** Specify the nature of the dependencies between tasks. This includes:\\
\text{\ \ \ }- *Data Dependency:* When one task produces outputs that are required as inputs for another task.\\
\text{\ \ \ }- *Methodological Dependency:* When a later task builds upon a theoretical framework, assumptions, or models established by an earlier task.\\
\text{\ \ \ }- *Computational Dependency:* When a task requires prior computations or optimizations to be completed before proceeding.\\
\text{\ \ \ }- *Structural Dependency:* When a task is logically required to be completed before another due to hierarchical or sequential constraints.\\
\text{\ \ \ }- *Code Dependency:* When one task relies on code structures, functions, or modules that are defined or executed in a preceding task. This includes shared variables, functions, or libraries that must be defined before their use in later tasks.\\
3. **Ensure Completeness:** Verify that all tasks in the decomposition are accounted for in the dependency analysis and that no essential dependencies are missing.
\\ \\
\#\# Output Format: \\
Respond as comprehensively and in as much detail as possible. Do not format your response in Markdown. Using plain text, without any Markdown formatting or syntax. Written as {tasknum} cohesive paragraphs, each paragraph is a dependency analysis of a task.
\\ \\
The response should be comprehensive and written in a clear, well-structured format without bullet points, ensuring a logical flow of dependency relationships and their implications.
\end{tcolorbox}
\noindent\begin{minipage}{\linewidth}
\captionof{figure}{The prompt used in the Task Dependency Analysis step. It guides the agent to identify and describe data, methodological, computational, and structural dependencies among subtasks, ensuring a coherent and executable modeling workflow.}
\label{prompt:TDAP}
\end{minipage}

\begin{tcolorbox}[colframe=blue!20!gray, colback=blue!5!white, coltitle=white, fonttitle=\bfseries, title=DAG Construction Prompt, breakable]
A well-structured Directed Acyclic Graph (DAG) is essential for visualizing and optimizing the dependencies between different tasks in a mathematical modeling process. Given a problem and its solution decomposition into {tasknum} subtasks, construct a DAG that accurately represents the dependency relationships among these tasks. The DAG should capture all necessary dependencies while ensuring that no cycles exist in the structure.
\\ \\
\#\# Input Information:\\
- **Mathematical Modeling Problem:** \{modeling\_problem\}\\
- **Problem Analysis:** \{problem\_analysis\}\\
- **Modeling Solution:** \{modeling\_solution\}\\
- **Decomposed Tasks:** \{task\_descriptions\}\\
- **Dependency Analysis:** \{task\_dependency\_analysis\}\\
\\ \\
\#\# Output Format (STRICT REQUIREMENT):\\
You **MUST** return a valid JSON-formatted adjacency list **without** any additional text, explanations, or comments. **Only** output the JSON object.
\\ \\
\#\#\# JSON Format (Strictly Follow This Format):\\
```json\\
\{\{\\
\text{\ \ }"task\_ID": [dependent\_IDs],\\
\text{\ \ }...\\
\}\}\\
\\
\#\# Example Output: \\
```json\\
\{\{\\
"1": []\\
"2": ['1']\\
"3": ['1']\\
"4": ['2', '3']\\
\}\}\\
```
\end{tcolorbox}
\noindent\begin{minipage}{\linewidth}
\captionof{figure}{The prompt used for constructing the Task Dependency Graph. It instructs the agent to generate a DAG in strict JSON format, capturing all task-level dependencies derived from prior analysis to enable structured visualization and execution planning.}
\label{prompt:DAGCP}
\end{minipage}

\begin{tcolorbox}[colframe=blue!20!gray, colback=blue!5!white, coltitle=white, fonttitle=\bfseries, title=Model Formulas Construction Prompt, breakable]
\# Reference Modeling Methods:\\
\{modeling\_methods\}
\\ \\
\{data\_summary\}
\\ \\
\# Task Description:\\
\{task\_description\}
\\ \\
\# Task Analysis:\\
\{task\_analysis\}
\\ \\
---\\
\# The structure of code for Task \{task\_id\}:\\
\{code\_structure\}\\ \\
\# The result for Task \{task\_id\}:\\
\{task\_result\}\\
---\\
When formulating the mathematical model for the current task, it is essential to consider how this task depends on other tasks in the overall process.
\\ \\
You are collaborating as part of a multi-agent system to solve a complex mathematical modeling problem. Each agent is responsible for a specific task, and some preprocessing or related tasks may have already been completed by other agents. It is crucial that you **do not repeat any steps that have already been addressed** by other agents. Instead, rely on their outputs when necessary and focus solely on the specific aspects of the task assigned to you.
\\ \\
You are tasked with developing a set of precise, insightful, and comprehensive mathematical formulas that effectively model the problem described in the task. Begin by conducting an in-depth analysis of the system, process, or phenomenon outlined, identifying all relevant variables, their interdependencies, and the fundamental principles, laws, or constraints that govern the behavior of the system, as applicable in the relevant field. Clearly define all variables, constants, and parameters, and explicitly state any assumptions, approximations, or simplifications made during the formulation process, including any boundary conditions or initial conditions if necessary.
\\ \\
Ensure the formulation considers the full scope of the problem, and if applicable, incorporate innovative mathematical techniques. Your approach should be well-suited for practical computational implementation, addressing potential numerical challenges, stability concerns, or limitations in simulations. Pay careful attention to the dimensional consistency and units of all terms to guarantee physical or conceptual validity, while remaining true to the theoretical foundations of the problem.
\\ \\
In the process of deriving the mathematical models, provide a clear, step-by-step explanation of the reasoning behind each formula, highlighting the derivation of key expressions and discussing any assumptions or trade-offs that are made. Identify any potential sources of uncertainty, limitations, or approximations inherent in the model, and provide guidance on how to handle these within the modeling framework.
\\ \\
The resulting equations should be both flexible and scalable, allowing for adaptation to different scenarios or the ability to be tested against experimental or real-world data. Strive to ensure that your model is not only rigorous but also interpretable, balancing complexity with practical applicability. List all modeling equations clearly in LaTeX format, ensuring proper mathematical notation and clarity of presentation. Aim for a model that is both theoretically sound and practically relevant, offering a balanced approach to complexity and tractability in its use.\\
\{user\_prompt\}\\
Respond as comprehensively and in as much detail as possible, ensuring clarity, depth, and rigor throughout. Using plain text and LaTeX for formulas. Written as one or more cohesive paragraphs. Avoid structuring your answer in bullet points or numbered lists.
\end{tcolorbox}
\noindent\begin{minipage}{\linewidth}
\captionof{figure}{The prompt used for constructing the model formulas. It instructs the agent to derive detailed, rigorous, and task-specific mathematical formulations by integrating prior task outputs, domain principles, and computational considerations, ensuring both theoretical soundness and practical applicability.}
\label{prompt:MFCP}
\end{minipage}

\begin{tcolorbox}[colframe=blue!20!gray, colback=blue!5!white, coltitle=white, fonttitle=\bfseries, title=Model Formulas Critique Prompt, breakable]
\{data\_summary\}
\\ \\
\# Task Description:\\
\{task\_description\}
\\ \\
\# Task Analysis:\\
\{task\_analysis\}
\\ \\
\# Task Modeling Formulas:\\
\{modeling\_formulas\}
\\ \\
---
\\ \\
The goal of this task is to critically evaluate the modeling formulas used to represent a given mathematical modeling problem. Your analysis should address the following dimensions: accuracy and rigor, innovation and insight, and the applicability of the models to real-world scenarios.
\\ \\
1. Accuracy and Rigor:
\\ \\
- Formula Integrity:\\
\text{\ \ }Evaluate whether the mathematical models and the corresponding formulas are mathematically sound and consistent with the underlying assumptions of the problem. Are the formulas properly derived, free from logical errors, and reflective of the relevant domain knowledge?\\
\text{\ \ }- Are any simplifications or approximations made, and if so, are they justifiable within the context of the model's scope?\\
\text{\ \ }- Examine the assumptions made in formulating the model. Are these assumptions realistic, and how do they affect the model’s precision and robustness?
\\ \\
2. Innovation and Insight:
\\ \\
- Novelty of Approach:\\
\text{\ \ }Critique the originality of the modeling approach. Does the model present a new or unconventional way of solving the problem, or does it simply rely on established methodologies without offering new insights?\\
\text{\ \ }- Consider whether any innovative methods, such as the introduction of novel variables or the use of innovative computational techniques, contribute to improving the model.
\\ \\
- Theoretical Insight:\\
\text{\ \ }Evaluate the depth of the theoretical insights provided by the model. Does it offer a fresh perspective or new understanding of the problem? How well does it illuminate the key dynamics and relationships within the system under study?\\
\text{\ \ }- Does the model reveal previously unnoticed phenomena, or does it suggest new directions for further research?
\\ \\
- Integration of Existing Knowledge:\\
\text{\ \ }Assess the extent to which the model integrates existing mathematical, theoretical, and empirical work. Does it build on prior research, and if so, does it do so in a way that adds substantial value or clarity? Are there gaps where additional cross-disciplinary knowledge could enhance the model?
\\ \\
---
\\ \\
3. Applicable:
\\ \\
- Real-World Relevance:\\
\text{\ \ }Evaluate the model’s practical applicability. How well does it apply to real-world problems, and to what extent does it provide actionable insights for decision-making or problem-solving in the field?  
\\ \\
Critique the analysis without offering any constructive suggestions—your focus should solely be on highlighting weaknesses, gaps, and limitations within the formulas.
\end{tcolorbox}
\noindent\begin{minipage}{\linewidth}
\captionof{figure}{The prompt used for criticizing the model formulas. It guides the agent to identify weaknesses in mathematical soundness, theoretical depth, and real-world applicability of the formulas, fostering rigorous evaluation without offering corrective suggestions.}
\label{prompt:MFCrP}
\end{minipage}

\begin{tcolorbox}[colframe=blue!20!gray, colback=blue!5!white, coltitle=white, fonttitle=\bfseries, title=Model Formulas Improvement Prompt, breakable]
\{data\_summary\}
\\ \\
\# Task Description:\\
\{task\_description\}
\\ \\
\# Task Analysis:\\
\{task\_analysis\}
\\ \\
\# Task Modeling Formulas:\\
\{modeling\_formulas\}
\\ \\
\# Task Modeling Formulas Critique:\\
\{modeling\_formulas\_critique\}
\\ \\
---
\\ \\
Based on the provided critique and analysis, refine the existing modeling formulas to address the identified limitations and gaps. 
\\ \\
Respond as comprehensively and in as much detail as possible, ensuring clarity, depth, and rigor throughout. Using plain text and LaTeX for formulas. Written as one or more cohesive paragraphs. Avoid structuring your answer in bullet points or numbered lists.\\
\{user\_prompt\}
Provide a new version of the task modeling formulas that integrates these improvements directly. DO NOT mention any previous formulas content and deficiencies.
\\ \\
IMPROVED TASK MODELING FORMULAS:
\end{tcolorbox}
\noindent\begin{minipage}{\linewidth}
\captionof{figure}{The prompt used for improving the model formulas.}
\label{prompt:MFIP}
\end{minipage}

\begin{tcolorbox}[colframe=blue!20!gray, colback=blue!5!white, coltitle=white, fonttitle=\bfseries, title=Model Construction Prompt, breakable]
\{data\_summary\}
\\ \\
\# Task Description:\\
\{task\_description\}
\\ \\
\# Task Analysis:\\
\{task\_analysis\}
\\ \\
\# Task Modeling Formulas:\\
\{modeling\_formulas\}
\\ \\
---\\
\# The structure of code for Task \{task\_id\}:\\
\{code\_structure\}\\ \\
\# The result for Task \{task\_id\}:\\
\{task\_result\}\\
---\\
Please consider the dependencies between the current task and the preceding tasks.
\\ \\
You are collaborating as part of a multi-agent system to solve a complex mathematical modeling problem. Each agent is responsible for a specific task, and some preprocessing or related tasks may have already been completed by other agents. It is crucial that you **do not repeat any steps that have already been addressed** by other agents. Instead, rely on their outputs when necessary and focus solely on the specific aspects of the task assigned to you.
\\ \\
Please continue the modeling formula section by building upon the previous introduction to the formula. Provide comprehensive and detailed explanations and instructions that elaborate on each component of the formula. Describe the modeling process thoroughly, including the underlying assumptions, step-by-step derivations, and any necessary instructions for application. Expand on the formula by incorporating relevant mathematical expressions where appropriate, ensuring that each addition enhances the reader’s understanding of the model. Make sure to seamlessly integrate the new content with the existing section, maintaining a natural flow and avoiding any repetition or conflicts with previously covered material. Your continuation should offer a clear and in-depth exploration of the modeling formula, providing all necessary details to facilitate a complete and coherent understanding of the modeling process.\\
\{user\_prompt\}\\
Respond as comprehensively and in as much detail as possible. Do not format your response in Markdown. Using plain text, without any Markdown formatting or syntax. Written as one or more cohesive paragraphs. Avoid structuring your answer in bullet points or numbered lists.
\end{tcolorbox}
\noindent\begin{minipage}{\linewidth}
\captionof{figure}{The prompt used for improving the model formulas. It instructs the agent to revise existing formulas by directly integrating feedback from prior critique, ensuring the final formulation is more rigorous, complete, and aligned with problem-specific constraints.}
\label{prompt:MCP}
\end{minipage}

\begin{tcolorbox}[colframe=blue!20!gray, colback=blue!5!white, coltitle=white, fonttitle=\bfseries, title=Code Generation Prompt, breakable]
\# Dataset Path:\\
\{dataset\_path\}\\
\\
\# Data Description:\\
\{data\_summary\}\\
\\
\# Variable Description:\\
\{variable\_description\}\\
\\
\# Other files (Generated by Other Agents):\\
\{dependent\_file\_prompt\}\\
\\
\# Task Description:\\
\{task\_description\}\\
\\
\# Task Analysis:\\
\{task\_analysis\}\\
\\
\# Task Modeling Formulas:\\
\{modeling\_formulas\}\\
\\
\# Task Modeling Process:\\
\{modeling\_process\}\\
\\
\# Code Template:\\
\{code\_template\}\\
\\
---\\
\\
\#\# Role \& Collaboration:\\
You are an expert programmer working as part of a multi-agent system. Your role is to implement the code based on the provided dataset (**refer to the Dataset Path, Dataset Description, and Variable Description**) **or preprocessed files generated by other agents** (**refer to "Other Files"**), along with the modeling process and given code template. Other agents will use your results to make decisions, but they will **not** review your code. Therefore, it is crucial that:\\
1. **Ensure the code is executable** and will successfully run without errors, producing the expected results. **It should be tested to verify it works in the intended environment**.\\
2. **Reuse files from "Other Files" whenever possible** instead of redoing tasks that have already been completed by other agents.\\
3. **All data processing steps must save the processed results to local files (CSV, JSON, or pickle) for easy access by other agents.**\\
4. **The output should be as detailed as possible**, including intermediate results and final outputs.\\
5. **Ensure transparency** by logging key computation steps and providing clear outputs.\\

\#\# Implementation Guidelines:\\
- **Prioritize using files from "Other Files" before processing raw data** to avoid redundant computation.\\
- Follow the provided **modeling formulas** and **modeling process** precisely.\\
- The **code must be executable**: ensure that the Python code you generate runs without errors. Do not just focus on producing the correct output format; **focus on producing a working solution** that can be executed successfully in a Python environment.\\
- **Store intermediate and final data processing results to local** in appropriate formats (e.g., CSV, JSON, or pickle).\\
- Provide **detailed print/logging outputs** to ensure that other agents can understand the results without needing to read the code.\\
\{user\_prompt\}\\
\\
\#\# Expected Response Format:\\
You **MUST** return the Python implementation in the following format:\\
```python\\
\# Here is the Python code.\\
```
\end{tcolorbox}
\noindent\begin{minipage}{\linewidth}
\captionof{figure}{The prompt used for generating code. It instructs the agent to produce fully executable Python implementations aligned with prior modeling outputs, while ensuring correctness, reproducibility, and interoperability within a multi-agent system through structured input, logging, and file-based output handling.}
\label{prompt:CGP}
\end{minipage}

\begin{tcolorbox}[colframe=blue!20!gray, colback=blue!5!white, coltitle=white, fonttitle=\bfseries, title=Code Debugging Prompt, breakable]
\# Code Template:\\
\{code\_template\}\\
\\
\# Modeling Process:\\
\{modeling\_process\}\\
\\
\# Current Code:\\
\{code\}\\
\\
However, there are some bugs in this version. Here is the execution result:\\
\# Execution Result:\\
\{observation\}\\
\\
---\\
\\
You are a helpful programming expert. Based on the provided execution result, please revise the script to fix these bugs. Your task is to address the error indicated in the result, and refine or modify the code as needed to ensure it works correctly.\\
\{user\_prompt\}\\
Please respond exactly in the following format:\\
```python\\
\# Provide the corrected python code here.\\
```
\end{tcolorbox}
\noindent\begin{minipage}{\linewidth}
\captionof{figure}{The prompt used for debugging code. It instructs the agent to identify and fix execution errors based on observed outputs, ensuring that the corrected Python script is functional, aligned with the modeling process, and ready for downstream use.}
\label{prompt:CDP}
\end{minipage}

\begin{tcolorbox}[colframe=blue!20!gray, colback=blue!5!white, coltitle=white, fonttitle=\bfseries, title=Code Structure Extraction Prompt, breakable]
You are a programming expert. Please extract the structure from the following code and output it in the following JSON format, please return an empty list if the corresponding item is not available.:\\
The code is:\\
```python\\
\{code\}\\
```\\
The output format is:\\
```json\\
\{\{\\
\text{\ \ \ \ }"script\_path": \{save\_path\}\\
\text{\ \ \ \ }"class": [\\
\text{\ \ \ \ }\{\{\\
\text{\ \ \ \ \ \ }"name": class name,\\
\text{\ \ \ \ \ \ }"description": description of class,\\
\text{\ \ \ \ \ \ }"class\_functions": [\\
\text{\ \ \ \ \ \ \ \ }\{\{\\
\text{\ \ \ \ \ \ \ \ \ \ }"name": function name,\\
\text{\ \ \ \ \ \ \ \ \ \ }"description": description of class function,\\
\text{\ \ \ \ \ \ \ \ \ \ }"parameters": [\\
\text{\ \ \ \ \ \ \ \ \ \ \ \ }\{\{\\
\text{\ \ \ \ \ \ \ \ \ \ \ \ \ \ }"name": param name,\\
\text{\ \ \ \ \ \ \ \ \ \ \ \ \ \ }"type": param type,\\
\text{\ \ \ \ \ \ \ \ \ \ \ \ \ \ }"description": description of param,\\
\text{\ \ \ \ \ \ \ \ \ \ \ \ }\}\},\\
\text{\ \ \ \ \ \ \ \ \ \ \ \ }...\\
\text{\ \ \ \ \ \ \ \ \ \ }],\\
\text{\ \ \ \ \ \ \ \ \ \ }"returns": \{\{\\
\text{\ \ \ \ \ \ \ \ \ \ \ \ }"description": "return of the function."\\
\text{\ \ \ \ \ \ \ \ \ \ }\}\},\\
\text{\ \ \ \ \ \ \ \ }\}\}\\
\text{\ \ \ \ \ \ }]\\
\text{\ \ \ \ }\}\}\\
\text{\ \ }],\\
\text{\ \ }"function": [\\
\text{\ \ \ \ }\{\{\\
\text{\ \ \ \ \ \ }"name": function name,\\
\text{\ \ \ \ \ \ }"description": description of class function,\\
\text{\ \ \ \ \ \ }"parameters": [\\
\text{\ \ \ \ \ \ \ \ }\{\{\\
\text{\ \ \ \ \ \ \ \ \ \ }"name": param name,\\
\text{\ \ \ \ \ \ \ \ \ \ }"type": param type,\\
\text{\ \ \ \ \ \ \ \ \ \ }"description": description of param,\\
\text{\ \ \ \ \ \ \ \ }\}\},\\
\text{\ \ \ \ \ \ \ \ }...\\
\text{\ \ \ \ \ \ }],\\
\text{\ \ \ \ \ \ }"returns": \{\{\\
\text{\ \ \ \ \ \ \ \ }"description": "return of the function."\\
\text{\ \ \ \ \ \ }\}\},\\
\text{\ \ \ \ }\}\}\\
\text{\ \ }],\\
\text{\ \ }"file\_outputs": [\\
\text{\ \ \ \ }\{\{\\
\text{\ \ \ \ \ \ }"path": "file\_path",\\
\text{\ \ \ \ \ \ }"file\_description": "description of the file",\\
\text{\ \ \ \ \ \ }"column\_name": ["column\_name\_if\_csv\_else\_None"]\\
\text{\ \ \ \ }\}\},\\
\text{\ \ \ \ }...\\
\text{\ \ }]\\
\}\}\\
```
\end{tcolorbox}
\noindent\begin{minipage}{\linewidth}
\captionof{figure}{The prompt used for extracting the structure of code.}
\label{prompt:CSEP}
\end{minipage}

\begin{tcolorbox}[colframe=blue!20!gray, colback=blue!5!white, coltitle=white, fonttitle=\bfseries, title=Result Interpretation Prompt, breakable]
\# Task Description:\\
\{task\_description\}\\
\\
\# Task Analysis:\\
\{task\_analysis\}\\
\\
\# Task Modeling Formulas:\\
\{task\_formulas\}\\
\\
\# Task Modeling:\\
\{task\_modeling\}\\
\\
\# Code Execution Result:\\
\{execution\_result\}\\
\\
---\\
\\
Based on the task description, analysis, modeling framework, and code execution result, present a comprehensive and detailed account of the intermediate results, calculations, and outcomes generated during the task. Clearly articulate the results of any computations or operations performed, providing numerical values, data trends, or statistical measures as necessary. If visual representations such as graphs, charts, or tables were used to communicate the results, ensure they are clearly labeled and explained, highlighting their relevance to the overall task. Discuss the intermediate steps or processes that led to the results, including any transformations or assumptions made during calculations. If applicable, compare and contrast these results with expected outcomes or previously known results to gauge the task’s success. Provide a thoughtful interpretation of the findings, considering how they contribute to advancing understanding or solving the problem at hand, and highlight any areas where further investigation or refinement may be needed.\\
\{user\_prompt\}\\
Respond as comprehensively and in as much detail as possible. Do not format your response in Markdown. Using plain text and LaTeX for formulas only, without any Markdown formatting or syntax. Written as one or more cohesive paragraphs. Avoid structuring your answer in bullet points or numbered lists.
\end{tcolorbox}
\noindent\begin{minipage}{\linewidth}
\captionof{figure}{The prompt used for extracting the structure of code. It guides the agent to parse and represent the structural elements of the code, including functions, classes, parameters, and output files, in a standardized JSON format to support traceability, reuse, and documentation in downstream tasks.}
\label{prompt:RIP}
\end{minipage}

\begin{tcolorbox}[colframe=blue!20!gray, colback=blue!5!white, coltitle=white, fonttitle=\bfseries, title=Solution Formulation Prompt, breakable]
\# Task Description:\\
\{task\_description\}\\
\\
\# Task Analysis:\\
\{task\_analysis\}\\
\\
\# Task Modeling Formulas:\\
\{task\_formulas\}\\
\\
\# Task Modeling:\\
\{task\_modeling\}\\
\\
\# Task Result:\\
\{task\_result\}\\
\\
---\\
\\
Craft a comprehensive and insightful answer section that synthesizes the findings presented in the results section to directly address the research questions and objectives outlined at the outset of the study. Begin by clearly stating the primary conclusions drawn from the analysis, ensuring that each conclusion is explicitly linked to specific aspects of the results. Discuss how these conclusions validate or challenge the initial hypotheses or theoretical expectations, providing a coherent narrative that illustrates the progression from data to insight.\\
\\
Evaluate the effectiveness and reliability of the mathematical models employed, highlighting strengths such as predictive accuracy, robustness, or computational efficiency. Address any limitations encountered during the modeling process, explaining how they may impact the validity of the conclusions and suggesting potential remedies or alternative approaches. Consider the sensitivity of the model to various parameters and the extent to which the results are generalizable to other contexts or applications.\\
\\
Analyze potential biases that may have influenced the results, including data bias, model bias, and computational bias. Discuss whether the dataset is representative of the problem space and whether any imbalances, selection biases, or sampling limitations might have affected the conclusions. Examine modeling assumptions, parameter choices, and architectural constraints that could introduce systematic deviations in the results. Assess how numerical precision, algorithmic approximations, or implementation details might influence the stability and fairness of the model’s predictions.\\
\\
Discuss strategies to mitigate identified biases and improve the reliability of the conclusions. Consider adjustments in data preprocessing, such as resampling, normalization, or augmentation, to address distribution imbalances. Explore refinements to the modeling process, including regularization techniques, fairness constraints, and sensitivity analyses, to ensure robustness across different scenarios. Evaluate the impact of alternative modeling approaches and discuss the extent to which the proposed methods can generalize beyond the given dataset or problem context.\\
\\
Explore the broader implications of the findings for the field of study, identifying how they contribute to existing knowledge, inform future research directions, or influence practical applications. Discuss any unexpected outcomes and their significance, offering interpretations that may reveal new avenues for exploration or theoretical development. Reflect on the societal, economic, or environmental relevance of the results, if applicable, and propose recommendations based on the study’s insights.\\
\\
Conclude the section by summarizing the key takeaways, emphasizing the contribution of the research to solving the problem at hand, and outlining the next steps for further investigation or implementation. Ensure that the discussion is logically structured, with each paragraph building upon the previous ones to form a cohesive and persuasive argument that underscores the study’s value and impact.\\
\\
The content of this Task Answer section should be distinct and not merely a repetition of the Task Result section. Ensure that there is no duplication.\\
\\
\{user\_prompt\}\\
\\
Respond as comprehensively and in as much detail as possible. Do not format your response in Markdown. Using plain text and LaTeX for formulas only, without any Markdown formatting or syntax. Written as one or more cohesive paragraphs. Avoid structuring your answer in bullet points or numbered lists.
\end{tcolorbox}
\noindent\begin{minipage}{\linewidth}
\captionof{figure}{The prompt used for formulating the solution. It guides the agent to synthesize modeling results into a coherent, bias-aware, and insight-driven conclusion that addresses research objectives, evaluates model reliability, and reflects on broader implications.}
\label{prompt:SFP}
\end{minipage}

\begin{tcolorbox}[colframe=blue!20!gray, colback=blue!5!white, coltitle=white, fonttitle=\bfseries, title=Chart Guidelines Generation Prompt, breakable]
\#\# Instruction\\
Create a highly detailed and comprehensive chart that effectively visualizes the complex mathematical relationships and insights presented in the provided mathematical modeling paper. Begin by selecting the most appropriate type of chart—such as a line graph, bar chart, scatter plot, heatmap, or 3D surface plot—based on the nature of the data and the specific relationships being analyzed. Clearly define the variables involved, including their units and scales, and incorporate any derived metrics that enhance interpretability. Ensure that the axes are labeled accurately and descriptively, with appropriate units and scales, whether linear or logarithmic, to best represent the data distribution and relationships. Include a clear and concise legend that distinguishes between different datasets or variables, using distinct colors or patterns that are both aesthetically pleasing and easily distinguishable. Utilize gridlines to aid in the accurate reading of values, and choose a color scheme that enhances readability while maintaining visual appeal.\\
\\
Emphasize the core purpose of the chart, whether it is to highlight trends over time, compare different values, show distributions, illustrate correlations, validate theoretical models, or support key arguments within the paper. Articulate the intended message of the chart clearly, ensuring that every design choice—from the type of chart to the specific visual elements used—aligns with the objectives of the mathematical modeling paper. Incorporate multiple lines or bars if comparing different datasets, use shading or contouring for density representation, and add error bars to indicate uncertainty where applicable. Include annotations to highlight significant data points, trends, or anomalies that are critical to the analysis, providing context and explanations that guide the viewer’s understanding.\\
\\
Balance aesthetics with functionality by selecting colors and contrasts that not only make the chart visually compelling but also enhance readability and comprehension. Avoid unnecessary complexity by keeping the design clean and focused, ensuring that the chart remains clear and easy to interpret without sacrificing accuracy or depth of information. If beneficial, incorporate supplementary visual aids such as trend lines, regression curves, or overlays of empirical and theoretical results to strengthen the analysis and provide additional layers of insight. The final chart should serve as a precise and compelling visualization that effectively conveys the mathematical insights, facilitates understanding, and robustly supports the overall narrative and conclusions of the mathematical modeling paper.\\
\\
\{user\_prompt\}\\
\\
\#\# Paper Content\\
<paper>  \\
\{paper\_content\}  \\
</paper>  \\
\\
\#\# Existing Charts\\
\{existing\_charts\}  \\
\\
\#\# Create a New Chart\\
\\
Please create a chart that aligns closely with the above paper content while avoiding redundancy with existing charts. Follow the markdown format below to describe your chart:  \\
\\
**Chart Title**  \\
$[$Provide a clear and descriptive title for the chart$]$  \\
\\
**Chart Type**  \\
$[$Specify the type of chart$]$  \\
\\
**Purpose**  \\
$[$Describe the core purpose of the chart in a paragraph$]$  \\
\\
**Data or Variables**  \\
$[$Describe the data or variables used in the chart in a paragraph$]$  \\
\\
**Chart Presentation Guidelines**  \\
$[$A comprehensive guide on chart presentation, covering data representation, key layout elements, units, axis labels, legends, gridlines, annotations, and other essential considerations for effective visualization.$]$\\
\\
**Intended Message**  \\
$[$Articulate the key message or insight the chart is intended to convey in a paragraph$]$
\end{tcolorbox}
\noindent\begin{minipage}{\linewidth}
\captionof{figure}{The prompt used for generating chart creation guidelines. It instructs the agent to design detailed, purpose-driven visualizations that align with the mathematical insights of the paper, specifying data, layout, annotation, and interpretability considerations to ensure clarity, relevance, and analytical depth.}
\label{prompt:CGGP}
\end{minipage}

\begin{tcolorbox}[colframe=blue!20!gray, colback=blue!5!white, coltitle=white, fonttitle=\bfseries, title=Paper Chapter Creation Prompt, breakable]
You are tasked with creating a publication-quality LaTeX chapter for a mathematical modeling research paper. Carefully transform the provided structured draft into a coherent, rigorous, and concise narrative chapter that aligns logically and seamlessly with the previously written content.\\
\\
\#\# Target Chapter:\\
\{chapter\_path\}\\
\\
\#\# Structured Draft:\\
<structured\_draft>\\
\{json\_context\}\\
</structured\_draft>\\
\\
\#\# Preceding Chapters (for seamless narrative integration and avoiding repetition):\\
<preceding\_content>\\
\{previous\_chapters\}\\
</preceding\_content>\\
\\
\#\# Requirements:\\
- Write exclusively in accurate, idiomatic LaTeX; avoid Markdown syntax and symbols entirely.\\
- Clearly indicate the chapter content corresponds precisely to the target chapter `\{chapter\_path\}`; do not repeat or reference explicitly the content of other chapters.\\
- Integrate any mathematical formulas properly using correct LaTeX environments (`$\textbackslash$begin\{align\}`). Truncate and wrap long formulas and symbols.\\
- Present the chapter as a continuous, fluent narrative without section headings, subsections, bullet points, or numbered lists, Response only chapter content, do not include headlines and anything else.\\
- Critically evaluate the structured draft, selecting only most high-quality important and relevant content. Remove all redundancy, eliminate low-value statements, and distill essential information clearly and succinctly.\\
- Maintain rigorous academic style, logical coherence, and clarity throughout, ensuring that the chapter integrates naturally with preceding chapters.\\
\\
\#\# Output Format:\\
```latex\\
CHAPTER\_CONTENT\_TEXT\\
```
\end{tcolorbox}
\noindent\begin{minipage}{\linewidth}
\captionof{figure}{The prompt used for creating a paper chapter. It guides the agent to transform a structured draft into a fluent, publication-ready LaTeX chapter that integrates rigorously with preceding content while ensuring clarity, conciseness, and academic coherence.}
\label{prompt:PCCP}
\end{minipage}

\begin{tcolorbox}[colframe=blue!20!gray, colback=blue!5!white, coltitle=white, fonttitle=\bfseries, title=Paper Chapter Creation with Preceding Prompt, breakable]
You are tasked with generating a publication-quality LaTeX chapter for a mathematical modeling paper. Write a cohesive, academically rigorous chapter that integrates seamlessly with the preceding content of the paper.\\
\\
\#\# Chapter to write: \\
\{chapter\_path\}\\
\\
\#\# Preceding Content: \\
<preceding\_content>\\
\{previous\_chapters\}\\
<preceding\_content>\\
\\
\#\# Writing Requirements:\\
- Use accurate and proper LaTeX syntax throughout, avoid all Markdown syntax or symbols.\\
- Present the content as a continuous, coherent narrative without using sections, subsections, or bullet points. Response only chapter content, do not include headlines and anything else.\\
- Make it clear that the section you need to write is `\{chapter\_path\}`. Do not involve the content of other chapters.
\end{tcolorbox}
\noindent\begin{minipage}{\linewidth}
\captionof{figure}{The prompt used for creating a paper chapter with preceding content. It guides the agent to generate a cohesive, LaTeX-formatted chapter that maintains academic rigor and continuity with prior sections, ensuring seamless narrative flow while adhering strictly to chapter boundaries.}
\label{prompt:PCCPP}
\end{minipage}

\begin{tcolorbox}[colframe=blue!20!gray, colback=blue!5!white, coltitle=white, fonttitle=\bfseries, title=Paper Notation Creation Prompt, breakable]
You are an AI assistant trained to extract and typeset the Notations table from a mathematical modeling paper in LaTeX format. Your task is to take the input paper and output a properly formatted LaTeX table displaying the notations used in the paper. \\
\\
1. Well-structured and easy to read.\\
2. Properly typeset for LaTeX documents.\\
3. Adaptive in size and position to fit neatly into any document.\\
4. Truncate and wrap long formulas, symbols and text in the table for better readability.\\
\\
<paper>\\
\{previous\_chapters\}\\
</paper>\\
\\
Exmple of Table Format:\\
```latex\\
$\textbackslash$begin\{table\}$[$H$]$\\
\text{\ \ \ \ }$\textbackslash$centering\\
\text{\ \ \ \ }$\textbackslash$renewcommand\{$\textbackslash$arraystretch\}\{1.3\}\\
\text{\ \ \ \ }$\textbackslash$begin\{tabular\}\{>\{$\textbackslash$raggedright$\textbackslash$arraybackslash\}p\{3cm\}> \{$\textbackslash$raggedright$\textbackslash$arraybackslash\}p\{11cm\}\}\\
\text{\ \ \ \ \ \ \ \ }$\textbackslash$toprule\\
\text{\ \ \ \ \ \ \ \ }$\textbackslash$textbf\{Notation\} \& $\textbackslash$textbf\{Description\} \\
\text{\ \ \ \ \ \ \ \ }$\textbackslash$midrule\\
\text{\ \ \ \ \ \ \ \ }( f(x) ) \& description... \\
\text{\ \ \ \ \ \ \ \ }$\textbackslash$bottomrule\\
\text{\ \ \ \ }$\textbackslash$end\{tabular\}\\
\text{\ \ \ \ }$\textbackslash$caption\{Table of Notations\}\\
\text{\ \ \ \ }$\textbackslash$label\{tab:notations\}\\
$\textbackslash$end\{table\}\\
```\\
\\
Response only latex table content, do not include headlines and anything else.
\end{tcolorbox}
\noindent\begin{minipage}{\linewidth}
\captionof{figure}{The prompt used for creating paper notations. It instructs the agent to extract and format a LaTeX-compatible notations table from the paper, ensuring clarity, structural consistency, and integration within mathematical modeling documents.}
\label{prompt:PNCP}
\end{minipage}

\begin{tcolorbox}[colframe=blue!20!gray, colback=blue!5!white, coltitle=white, fonttitle=\bfseries, title=Paper Meta Information Prompt, breakable]
You are an expert academic writer tasked with analyzing paper chapters and generating key metadata for a mathematical modeling paper.\\
\\
\# Input Chapters\\
\{paper\_chapters\}\\
\\
Based on the content of these chapters, please generate:\\
1. A concise, descriptive title that reflects the paper's main focus\\
2. A comprehensive and detailed summary highlighting key findings and methodology\\
3. 4-6 relevant keywords that capture the paper's main themes\\
\\
Returns the Legal JSON Format:\\
```Json\\
\{\{\\
\text{\ \ \ \ }"title": "A clear, concise title",\\
\text{\ \ \ \ }"summary": "A well-structured summary covering the following information: $\textbackslash$n- Restatement and Clarification of the Problem: Describe the problem to be solved in your own words.$\textbackslash$n- Explanation of Assumptions and Their Rationality: Highlight the assumptions made in the modeling process and clearly list all the variables required for the model.$\textbackslash$n- Model Design and Rationality Argumentation: Specify the type of model used or describe the construction of a new model, explain how it was established and the rationale behind its design.$\textbackslash$n- Description of Model Testing and Sensitivity Analysis: Include error analysis and other testing items.",\\
\text{\ \ \ \ }"keywords": "keyword1; keyword2; keyword3; keyword4..."\\
\}\}\\
```\\
\\
Requirements:\\
- Title should be specific and academic in tone\\
- Summary should follow standard academic abstract structure and be approximately 400 words\\
- Keywords should be ordered from general to specific\\
- must return a strictly legal JSON
\end{tcolorbox}
\noindent\begin{minipage}{\linewidth}
\captionof{figure}{The prompt used for creating paper meta information. It instructs the agent to generate a publication-ready title, abstract, and keyword set from chapter content, producing a structured and legally formatted JSON summary aligned with academic conventions.}
\label{prompt:PMIP}
\end{minipage}
\end{document}